\documentclass[final,3p,times]{elsarticle_tech}
   
\usepackage{amsmath}  
\usepackage{overpic,rotating,contour,subfigure}                         
\usepackage{graphics,subfigure}
\usepackage{graphicx}
\usepackage{epsfig}
\usepackage{amssymb}  
\usepackage{epsfig} 
\usepackage{psfrag,rotating}
\usepackage{amssymb,floatflt,enumerate}
\usepackage{amsmath,amscd,psfrag,leqno} 
\usepackage[mathscr]{eucal}   
\usepackage{shadethm}   
\usepackage{graphicx,subfigure}
\usepackage{overpic,contour}

\def\Beweisende{\square}            
\def\BewEnde{\hfill{\Beweisende}}

\def\phm{{\hphantom{-}}} 

\def\phi{\varphi}

\def\RR{{\mathbb R}}

\def\CC{{\mathbb C}}

\def\Vkt#1{{\mathbf #1}} 

\newcommand{\go}[1]{{\sf #1}}

\definecolor{rot}{rgb}{1,0,0}

\newtheorem{thm}{Theorem}
\newtheorem{lem}{Lemma}
\newtheorem{definition}{Definition}
\newtheorem{ex}{Example}
\newtheorem{cor}{Corollary}

\newdefinition{rmk}{Remark}
\newproof{pf}{Proof}
\newproof{pot}{Proof of Theorem \ref{thm2}}

\begin{document} 
 
\begin{frontmatter}
 
\title{Self-motions of pentapods with linear platform}

\author[TU]{Georg Nawratil}
\author[RADON]{Josef Schicho}
\address[TU]{Institute of Discrete Mathematics and Geometry, Vienna University of Technology, 
Wiedner Hauptstrasse 8-10/104, 1040 Vienna, Austria}
\address[RADON]{Johann Radon Institute for Computational and Applied Mathematics, Austrian Academy of Sciences, 
Altenberger Strasse 69,
4040 Linz, Austria}

\begin{abstract} 
We give a full classification of all pentapods with linear platform possessing a self-motion 
beside the trivial rotation about the platform. 
Recent research necessitates a contemporary and accurate re-examination of old results on this topic given by Darboux, Mannheim, Duporcq and Bricard, which 
also takes the coincidence of platform anchor points into account.   
For our study we use bond theory with respect to a novel kinematic mapping for pentapods with linear platform, beside the 
method of  singular-invariant leg-rearrangements. Based on our results we design  
pentapods with linear platform, which have a simplified direct kinematics concerning their number of (real) solutions. 
\end{abstract}

\begin{keyword}
Pentapod, Borel Bricard Problem, Bond Theory, Self-Motion, Direct Kinematic, Kinematic Mapping
\end{keyword}

\end{frontmatter}

\section{Introduction}\label{intro}
 
The geometry of a pentapod (see Fig.\ \ref{fig0}a) is given by the five base anchor points $\go M_i$ with coordinates
$\Vkt M_i:=(A_i,B_i,C_i)^T$ with respect to the fixed system $\Sigma_0$ and by the five collinear platform anchor points $\go m_i$ 
with coordinates $\Vkt m_i:=(a_i,0,0)^T$ with respect to the moving system $\Sigma$ (for $i=1,\ldots ,5$). 
All pairs $(\go M_i,\go m_i)$ of corresponding anchor points are connected by SPS-legs (or alternatively SPU-legs with  aligned universal joints; 
cf.\ \cite[Fig.\ 1]{btt}), where only the prismatic joints are active.

If the geometry of the manipulator is given, as well as the lengths of the five pairwise distinct legs, a 
pentapod has generically mobility 1 according to the formula of Gr\"ubler. 
In the discussed case of pentapods with linear platform the degree of freedom corresponds to the rotation about 
the carrier line $\go p$ of the five platform anchor points. 
This rotational motion is irrelevant for applications with axial symmetry as e.g.\ 5-axis milling, spot-welding, laser or water-jet engraving/cutting, 
spray-based painting, etc.\ (cf.\ \cite{borras3,baer}). 
Therefore these mechanisms are of great practical interest. 
In this context configurations should be avoided, where the manipulator gains an additional uncontrollable mobility (beside the rotational motion around $\go p$), 
which is referred as self-motion within this article. 
Before we give a review on pentapods possessing these special motions in Section \ref{rev:pent_self}, we repeat a few basics in geometry, which 
are essential for the understanding of the paper.

\subsection{Geometric basics}\label{geom:basic}
\begin{figure}[top]
\begin{center} 
\subfigure[]{ 
 \begin{overpic}
    [width=55mm]{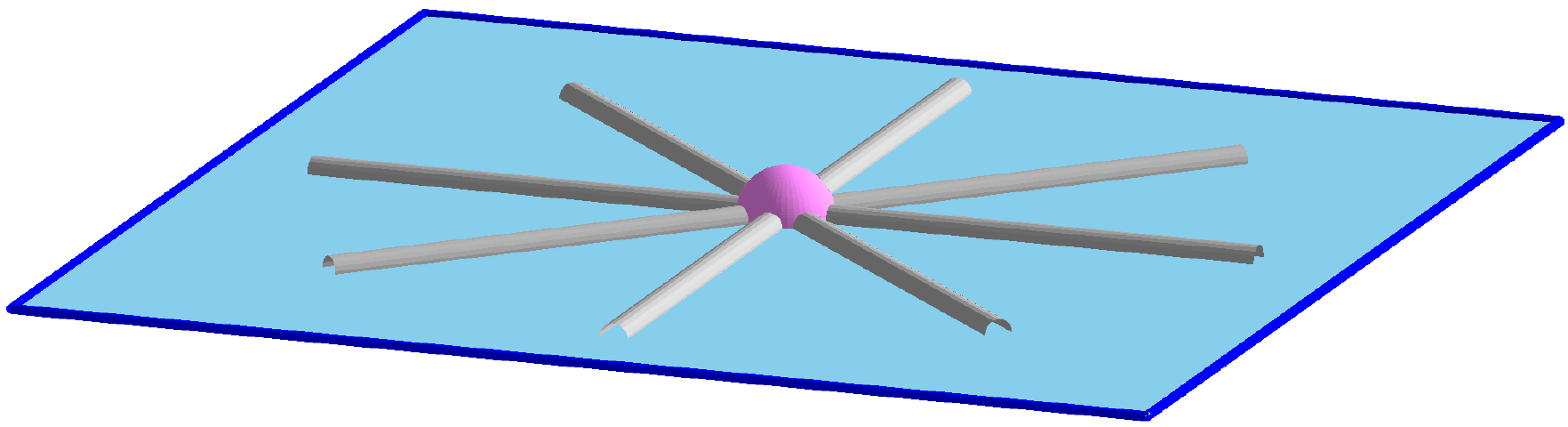}
  \end{overpic} 
 }
\quad
\subfigure[]{
\begin{overpic}
    [width=43mm]{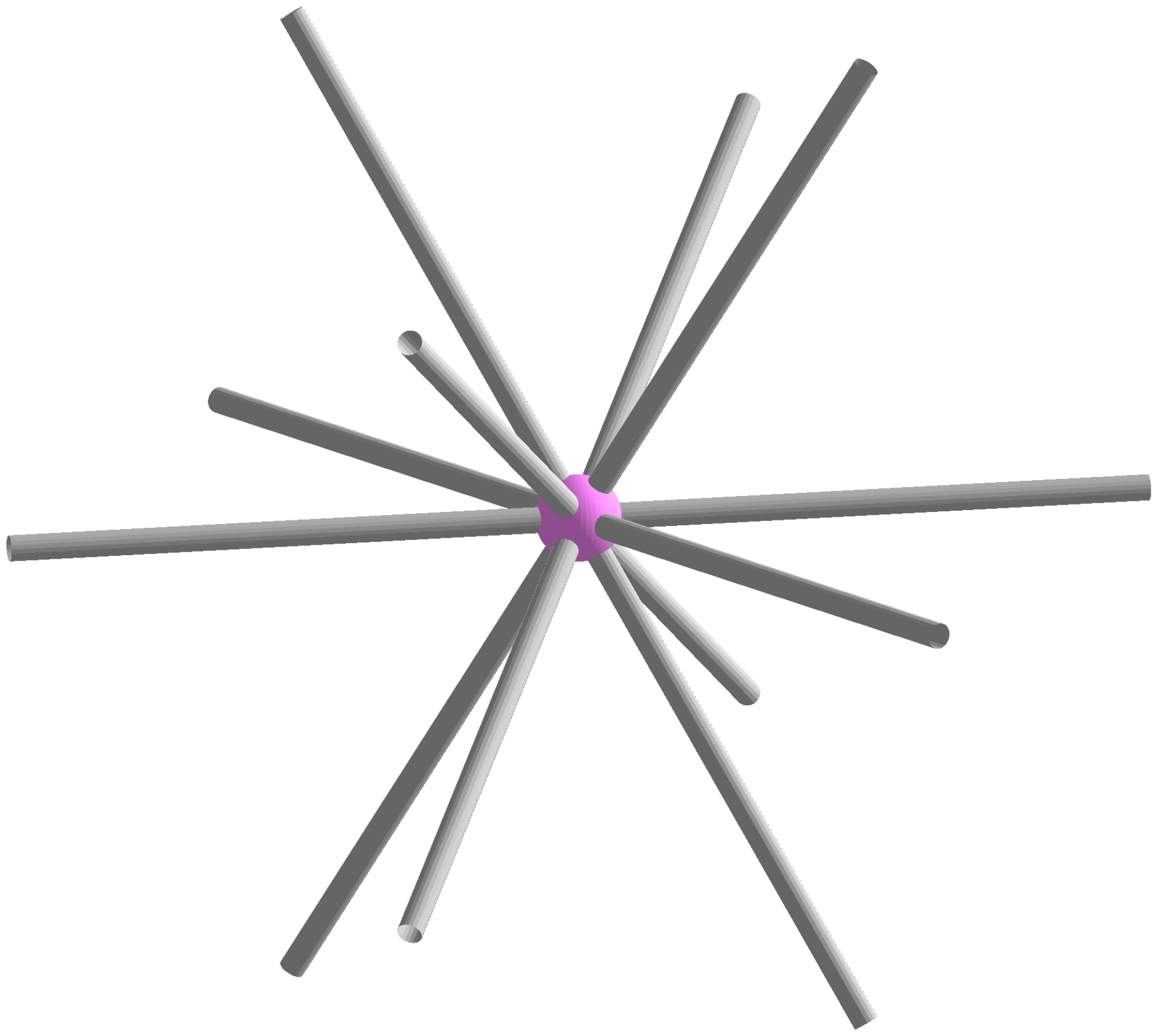}
  \end{overpic}
} 
\quad
\subfigure[]{
\begin{overpic}
    [width=30mm]{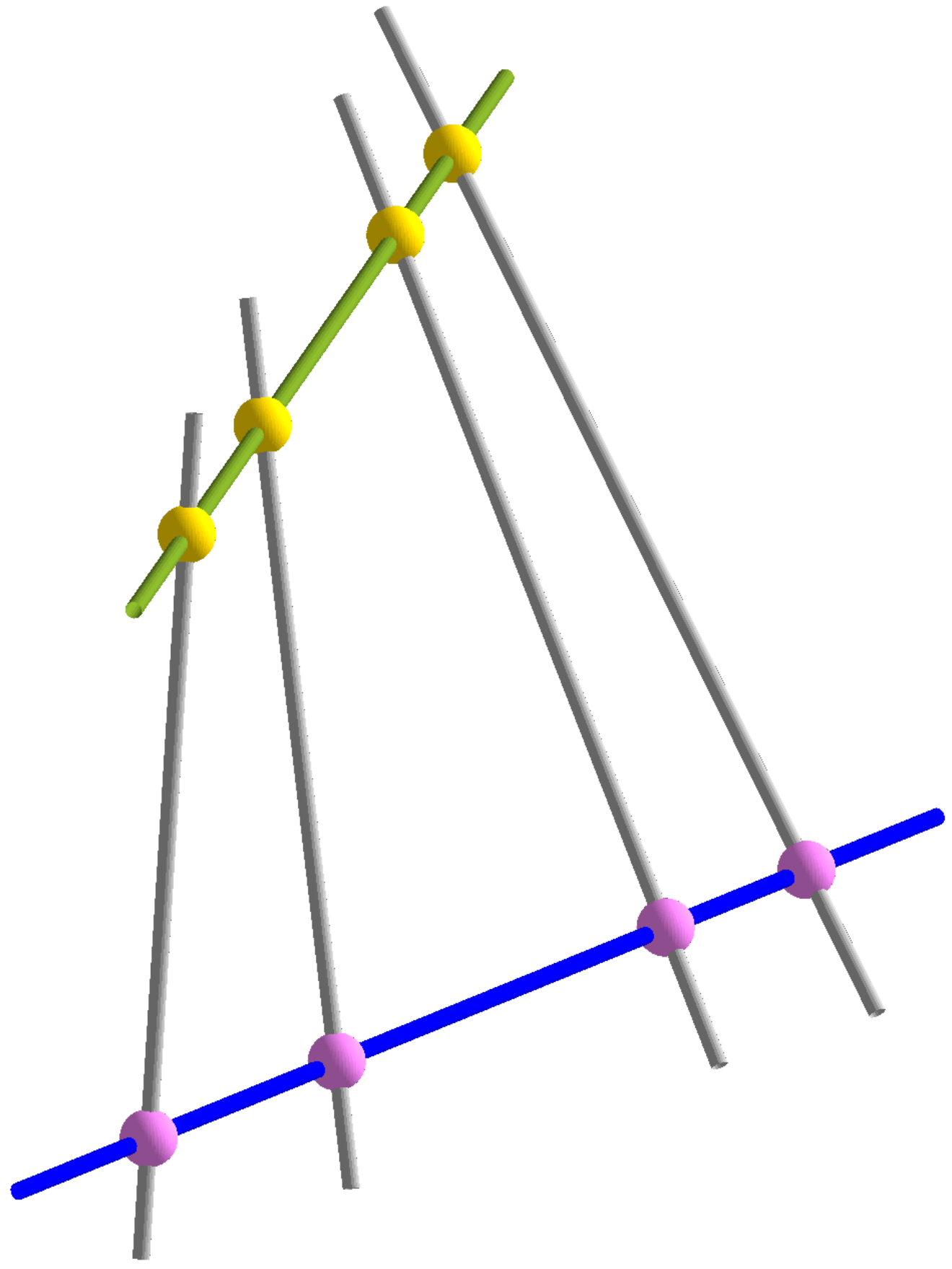}
\begin{small}
\put(-7.5,4){$\go g\kappa$}
\put(13,3){$\go G_i\kappa$}
\put(29,9){$\go G_j\kappa$}
\put(36.5,27){$\go G_k\kappa$}
\put(67.5,25.5){$\go G_l\kappa$}
\put(5,47){$\go g$}
\put(4,58){$\go G_i$}
\put(24,63.5){$\go G_j$}
\put(20,80){$\go G_k$}
\put(39,84){$\go G_l$}
\end{small} 
  \end{overpic}
}  

\end{center} 
\caption{(a) A pencil of lines is the 1-parametric set of all coplanar lines 
through a common point, which is the co-called vertex of the pencil. 
(b) A bundle of lines is the 2-parametric set of all lines 
through a common point, which is the co-called vertex of the bundle. 
(c) A regulus of lines. 
}
  \label{fig-1}
\end{figure}
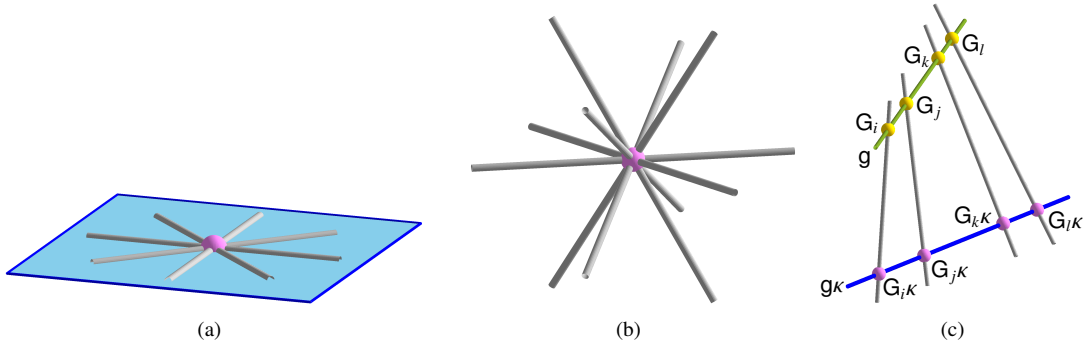

We consider the projective closure of the Euclidean 3-space, which means that we add a point at infinity  to each line, which is a so-called ideal point. 
Moreover two lines are parallel or coincide if and only if they have the same ideal point. 
The set of ideal points of a pencil of lines (cf.\ Fig.\ \ref{fig-1}a) 
form a so-called ideal line of the carrier plane of this pencil.  
Again two planes are parallel or coincide if and only if they have the ideal line in common. 
The set of ideal points of a bundle of lines (cf.\ Fig.\ \ref{fig-1}b) 
constitute the so-called ideal plane. 
Summed up we can say that we obtain the projective closure 
by addition of the ideal plane. Points, lines and planes, which are no ideal elements are called finite. 

We can also introduce projective point coordinates by homogenizing the coordinates $(x,y,z)$ of a finite point by 
$(1:x:y:z)$. The ideal point of a finite line in direction $(u,v,w)$ has the coordinates $(0:u:v:w)$. 
Now we can define a regular/singular projectivity (projective mapping) by multiplication of the projective point coordinates
with a regular/singular $4\times 4$ matrix. 
If the set of ideal points is mapped onto itself, then the projectivity is called an affinity 
(affine transformation).

Under a regular projectivity a line $\go g$ is mapped onto a line $\go g\kappa$ due to the linearity of the mapping. 
The set of lines $[\go G_i,\go G_i\kappa]$ with $\go G_i\in\go g$ is called a regulus of lines (cf.\ Fig.\ \ref{fig-1}c) if $\go g$ and $\go g\kappa$ are skew. 
In this context it should be noted that the cross-ratio (CR) of four collinear points (cf.\ \cite[page 20]{pottmann_wallner}) 
is invariant under a regular projectivity; i.e.\ 
$CR(\go G_i,\go G_j,\go G_k,\go G_l)=CR(\go G_i\kappa,\go G_j\kappa,\go G_k\kappa,\go G_l\kappa)$.

Conic sections are well-known geometric objects. If the ideal line of the carrier plane of the conic section touches 
the conic, then it is a parabola. If there are conjugate complex (resp.\ two real) intersection points, we get an 
ellipse (resp.\ a hyperbola). The conjugate complex intersection points of a circle with the ideal line are 
the co-called cyclic points of the circle's carrier plane. 

A further geometric object used within the article at hand is a so-called cubic ellipse. 
According to \cite{gruenwald} a cubic ellipse is a space curve of degree 3, which intersects the ideal plane in one 
real and two conjugate complex ideal points\footnote{Therefore the projection in direction of the real ideal point yields an ellipse.}. 
If the latter ones are the cyclic points of a plane (not) orthogonal to the 
direction of the real ideal point, then the cubic ellipse is called straight cubic circle (skew cubic circle). 
For more details we refer to \cite{rulf}.

Finally we need the notation of a so-called M\"obius transformation $\tau$ of the plane. If we combine the planar Cartesian coordinates $(u,v)$ to a 
complex number $w:=u+iv$, then $\tau(w)$ can be defined as a  rational function of the form 
\begin{equation}\label{tau}
\tau:\quad w\mapsto \frac{z_1w+z_2}{z_3w+z_4},
\end{equation}
with complex numbers $z_1, \ldots, z_4$ satisfying $z_1z_4 - z_2z_3 \neq 0$. 
Moreover it should be noted that $\tau$ maps straight lines onto straight lines or circles and 
that a M\"obius transformation is uniquely defined by three pairwise distinct points $w_1,w_2,w_3$ and their 
pairwise distinct images.

\subsection{Review, motivation and outline}\label{rev:pent_self}
The self-motions of pentapods with linear platform represent interesting solutions to the 
still unsolved problem posed by the French Academy of Science for the {\it Prix Vaillant} of the year 1904, which is also known as 
Borel-Bricard problem (cf.\ \cite{borel,bricard,husty_bb}) and reads as follows: 
{\it "Determine and study all displacements of a rigid body in which distinct points of the body move on spherical paths."}

For the special case of five collinear points the Borel-Bricard problem seemed to be solved since more than 100 years, 
due to the following results (cf.\ \cite[page 415]{krames}): If five points of a line have
spherical trajectories then this property holds for all points of the line. The centers are located on
a straight line (cf.\ Darboux \cite[page 222]{koenigs}), a conic section (cf.\ Mannheim \cite[pages 180ff]{mannheim}) 
or a straight cubic circle (cf.\ Duporcq \cite{duporcq}; see also Bricard \cite[Chapter III]{bricard}).

\begin{figure}[top]
\begin{center} 
\subfigure[]{ 
 \begin{overpic}
    [width=37mm]{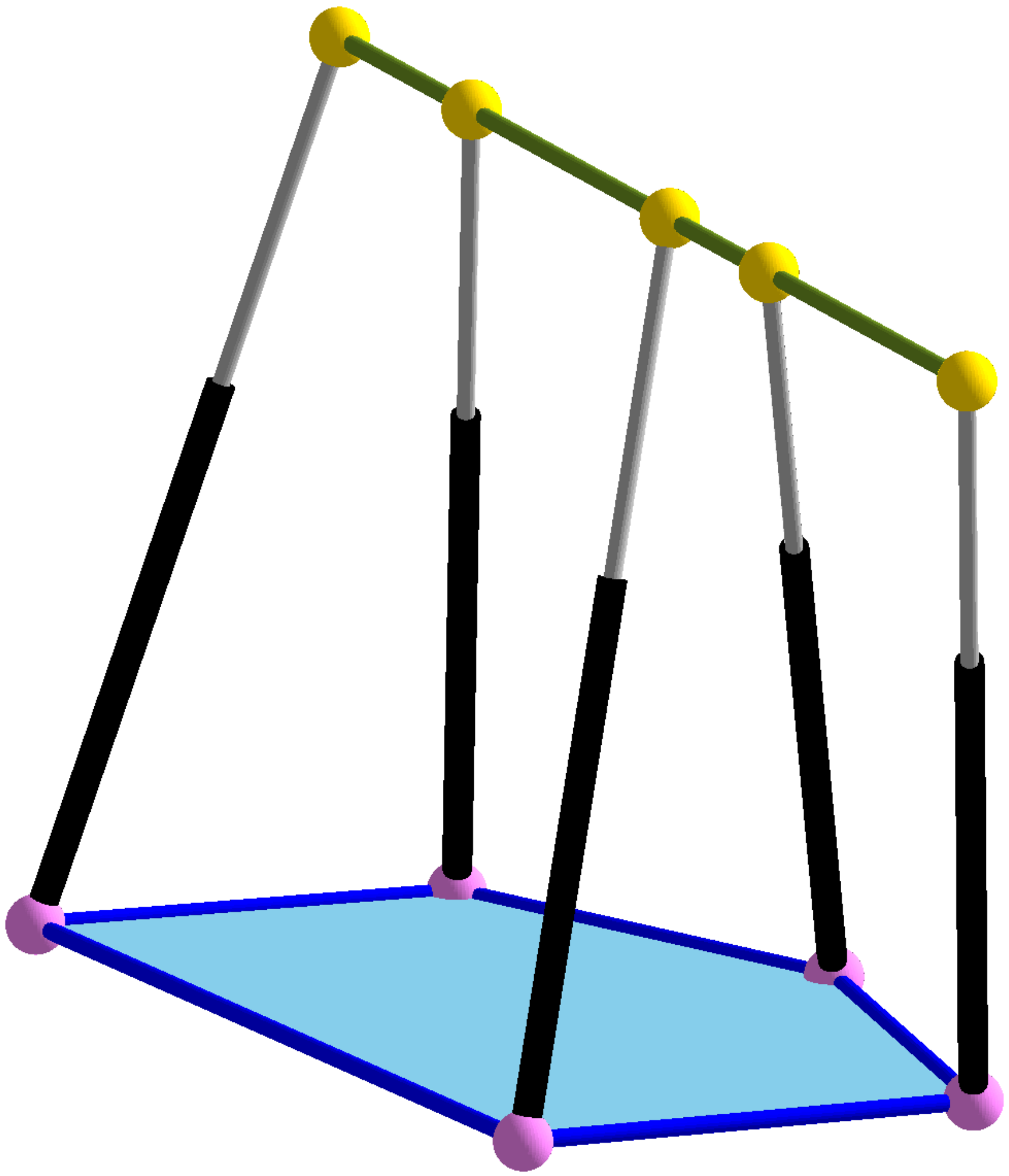}
\begin{small}
\put(32,0){$\go M_3$}
\put(87,4){$\go M_1$}
\put(60.5,21){$\go M_2$}
\put(1,13){$\go M_5$}
\put(28.5,27){$\go M_4$}
\put(17.5,95){$\go m_5$}
\put(44,90.5){$\go m_4$}
\put(45.5,80.5){$\go p$}
\put(60,82){$\go m_3$}
\put(69,76.5){$\go m_2$}
\put(86,66){$\go m_1$}
\end{small} 
  \end{overpic}
}
\hfill
\subfigure[]{
\begin{overpic}
    [width=35mm]{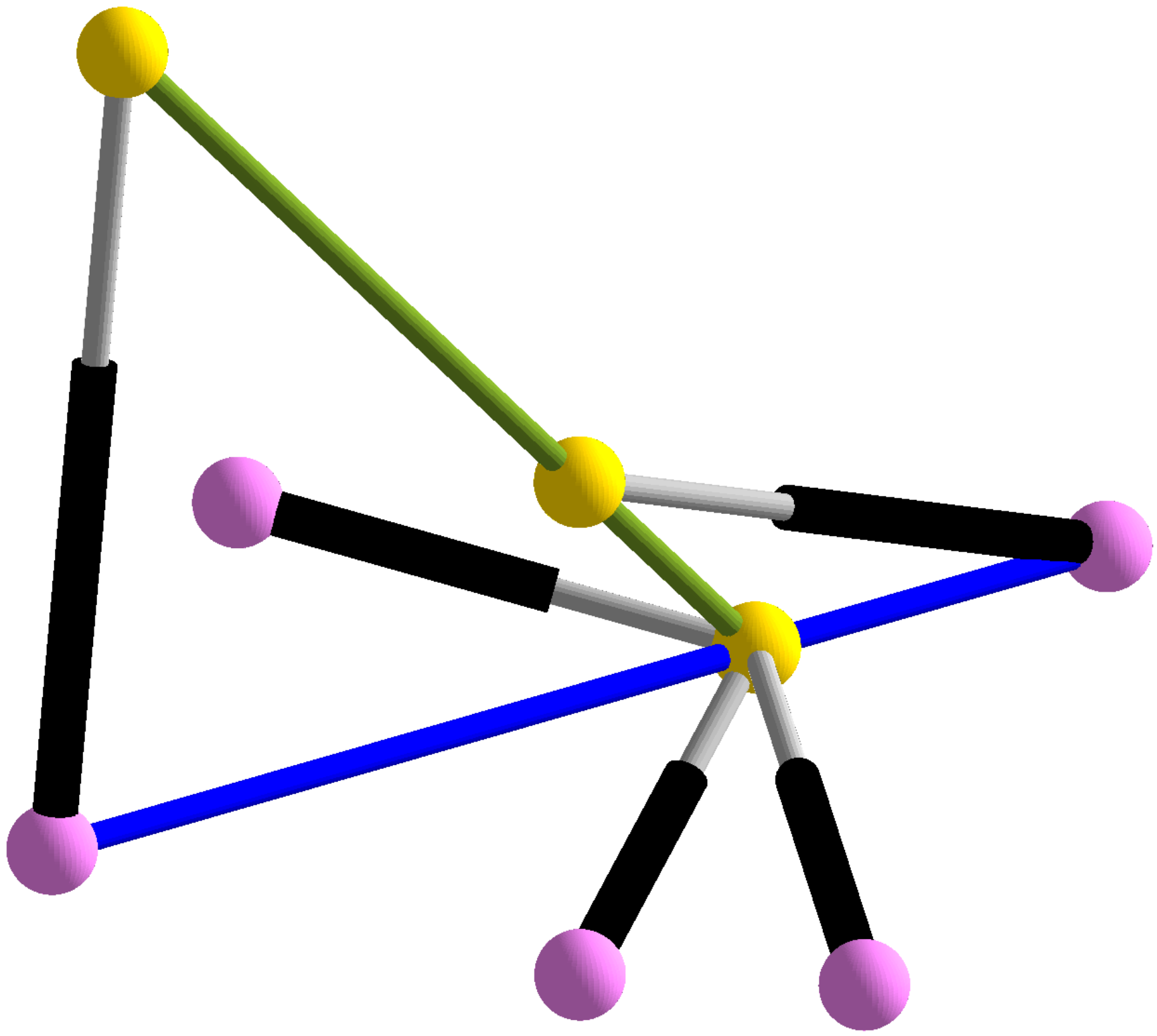}
\begin{small}
\put(91,48){$\go M_4$}
\put(79.5,1){$\go M_1$}
\put(2,4){$\go M_5$}
\put(16.5,35){$\go M_3$}
\put(36,1.5){$\go M_2$}
\put(16,84){$\go m_5$}
\put(76,18){$=\go m_3$}
\put(49.5,52.7){$\go m_4$}
\put(71,27){$\go m_1=\go m_2$}
\put(33.5,15.5){$\go h$}\end{small} 
  \end{overpic}
} 
\hfill
\subfigure[]{
\begin{overpic}
    [width=35mm]{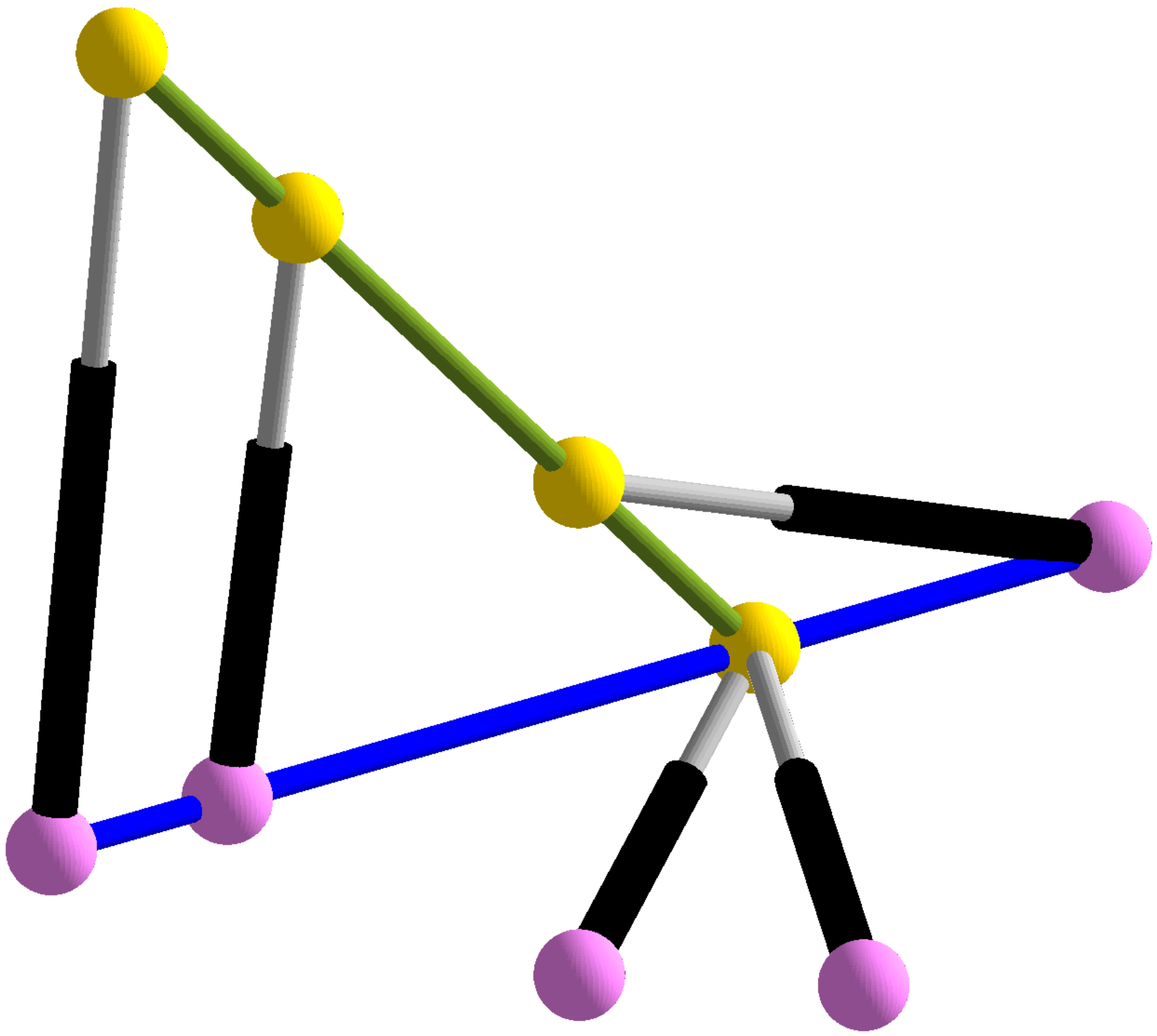}
\begin{small}
\put(91,48){$\go M_4$}
\put(79.5,1){$\go M_1$}
\put(2,4){$\go M_5$}
\put(17.5,9){$\go M_3$}
\put(36,1.5){$\go M_2$}
\put(16,84){$\go m_5$}
\put(30.5,70){$\go m_3$}
\put(49.5,52.7){$\go m_4$}
\put(71,27){$\go m_1=\go m_2$}
\put(33.5,15.5){$\go h$}
\end{small} 
  \end{overpic}
} 
\hfill
\subfigure[]{
\begin{overpic}
    [width=35mm]{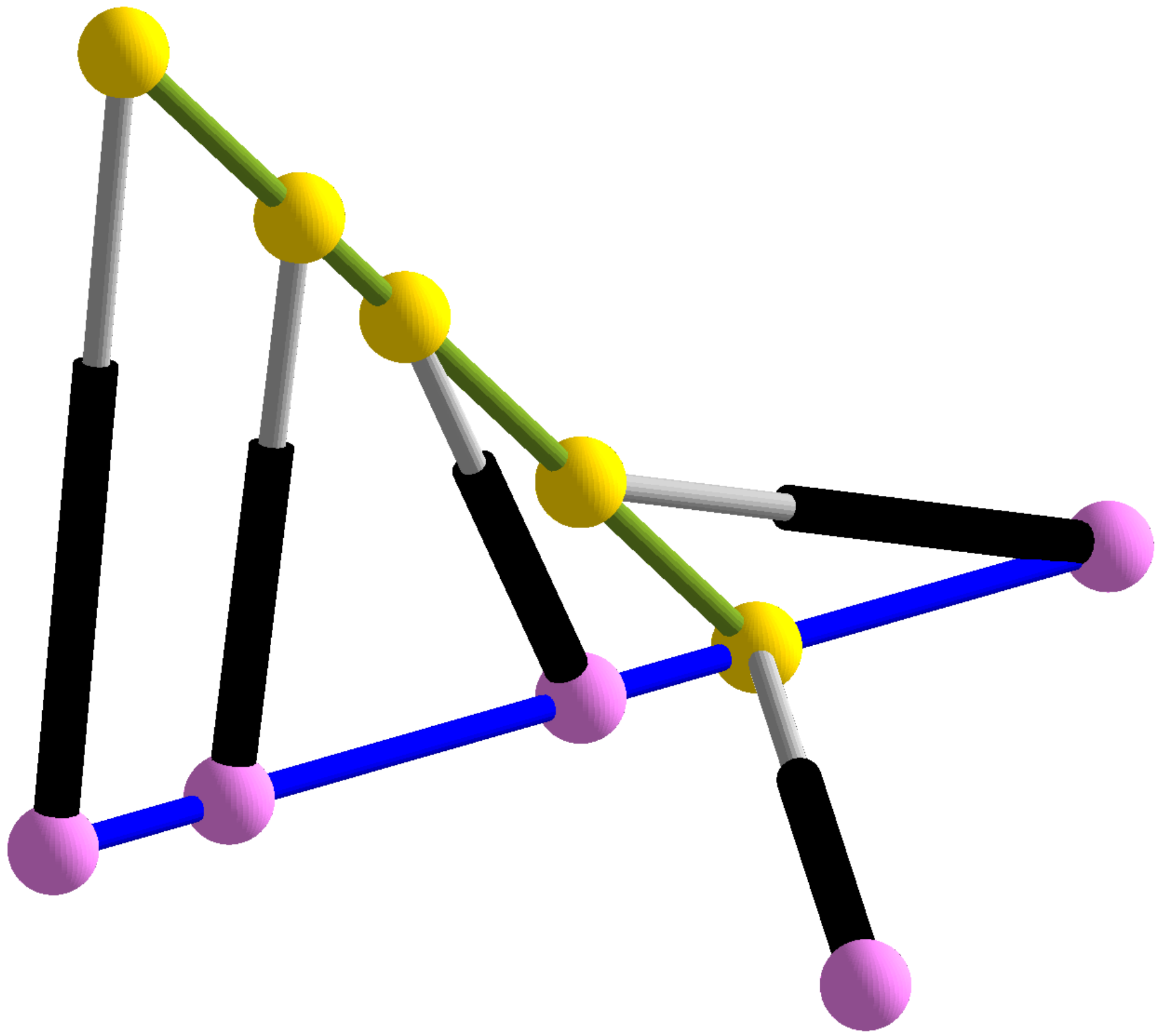}
\begin{small}
\put(91,48){$\go M_4$}
\put(79.5,1){$\go M_1$}
\put(2,4){$\go M_5$}
\put(17.5,9){$\go M_3$}
\put(47.5,17.5){$\go M_2$}
\put(16,84){$\go m_5$}
\put(30.5,70){$\go m_3$}
\put(40,61.5){$\go m_2$}
\put(49.5,52.7){$\go m_4$}
\put(71,27){$\go m_1$}
\put(33.5,15.5){$\go h$}
\end{small} 
  \end{overpic}
} 
\end{center} 
\caption{(a) Pentapod with linear platform $\go p$ (and planar base). 
(b,c,d) The designs ($\alpha,\beta,\gamma$) are illustrated in a pose of a self-motion.}
  \label{fig0}
\end{figure}
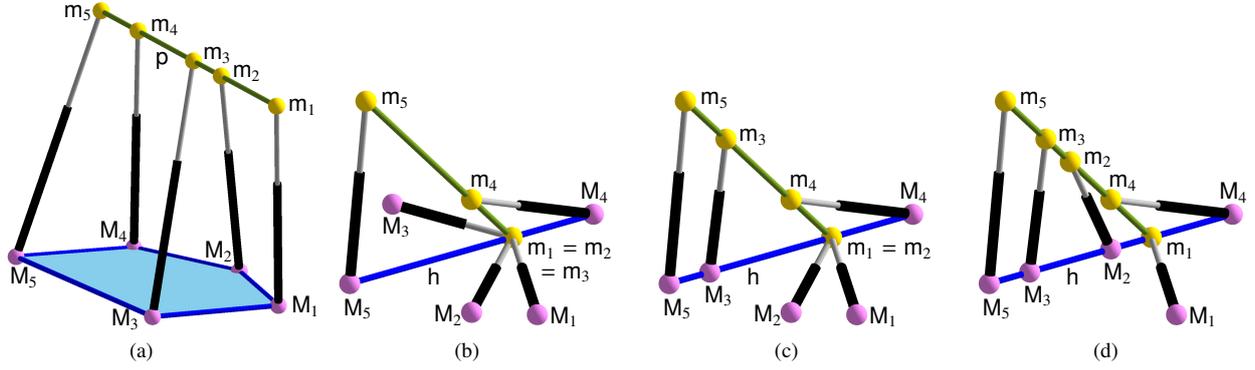

In a recent publication \cite{gns1} the authors determined all pentapods with mobility 2, where neither all platform anchor points nor 
all base anchor points are collinear. As a side result of this study we obtained the following three designs of 
pentapods with linear platform possessing a self-motion of $\go p$:
\begin{itemize}
	\item[($\alpha$)]
	$\go m_1=\go m_2=\go m_3$:  The self-motion is obtained if
	$\go m_1=\go m_2=\go m_3$ is located on the line $\go h$ spanned by $\go M_4$ and $\go M_5$. 
	\item[($\beta$)]
	$\go m_1=\go m_2$ and $\go M_3,\go M_4,\go M_5$ collinear:  The self-motion is obtained if
	$\go m_1=\go m_2$ is located on the line $\go h$  spanned by $\go M_3,\go M_4,\go M_5$.
	\item[($\gamma$)]
	$\go M_2,\go M_3,\go M_4,\go M_5$ collinear: The self-motion is obtained if 
	$\go m_1$ is located on the line $\go h$  spanned by $\go M_2,\go M_3,\go M_4,\go M_5$.	
\end{itemize}
In all three cases, which are illustrated in Fig.\ \ref{fig0}, the following legs can be added without restricting the spherical self-motion: 
Every point of $\go p$ can be connected with any point of the line $\go h$ with exception of 
the point $\go m_1$ (= center of spherical motion), which can be linked with any point of the fixed 3-space.

This already shows that the cases ($\alpha,\beta,\gamma$) are not covered by the above cited results of 
Darboux, Mannheim, Duporcq and Bricard, respectively.
The reason for this is partially hidden in the formulation of the problem, as 100 years ago they studied the conditions for
five points of a line to run on spherical trajectories, which already implies that the collinear points are pairwise distinct. 
Therefore they missed the cases ($\alpha,\beta$), but this still does not explain the absence of case ($\gamma$). 
All in all this shows the need of a contemporary and accurate re-examination of the old results, which 
also takes the coincidence of platform anchor points into account. 
This is done within the paper at hand, which is structured as follows: 

In Section \ref{sec:archsing} we give a short review on architecturally singular pentapods as they imply further solutions to our problem. 
Based on the method of singular-invariant leg-replacements we determine  in Section \ref{p5coplanar}
all non-architecturally singular pentapods with linear platform and planar base possessing self-motions. 
The same method is used in Section \ref{p5notcoplanar} to distinguish five different types of pentapods with linear platform and non-planar base.
In Section \ref{sec:dkp} we introduce the theory of bonds with respect to a novel kinematic mapping for pentapods with linear platform. 
This theory is used in Section \ref{sec:contemp_duporcq} for the determination of all non-architecturally singular pentapods with linear 
platform and non-planar base possessing self-motions.
Finally we use the presented results/methods to design pentapods with a linear platform, 
which have a simplified direct kinematics with respect to the number of (real) solutions (cf.\ Section \ref{conclusion}).

\subsection{Architecturally singular pentapods}\label{sec:archsing}

A pentapod is called architecturally singular if in any pose of the platform the rank of its Jacobian matrix is less than five.
This condition also has a line-geometric interpretation as the Jacobian is composed of the Pl\"ucker coordinates of 
the five carrier lines of the legs (cf.\ \cite{pottmann_wallner}). 
All architecturally singular pentapods are known (cf.\ \cite[Theorem 3]{kargernonplanar} under consideration of \cite{nawratilpenta}) 
as well as their properties of being redundant and invariant under projectivities of the platform and the base.
If we take additionally the collinearity of the platform into account (i.e.\ $\go m_1,\ldots, \go m_5$ collinear) we end up with the following list: 

\begin{cor}\label{cor:arch}
If a pentapod with linear platform is architecturally singular, then it has one of the following designs
\footnote{After a possible necessary renumbering of anchor points.}:
\begin{enumerate}
\item
$\go M_1=\go M_2=\go M_3$: The first three legs belong to a pencil of lines. 
\item
$\go m_1=\go m_2=\go m_3$ and $\go M_1,\go M_2,\go M_3$ are collinear: The first three legs belong to a pencil of lines.
\item
$\go M_1,\ldots,\go M_4$ are collinear and the following 
cross-ratio condition holds: 
\begin{equation}
CR(\go m_1,\go m_2,\go m_3,\go m_4)=CR(\go M_1,\go M_2,\go M_3,\go M_4).
\end{equation}
The first four lines belong to a regulus of lines. In the special case $\go m_1=\go m_2$ and 
$\go M_3=\go M_4$ the regulus splits up into two pencils of lines. 
\item
$\go m_1=\go m_2=\go m_3=\go m_4$: The first four legs belong to a bundle of lines.
\item
$\go M_1,\ldots,\go M_5$ are collinear.
\item
$\go m_1=\go m_2=\go m_3$ and $\go M_4=\go M_5$.  
\item
$\go m_1=\go m_2$ and $\go m_4=\go m_5$. Moreover $\go M_1,\go M_2,\go M_5$ are collinear 
and $\go M_3,\go M_4,\go M_5$ are collinear.
\item
$\go m_1,\ldots,\go m_5$ are pairwise distinct, $\go M_1,\ldots,\go M_5$ are coplanar and no three of them are collinear. 
Moreover there is a projective correspondence between the points $\go m_i$ and $\go M_i$ for $i=1,\ldots ,5$. 
For more details on this most complicated case we refer to \cite[Item 10 of Theorem 3]{kargernonplanar} and \cite{borras_arch}. 
\item
$\go m_4=\go m_5$ holds and $\go M_1,\ldots,\go M_5$ are coplanar, where $\go M_1,\go M_2,\go M_3$ are collinear. 
Moreover the following cross-ratio condition holds: 
\begin{equation}
CR(\go m_1,\go m_2,\go m_3,\go m_4=\go m_5)=CR(\go M_1,\go M_2,\go M_3,\go M),
\end{equation}
with $\go M$ denoting the intersection point of $[\go M_4,\go M_5]$ and the carrier line of $\go M_1,\go M_2,\go M_3$. 
\end{enumerate}
In the cases (5-9) the five legs belong to a so-called congruence of lines (cf.\ \cite[Section 3.2.1]{pottmann_wallner}).
\end{cor}

Due to the above mentioned redundancy all these nine cases imply solutions to our problem, as they have a self-motion in 
each pose of $\go p$ (over $\CC$). Moreover, the architecturally singular cases 2,4,6,7,9 are also not covered by 
the old results due to the coincidence of platform anchor points. The remaining cases are discussed in more detail: 

\begin{itemize}
\item[Ad 3:] 
All lines of the regulus can be added without restricting the self-motion. Therefore this case corresponds to the 
result of Darboux \cite[page 222]{koenigs}. 
\item[Ad 8:]
The projective correspondence can be extended to all points of the linear platform $\go p$ and therefore they are mapped onto 
a conic determined by $\go M_1,\ldots ,\go M_5$. Now all legs connecting corresponding anchor points 
can be attached without changing the self-motion. This equals the solution given by Mannheim \cite[pages 180ff]{mannheim}.
\item[Ad 1:] This case can be interpreted as a special case of Mannheim's solution, as following legs can be added without restricting the self-motion: 
Every point of $\go p$ can be connected with $\go M_1=\go M_2=\go M_3$ with exception of the point $\go m_i$ which can  
be linked with any point of the line $[\go M_1,\go M_i]$ for $i=4,5$. 
Therefore the conic splits up into the two lines $[\go M_1,\go M_4]$ and $[\go M_1,\go M_5]$. 
\item[Ad 5:]
This trivial case can also be seen as a  special case of Mannheim's solution, as the conic degenerates into the double counted carrier line of 
$\go M_1,\ldots ,\go M_5$.
\end{itemize}

As all architecturally singular pentapods with linear platform are already known, we can restrict our study done in the remainder of the article 
to non-architecturally singular manipulators. 
Moreover, as the designs ($\alpha,\beta,\gamma$) are not architecturally singular, we can make the following three additional assumptions in order 
to exclude these already known cases:  
\begin{enumerate}[(i)]
\item
No three platform anchor points coincide.
\item
If two platform anchor points coincide, the remaining three base anchor points are not collinear.
\item
No four base anchor points are collinear. 
\end{enumerate}

\begin{definition}\label{df:P}
We define by $\mathcal{P}$ the set of all non-architecturally singular pentapods with a linear platform, which fulfill the assumptions (i,ii,iii). 
\end{definition}

We split the determination of all elements of $\mathcal{P}$ with self-motions in two parts with respect to the criterion if the base
anchor points are coplanar (= planar pentapod; e.g.\ Fig.\ \ref{fig0}a,d) or not (= non-planar pentapod; e.g.\ Fig.\ \ref{fig0}b,c).


\section{Planar pentapods of $\mathcal{P}$ with self-motions}\label{p5coplanar}

Within this section we prove the following theorem:

\begin{thm}\label{thm:planar}
A planar pentapod of $\mathcal{P}$ has a self-motion only in the following case: 
There exists an orthogonal-projection $\pi_{\varepsilon}$ of the base plane $\varepsilon$ and an 
orthogonal-projection $\pi_{\go p}$ of $\go p$ in a way that the projected point sets are congruent. 
In this case $\go p$ can perform a circular translation.
\end{thm}

For the proof of this theorem the following preparatory work has to be done:

\begin{lem}\label{prep} 
The anchor points of a planar pentapod with a linear platform, which fulfills the assumptions (i,ii,iii), 
can always be relabeled in a way that the following conditions hold: 
\begin{equation*}
\go M_1\neq \go M_2, \qquad
\go M_1,\go M_2,\go M_3\,\, \text{not collinear}, \qquad
\go M_1,\go M_2,\go M_4\,\, \text{not collinear}, \qquad
\go m_3\neq\go m_4.
\end{equation*}
\end{lem}

\noindent
{\sc Proof:} If two platform anchor points coincide we denote them with $\go m_4=\go m_5$. 
Then due to assumption (ii) $\go M_1,\go M_2,\go M_3$ are not collinear. 
Due two assumption (iii) one of the remaining two base points is not on the line 
spanned by $\go M_1$ and $\go M_2$. We denote this point by $\go M_4$ and we are done. 

Now we discuss the case where all five platform anchor points are pairwise distinct: 
\begin{enumerate}[$\star$]
\item
If two base points coincide\footnote{No three base points can coincide as it yields a contradiction to assumption (iii).} 
then we denote them with $\go M_3=\go M_4$. Due to assumption (iii) there are at least two further base points which span 
together with  $\go M_3=\go M_4$ a plane. We denote these points by $\go M_1$ and $\go M_2$, respectively. 
\item
If no base points coincide, but three of them are collinear, then we denote them by $\go M_1,\go M_2$ and $\go M_5$. 
Due to assumption (iii) we are done. 
\item
If no three base points are collinear, we can label the points arbitrarily. 
\hfill $\BewEnde$
\end{enumerate}

\noindent
Moreover we can choose the moving frame $\Sigma$ in a way that $\go m_1$ equals its origin. 
The fixed frame $\Sigma_0$ is selected in a way that $\go M_1$ 
equals the origin, $\go M_2$ is located on the $x$-axis and the remaining points belong to the 
$xy$-plane. Due to Lemma \ref{prep} we can assume w.l.o.g. that $A_2B_3B_4(a_3-a_4)\neq 0$ holds.

The proof of Theorem \ref{thm:planar} is based on the following result obtained by Borras et al.\ \cite{borras_arch}:  
{\it A leg of a given planar pentapod with linear platform is replaced by a leg with platform anchor point $(a,0,0)$ and base anchor point $(A,B,0)$ 
fulfilling  Eq.\ (6) of \cite{borras_arch}, which reads as follows under consideration of our special choice of 
coordinate systems $\Sigma$ and $\Sigma_0$: 
\begin{equation}\label{centraleq}
\left[
(D_2,D_3,0)+a(D_4,D_5,D_1)
\right]
\begin{pmatrix}
A \\ 
B \\ 
1
\end{pmatrix}=0
\end{equation}
with
\begin{equation}
\begin{split}
D_1:=det(\Vkt A, \Vkt B, \Vkt a\Vkt A ,\Vkt a\Vkt B), \quad
D_2:=-det(\Vkt a, \Vkt B, \Vkt a\Vkt A ,\Vkt a\Vkt B),\quad
D_3:=det(\Vkt a, \Vkt A, \Vkt a\Vkt A ,\Vkt a\Vkt B),\\
D_4:=-det(\Vkt a, \Vkt A, \Vkt B ,\Vkt a\Vkt B),\quad
D_5:=det(\Vkt a, \Vkt A, \Vkt B ,\Vkt a\Vkt A),
\end{split}
\end{equation}
and
\begin{equation}
\Vkt a:=\begin{pmatrix}
a_2 \\ a_3\\ a_4\\a_5 \end{pmatrix}, \quad
\Vkt A:=\begin{pmatrix}
A_2 \\ A_3\\ A_4\\A_5 \end{pmatrix}, \quad
\Vkt B:=\begin{pmatrix}
0 \\ B_3\\ B_4\\B_5 \end{pmatrix}, \quad
\Vkt a\Vkt A:=\begin{pmatrix}
a_2A_2 \\ a_3A_3\\ a_4A_4\\a_5A_5 \end{pmatrix}, \quad
\Vkt a\Vkt B:=\begin{pmatrix}
0 \\ a_3B_3\\ a_4B_4\\a_5B_5 \end{pmatrix},
\end{equation}
then the resulting planar pentapod with linear platform has the same singularity set (and direct kinematics solution) 
if it is not architecturally singular.} \\

\noindent
In the following we study Eq.\ (\ref{centraleq}) in more detail:

\begin{lem}\label{arch} 
A planar pentapod with linear platform fulfilling the assumptions (i,ii,iii) 
is architecturally singular if and only if $D_1=D_4=0$ holds (with respect to $\Sigma$ and $\Sigma_0$).
\end{lem}

\noindent
{\sc Proof:} 
Due to Lemma \ref{prep} we can assume w.l.o.g.\ that  the following determinant
\begin{equation}
\begin{vmatrix}
A_2 & 0 & 0  \\
A_3 & B_3 & a_3B_3  \\
A_4 & B_4 & a_4B_4  
\end{vmatrix}=A_2B_3B_4(a_3-a_4)
\end{equation}
is different from zero. Therefore $D_1=D_4=0$ implies $rk(\Vkt a,\Vkt A, \Vkt B, \Vkt a\Vkt A ,\Vkt a\Vkt B)<5$ 
which characterizes architecturally singularity (cf.\ \cite{nawratilpenta,borras_arch,roschel_mick}). 
\hfill $\BewEnde$\\

\noindent
Based on this lemma we can prove the next one, which reads as follows:

\begin{lem}\label{d2d3} 
For a planar pentapod of $\mathcal{P}$ the condition $D_2=D_3=0$ cannot hold (with respect to $\Sigma$ and $\Sigma_0$). 
\end{lem}

\noindent
{\sc Proof:} 
The proof is done by contradiction as follows: We show that for a planar pentapod with linear platform the condition $D_2=D_3=0$ (with respect to 
$\Sigma$ and $\Sigma_0$ $\Rightarrow$ $A_2B_3B_4(a_3-a_4)\neq 0$) implies either an architecturally singular design ($\Leftrightarrow$ $D_1=D_4=0$; cf.\ Lemma \ref{arch}) 
or a contradiction to the assumptions (i,ii,iii). 

It can easily be seen that $D_2=0$ is fulfilled for $a_2=0$ ($\Leftrightarrow$ $\go m_1=\go m_2$). Now  
$D_3=0$ simplifies to:
\begin{equation}
-a_3a_4a_5A_2(A_4B_5 - A_5B_4 - A_3B_5 + A_5B_3 + A_3B_4 - A_4B_3).
\end{equation}
$a_3a_4a_5=0$ contradicts assumption (i), $A_2$ cannot vanish, and the last factor implies 
the collinearity of $\go M_3,\go M_4,\go M_5$, which contradicts assumption (ii). 
Therefore we can assume for the remaining discussion that $a_2\neq 0$ holds. We distinguish the following 
cases:
\begin{enumerate}[1.]
\item
$a_5\neq 0$: Under this assumption we can solve $D_2=0$ for $A_5$. Then the numerator of $D_3$ factors into 
\begin{equation}\label{help1}
a_3a_4(A_2B_3 - A_2B_4 + A_3B_4 - A_4B_3)F[20],
\end{equation}
where the number in the brackets gives the number of terms.
As $F$ is also a factor of $D_1$ and $D_4$, the condition 
$F=0$ implies an architecturally singular design. Moreover $a_i=0$ ($\Leftrightarrow$ $\go m_1=\go m_i$) implies the collinearity of $\go M_2,\go M_j,\go M_5$ for 
pairwise distinct $i,j\in\left\{3,4\right\}$. Therefore $a_3a_4=0$ contradicts assumption (ii). The vanishing of the third factor 
of Eq.\ (\ref{help1}) implies the collinearity of $\go M_2,\ldots ,\go M_5$, which contradicts assumption (iii). 
\item
$a_5=0$ ($\Leftrightarrow$ $\go m_1=\go m_5$): Now $D_2$ and $D_3$ factors into: 
\begin{equation}
-a_2a_3a_4B_5(A_2B_3 - A_2B_4 + A_3B_4 - A_4B_3),\qquad
a_2a_3a_4A_5(A_2B_3 - A_2B_4 + A_3B_4 - A_4B_3).
\end{equation}
$a_2a_3a_4=0$ contradicts assumption (i) and the last factor implies 
the collinearity of $\go M_2,\go M_3,\go M_4$, which contradicts assumption (ii). 
Therefore $A_5=B_5=0$ has to hold, but in this case the first and the fifth leg coincide. 
This closes the proof of Lemma \ref{d2d3}. \hfill $\BewEnde$
\end{enumerate}

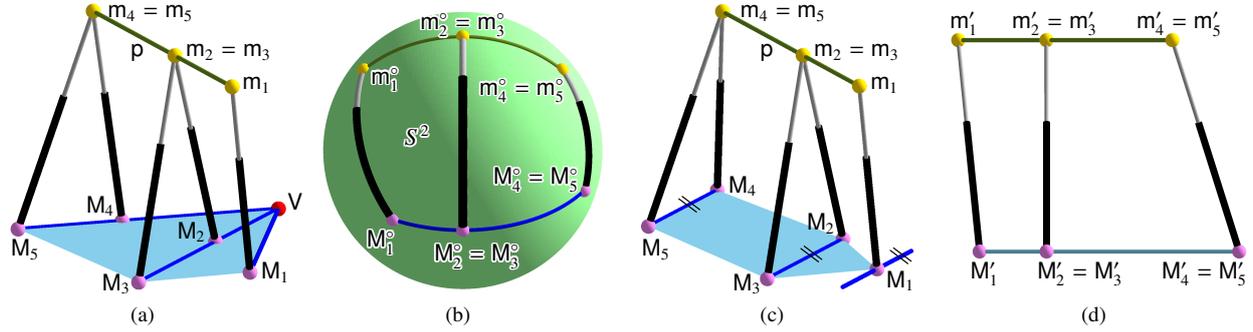
\begin{figure}[top]
\centering
\subfigure[]{ 
 \begin{overpic}
    [width=36.5mm]{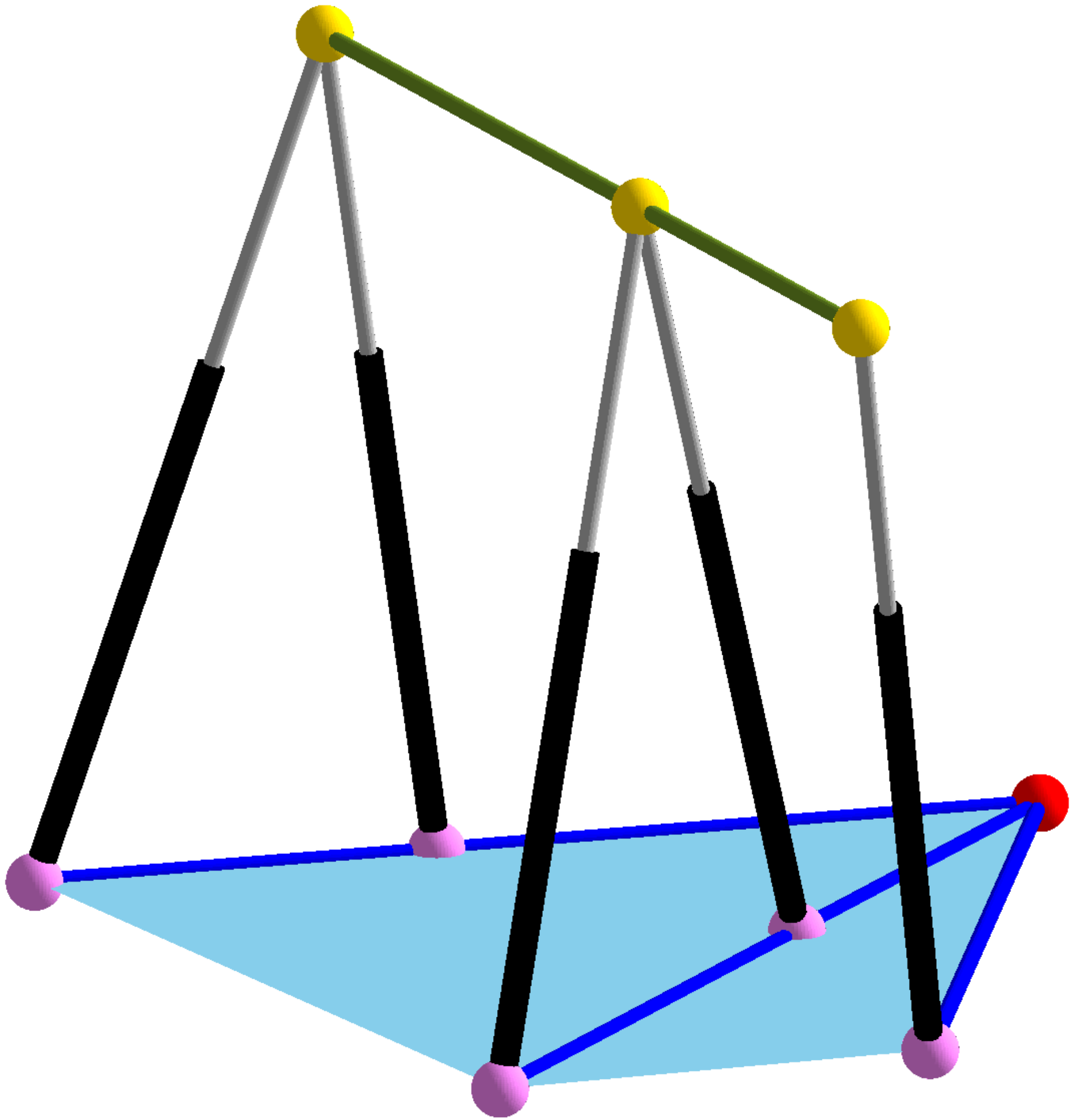}
\begin{small}
\put(32,0){$\go M_3$}
\put(87,4){$\go M_1$}
\put(58,18){$\go M_2$}
\put(96,28){$\go V$}
\put(0.5,12){$\go M_5$}
\put(26.5,27){$\go M_4$}
\put(34,95){$\go m_4=\go m_5$}
\put(42,81.5){$\go p$}
\put(61,81.5){$\go m_2=\go m_3$}
\put(80.5,70){$\go m_1$}
\end{small} 
  \end{overpic} 
}
  \hfill
\subfigure[]{  
  \begin{overpic}
    [width=37mm]{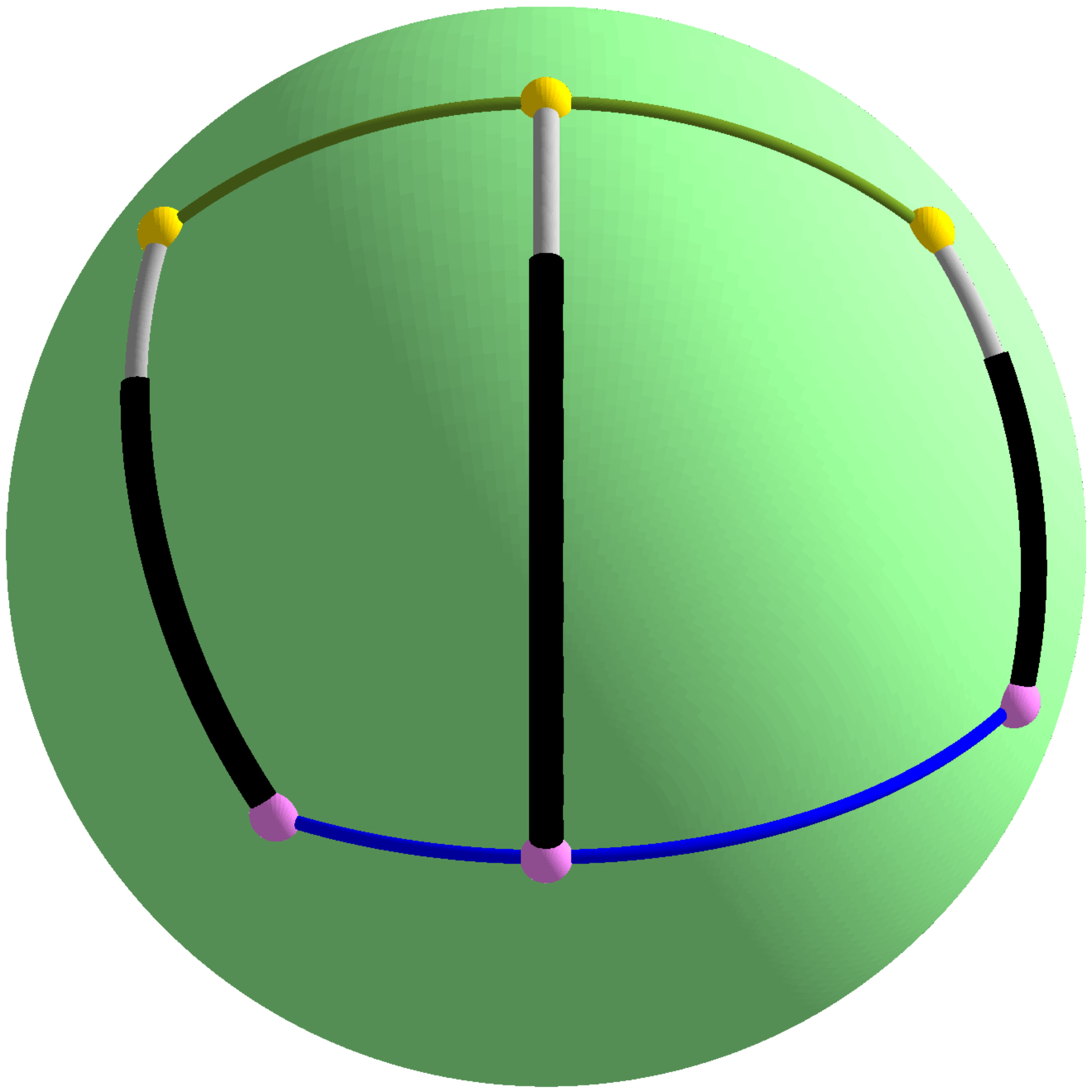}
  \contourlength{0.2mm}
  \begin{small} 
    \put(16,15.5){\contour{white}{$\go M_1^{\circ}$}}    
    \put(40,10.5){\contour{white}{$\go M_2^{\circ}=\go M_3^{\circ}$}}
    \put(61,38){\contour{white}{$\go M_4^{\circ}=\go M_5^{\circ}$}}
		\put(17.5,74){\contour{white}{$\go m_1^{\circ}$}}
		\put(35.0,93.5){\contour{white}{$\go m_2^{\circ}=\go m_3^{\circ}$}}
		\put(56,70){\contour{white}{$\go m_4^{\circ}=\go m_5^{\circ}$}}
    \put(29.5,52){\contour{white}{$S^2$}}
	\end{small}
  \end{overpic}
  }
\hfill
 \subfigure[]{  
  \begin{overpic}
    [width=36mm]{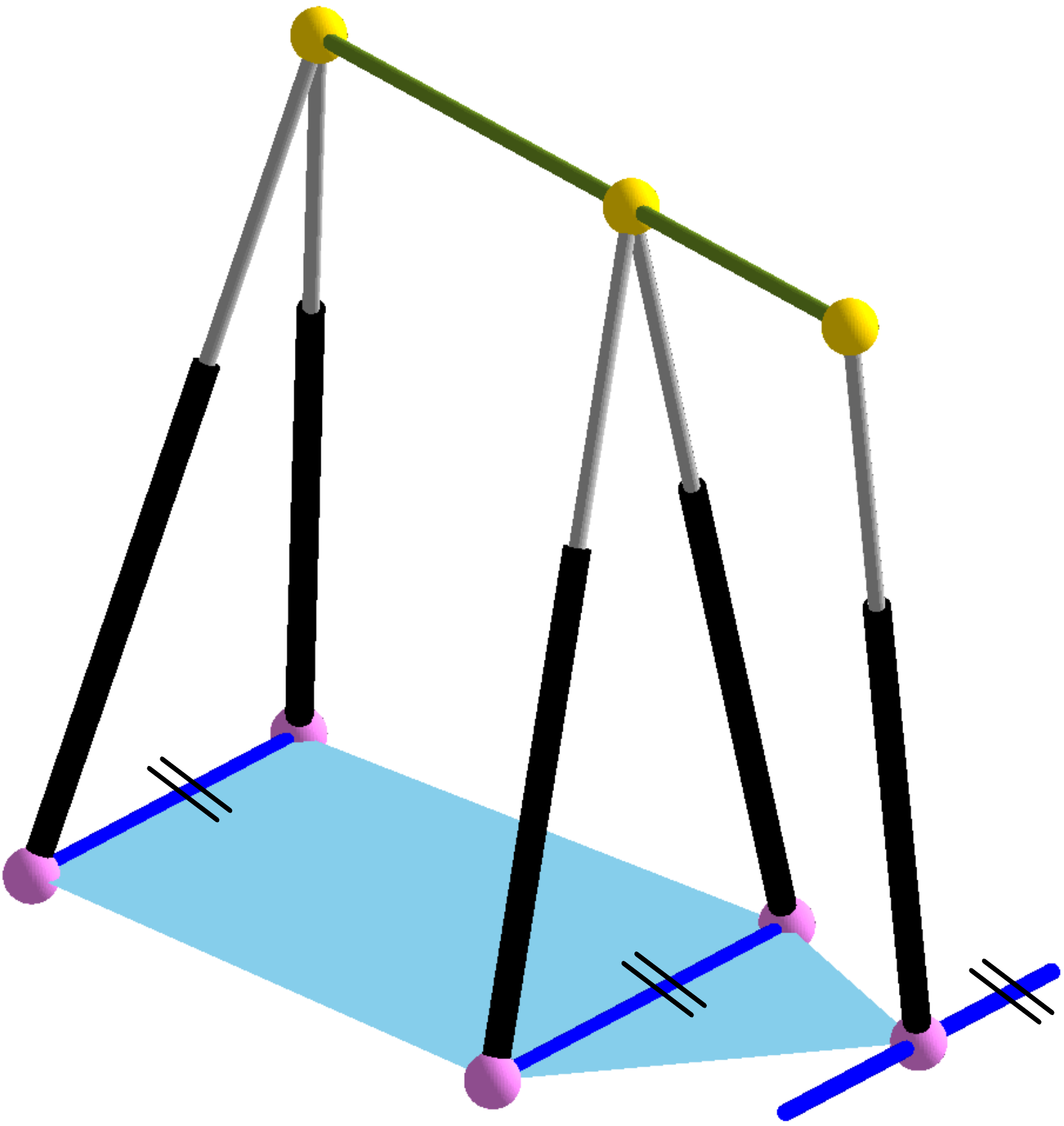}
  \contourlength{0.2mm}
  \begin{small} 
\put(32,0){$\go M_3$}
\put(85,1){$\go M_1$}
\put(57.5,21){$\go M_2$}
\put(0.5,12){$\go M_5$}
\put(30.5,34){$\go M_4$}
\put(34.5,95){$\go m_4=\go m_5$}
\put(41.5,81.5){$\go p$}
\put(60,81.5){$\go m_2=\go m_3$}
\put(79.5,70){$\go m_1$}
	\end{small}
  \end{overpic}
  }
\hfill
\subfigure[]{  
  \begin{overpic}
    [width=39mm]{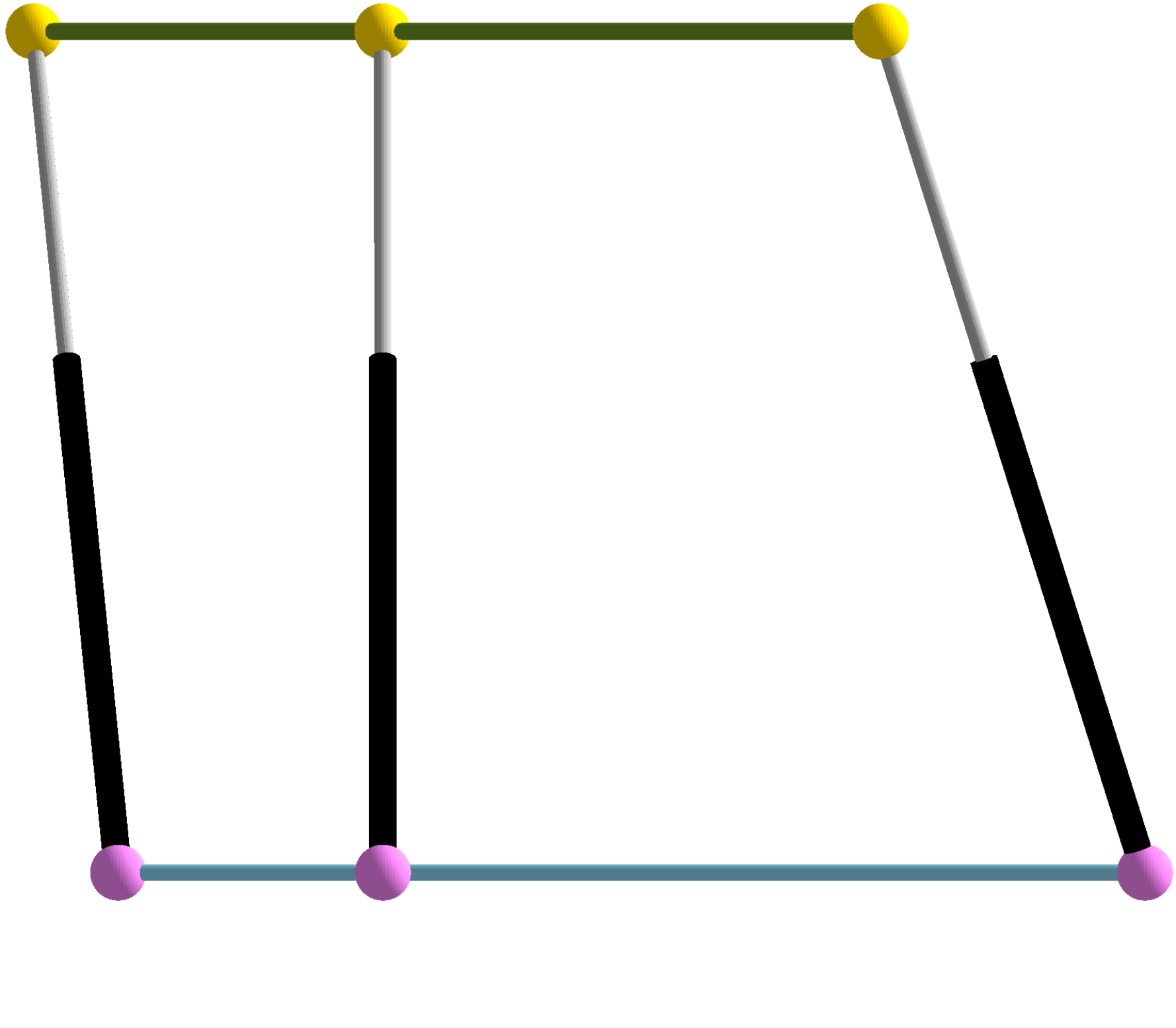}
  \contourlength{0.2mm}
  \begin{small} 
\put(71.5,4){$\go M_4^{\prime}=\go M_5^{\prime}$}
\put(29.5,4){$\go M_2^{\prime}=\go M_3^{\prime}$}
\put(7,4){$\go M_1^{\prime}$}
\put(63,89){$\go m_4^{\prime}=\go m_5^{\prime}$}
\put(21,89){$\go m_2^{\prime}=\go m_3^{\prime}$}
\put(0,89){$\go m_1^{\prime}$}
	\end{small}
  \end{overpic} 
  }
\caption{(a) $\go V$ is a finite point. (b) Spherical image of the pentapod given in (a) with respect to the unit sphere $S^2$ centered in $\go V$. 
(c) $\go V$ is an ideal point. (d) Orthogonal projection of the pentapod given in (c) onto a plane orthogonal to $\go V$.}
  \label{fig1}
\end{figure}

\noindent
Due to this lemma Eq.\ (\ref{centraleq}) determines for all planar pentapods of $\mathcal{P}$ 
a bijection between points of $\go p$ and lines in the base plane.\footnote{For $D_2=D_3=0$ 
Eq.\ (\ref{centraleq}) would factor into $a(D_4A+D_5B+D_1)=0$, which does not imply such a bijection.}
Due to the linear relation the lines generate a pencil with vertex $\go V$, which can also be an ideal point ($\Rightarrow$ parallel line pencil). 
According to \cite{borras2} we are now able to perform a series of leg-replacements in a way that we end up with 
a non-architecturally singular pentapod of the following type (see Fig.\ \ref{fig1}a,c): 
\begin{equation}\label{nach_legre}
\go m_2=\go m_3,\qquad
\go m_4=\go m_5,\qquad
\go M_2,\go M_3,\go V\,\, \text{are collinear}, \qquad
\go M_4,\go M_5,\go V\,\, \text{are collinear}.
\end{equation}

\noindent
Based on this preparatory work we can prove Theorem \ref{thm:planar} as follows:

\noindent
{\sc Proof of Theorem \ref{thm:planar}:} 
We distinguish the following two cases: 
\begin{enumerate}[1.]
\item
\underline{$\go V$ is a finite point:} 
In this case the motion of $\go p$ can only be spherical with center $\go V$. Therefore we can also consider the 
spherical 3-legged manipulator, which we obtain by projecting the pentapod onto the unit sphere $S^2$ centered in $\go V$. 
Therefore the spherical 3-legged manipulator has to have a self-motion. 

Note that the projected base points $\go M_i^{\circ}$ as well as the  projected platform points $\go m_i^{\circ}$ 
are located on great circles of $S^2$. It is well known (cf.\ \cite[Lemma 2]{nawratil_proj} and \cite[Theorem 5]{nawratil_bond}) that
the 3-legged spherical manipulators illustrated in (see Fig.\ \ref{fig1}b) can only have self-motions if two 
platform or base anchor points coincide. 
Now this is only possible if the line $\go p$ contains $\go V$. In this case the platform of the spherical manipulator collapse into a 
point and we only get the trivial rotation about the line $\go p$ as uncontrollable motion while $\go p$ itself remains fix. 
Therefore this case does not yield a solution. 
\item
\underline{$\go V$ is an ideal point:}  
Now the motion of $\go p$ can only be a planar one orthogonal to the direction of $\go V$. 
Therefore the corresponding planar 3-legged manipulator, which is obtained by an orthogonal projection of the pentapod onto 
a plane orthogonal to $\go V$, also has to have a self-motion. Note that the projected base points $\go M_i^{\prime}$ 
as well as the  projected platform points $\go m_i^{\prime}$ are collinear (see Fig.\ \ref{fig1}d).
According to \cite{nawratil_proj,nawratil_bond} this planar 3-legged manipulator 
can only have a self-motion in one of the following two cases:
\begin{enumerate}[$(a)$]
\item
Two platform or base anchor points coincide: This is only possible if the line $\go p$ contains $\go V$. 
Analogous considerations as in the spherical case show that we do not get a solution.
\item
The platform and the base are congruent and all legs have equal lengths: In this case the 
planar 3-legged manipulator has a circular translation. This already implies the solution given 
in Theorem \ref{thm:planar}. \hfill $\BewEnde$
\end{enumerate}
\end{enumerate}

\begin{rmk}\label{rmk:bond}
Note that the case given in Theorem \ref{thm:planar} is also not covered by the more than 100 year old results of 
Darboux, Mannheim, Duporcq and Bricard, respectively, even though no platform anchor points have to coincide. 
Therefore our study reveals a further lost case beside design ($\gamma$).  

But this case is not novel, as it is already contained within the more general 
characterization given in Theorem 2\footnote{This theorem is originally stated for hexapods but 
it also holds for pentapods, as its proof is also valid for $5$-legged manipulators.}
of \cite{nawratil_bond}, which reads as follows: {\it A pentapod has a translational self-motion 
if and only if the platform can be rotated about the center $\go m_1=\go M_1$ into a pose, 
where the vectors $\overrightarrow{\go M_i\go m_i}$ for $i=2,\ldots ,5$ 
fulfill the condition  $rk(\overrightarrow{\go M_2\go m_2},\ldots ,\overrightarrow{\go M_5\go m_5})\leq 1$.} 
Therefore a pentapod with a linear platform and a translational self-motion has to have a planar base.
\hfill $\diamond$
\end{rmk}


\section{Types of non-planar pentapods}\label{p5notcoplanar}

Based on the idea of singular-invariant leg-replacements, which can also be extended to the non-planar case (cf.\ \cite{borras3}),  
one can distinguish different types introduced in this section.

\begin{lem}\label{labeling}
The anchor points of a non-planar pentapod with a linear platform 
can always be relabeled in a way that the following conditions hold: 
\begin{enumerate}
\item
$\go M_1,\ldots ,\go M_4$ span a tetrahedron.
\item
If $\go M_1,\ldots ,\go M_5$ are pairwise distinct, then $\go M_1,\go M_i,\go M_5$ are not collinear for all $i\in\left\{2,3,4\right\}$. \\
If two base anchor points coincide, then $\go M_1$ is none of them. 
\end{enumerate} 
\end{lem}

As the proof is trivial it is left to the reader and we proceed with the coordinatization used within this section: 
Again we choose the origin of the moving system $\Sigma$ in $\go m_1$. 
The fixed frame $\Sigma_0$ is selected in a way that $\go M_1$ 
equals the origin, $\go M_2$ belongs to the $x$-axis and $\go M_3$ is located in the $xy$-plane. 
All in all this yields $A_2B_3C_4\neq 0$ and $\Vkt M_5\neq \Vkt o$ (zero vector). 

Moreover we introduce the following notation: $D_{ijk}$ denotes the determinant of the $4\times 7$ matrix 
\begin{equation}\label{matrix_cubic}
(\Vkt a,\Vkt A, \Vkt B, \Vkt C, \Vkt a\Vkt A, \Vkt a\Vkt B, \Vkt a\Vkt C) \quad \text{with} \quad
\Vkt C=(0,0,C_4,C_5)^T, \quad \Vkt a\Vkt C=(0,0,a_4C_4,a_5C_5)^T
\end{equation}
after removing the $i$-th, $j$-th and $k$-th column. Now we can state the following lemma:

\begin{lem}\label{noch}
For a non-planar pentapod of $\mathcal{P}$ 
the condition $D_{167}=D_{157}=D_{156}=0$ cannot hold (with respect to $\Sigma$ and $\Sigma_0$).
\end{lem}

\noindent
{\sc Proof:}
As $\Vkt M_2,\Vkt M_3,\Vkt M_4$ are linearly independent there exist a unique triple $(\lambda_2,\lambda_3,\lambda_4)$ with
$\lambda_2\Vkt M_2+\lambda_3\Vkt M_3+\lambda_4\Vkt M_4=\Vkt M_5$. Due to $\Vkt M_5\neq \Vkt o$ we have $(\lambda_2,\lambda_3,\lambda_4)\neq (0,0,0)$. 
Now it can easily be seen that the following equivalences hold: 
\begin{equation}
D_{167}=0\Leftrightarrow \sum_{i=2}^4 \lambda_ia_iA_i=a_5A_5, \quad
D_{157}=0\Leftrightarrow \sum_{i=2}^4 \lambda_ia_iB_i=a_5B_5, \quad
D_{156}=0\Leftrightarrow \sum_{i=2}^4 \lambda_ia_iC_i=a_5C_5. 
\end{equation}
Therefore $D_{167}=D_{157}=D_{156}=0$ implies $\lambda_2a_2\Vkt M_2+\lambda_3a_3\Vkt M_3+\lambda_4a_4\Vkt M_4=a_5\Vkt M_5$. 
As $a_5$ cannot equal zero\footnote{For $a_5=0$ we get $\go m_1=\go m_5$ and therefore $a_2a_3a_4\neq 0$ has to hold, as 
otherwise we get a contradiction to assumption (i).} we can divide both sides by $a_5$, which shows that the following relation has to hold: 
\begin{equation}
(\lambda_2,\lambda_3,\lambda_4)=
\left(
\frac{a_2}{a_5}\lambda_2,\frac{a_3}{a_5}\lambda_3,\frac{a_4}{a_5}\lambda_4
\right).
\end{equation}
In order to get no contradiction with assumption (i) and $\Vkt M_5\neq \Vkt o$ the implied three equations only have the following solution: 
$a_i=a_5$ and $\lambda_j=\lambda_k=0$ with pairwise distinct $i,j,k\in\left\{2,3,4\right\}$. 
For $\lambda_i\neq 1$ the points $\go M_1,\go M_i,\go M_5$ are collinear, which contradicts Lemma \ref{labeling}
and for $\lambda_i= 1$ the $i$-th leg and the fifth leg coincide; a contradiction. \hfill $\BewEnde$\\

\noindent
In the following we distinguish two cases with respect to the criterion 	whether 
$D_{567}$ vanishes or not. This subdivision was also used by Bricard in \cite[Items 12 and 13 of Chapter III]{bricard}  
as $D_{567}=0$ is equivalent to the following geometric condition, which we call the affine relation (AR):

\begin{itemize}
\item[(AR)]
There exists a singular affinity $\kappa$ with $\go M_i\mapsto \go m_i$ for $i=1,\ldots ,5$.
\end{itemize}

\subsection{$D_{567}\neq 0$}

Under this assumption we can use the following result of Borras and Thomas \cite{borras3}:
{\it A leg of a given non-planar pentapod with linear platform is replaced by a leg with platform anchor point $(a,0,0)$ and base anchor point $(A,B,C)$ 
fulfilling  Eq.\ (7) of \cite{borras3}, which reads as follows within our notation:
\begin{equation}\label{centraleq_nonplanar}
\begin{pmatrix}
D_{267}-aD_{567} & -D_{367} & D_{467} \\
D_{257} & -D_{357}-aD_{567} & D_{457} \\
D_{256} & -D_{356} & D_{456}-aD_{567}
\end{pmatrix}
\begin{pmatrix}
A \\ B \\ C
\end{pmatrix}
=
a
\begin{pmatrix}
D_{167} \\ D_{157} \\D_{156} 
\end{pmatrix}
\end{equation}
then the resulting pentapod has the same singularity set (and direct kinematics solution) 
if it is not architecturally singular.}\\

\noindent
Solving Eq.\ (\ref{centraleq_nonplanar}) with Crammer's rule yields: 
\begin{equation}\label{cramer}
A=\frac{d_1(a)}{d_0(a)}, \quad
B=\frac{d_2(a)}{d_0(a)}, \quad
C=\frac{d_3(a)}{d_0(a)}.
\end{equation}
Due to the assumption $D_{567}\neq 0$, the polynomial $d_0$ is cubic in the unknown $a$. The other 
polynomials $d_1,d_2,d_3$ are of degree 3 or less in $a$, but 
due to Lemma \ref{noch} one of them has to be cubic, which shows the following result: 

\begin{cor}\label{csc}
The locus of base anchor points of singular-invariant leg-replacements of a non-planar pentapod of $\mathcal{P}$, 
which does not fulfill the affine relation (AR), is a cubic space curve.
\end{cor} 

\begin{figure}[top]
\begin{center} 
\subfigure[]{ 
 \begin{overpic}
    [width=47mm]{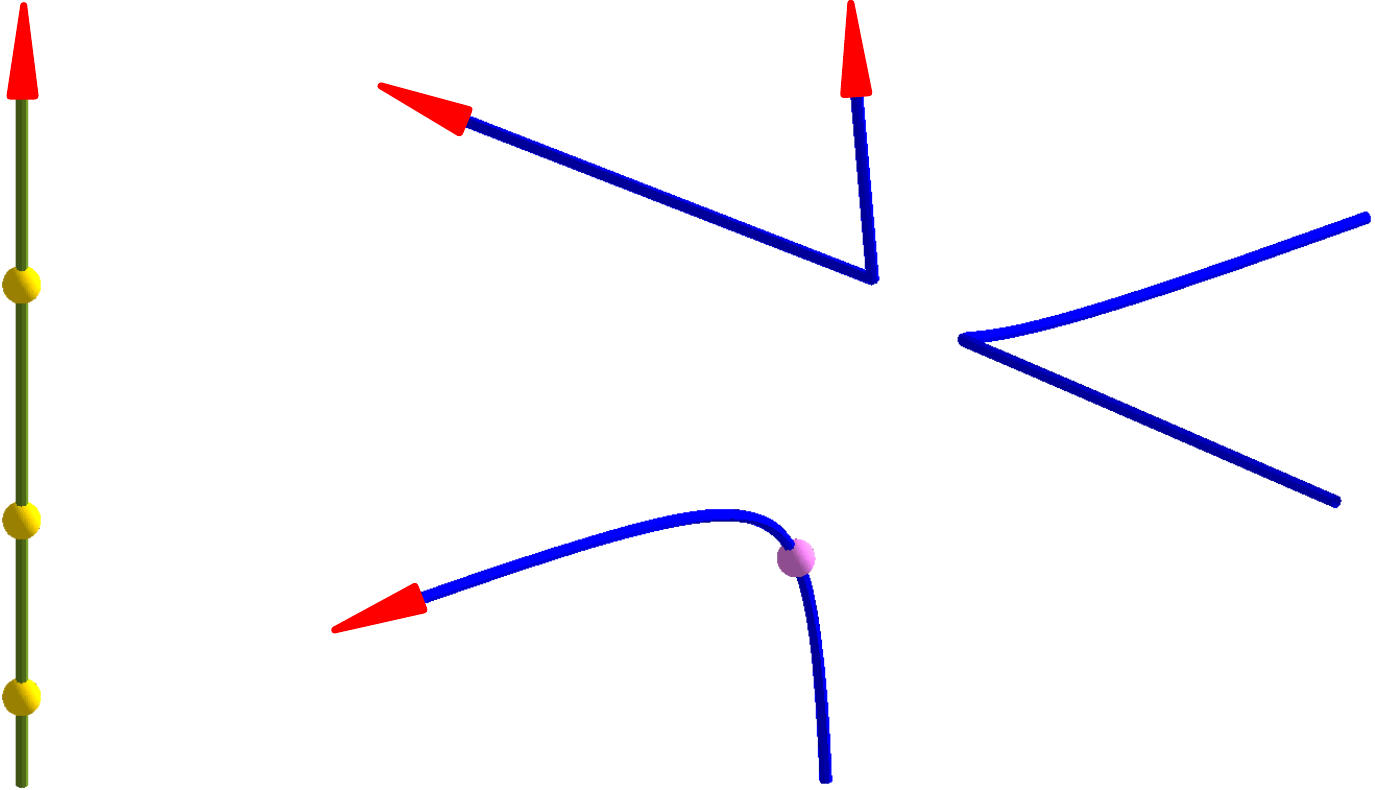}
\begin{small}
\put(3,0){$\go W_3\sigma^{-1}$}
\put(3,19){$\go W_2\sigma^{-1}$}
\put(3,30){$\go W_1\sigma^{-1}$}
\put(3,43){$\go p$}
\put(4,52){$\go U$}
\put(30,52){$\go W_1$}
\put(64,52){$\go W_2$}
\put(28,6){$\go W_3$}
\put(60.5,14){$\go U\sigma$}
\put(94,15){$\go s$}
\end{small}     
  \end{overpic} 
 }
\qquad\qquad
\subfigure[]{
\begin{overpic}
    [width=50mm]{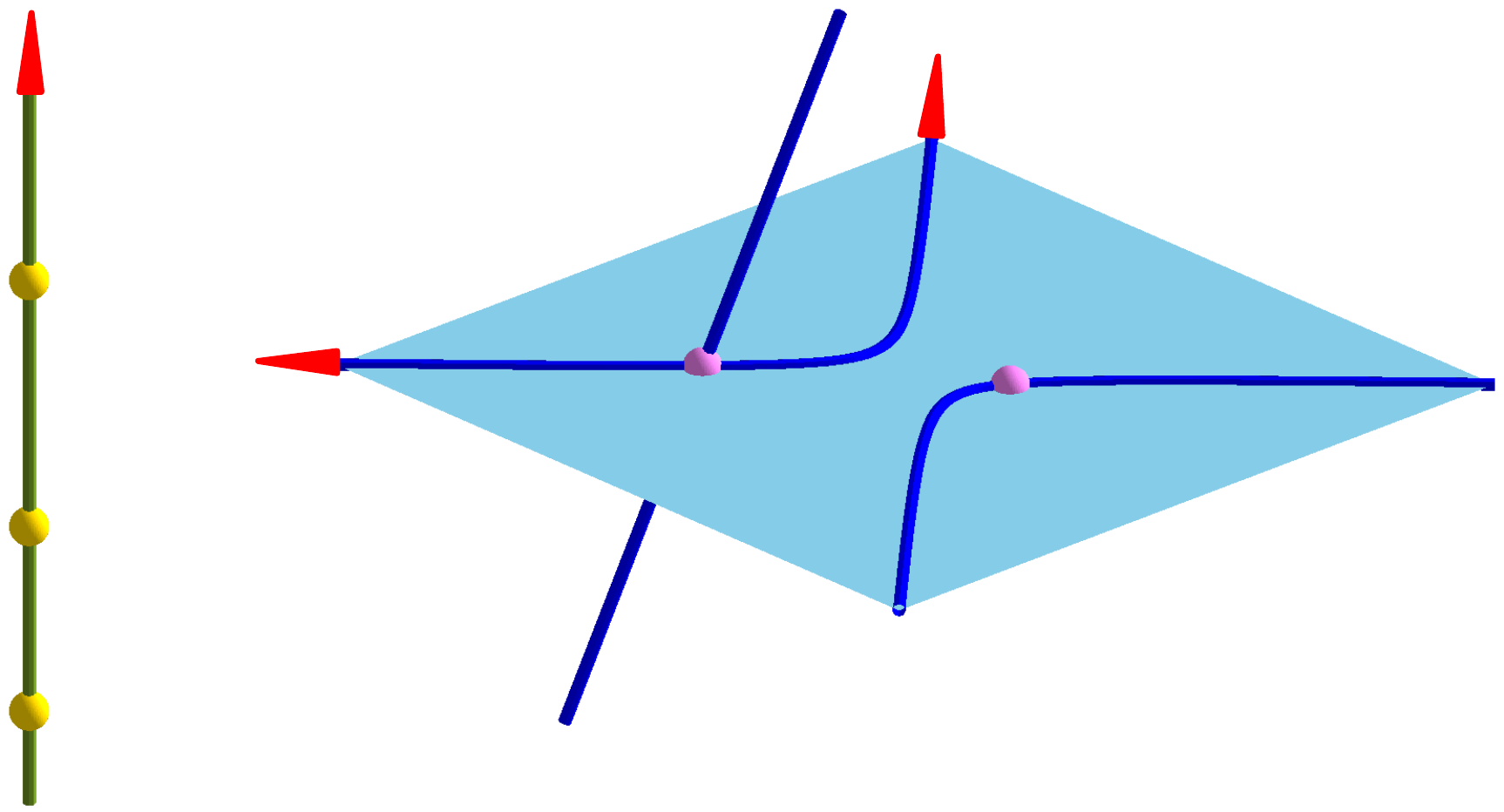}
\begin{small}
\put(3.5,0){$\go W_2\sigma^{-1}$}
\put(3.5,12){$\go P_1$}
\put(3.5,36){$\go W_1\sigma^{-1}$}
\put(3.5,26){$\go p$}
\put(4,48.5){$\go U$}
\put(31,6){$\go g_1$}
\put(55.5,16.5){$\go q$}
\put(65,22){$\go U\sigma$}
\put(17,23){$\go W_1$}
\put(64,46){$\go W_2$}
\put(44.5,23.3){$\go Q$}
\put(72,35){$\varepsilon$}
\end{small}         
  \end{overpic}
} 
\\
\subfigure[]{
\begin{overpic}
    [width=45mm]{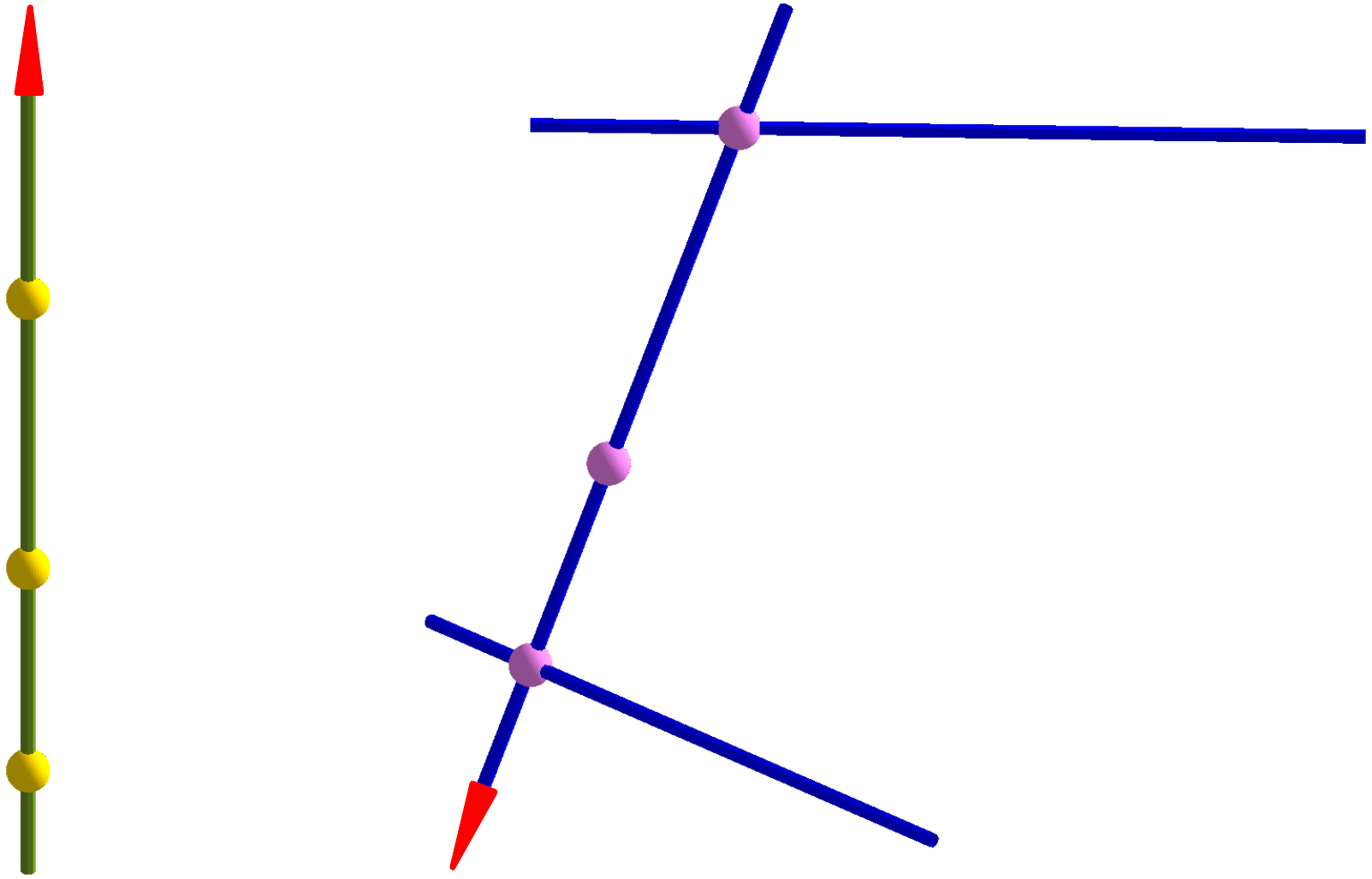}
\begin{small}
\put(5,5){$\go P_2$}
\put(5,19){$\go W_1\sigma^{-1}$}
\put(5,41){$\go P_1$}
\put(4,31){$\go p$}
\put(4.5,58){$\go U$}
\put(47,57.5){$\go L_1$}
\put(93.5,48){$\go g_1$}
\put(42.5,16){$\go L_2$}
\put(38,1){$\go W_1$}
\put(47,26){$\go U\sigma$}
\put(70,4){$\go g_2$}
\put(51.5,39){$\go l$}
\end{small} 
  \end{overpic}
}  
\qquad\qquad\qquad
\subfigure[]{
\begin{overpic}
    [width=44mm]{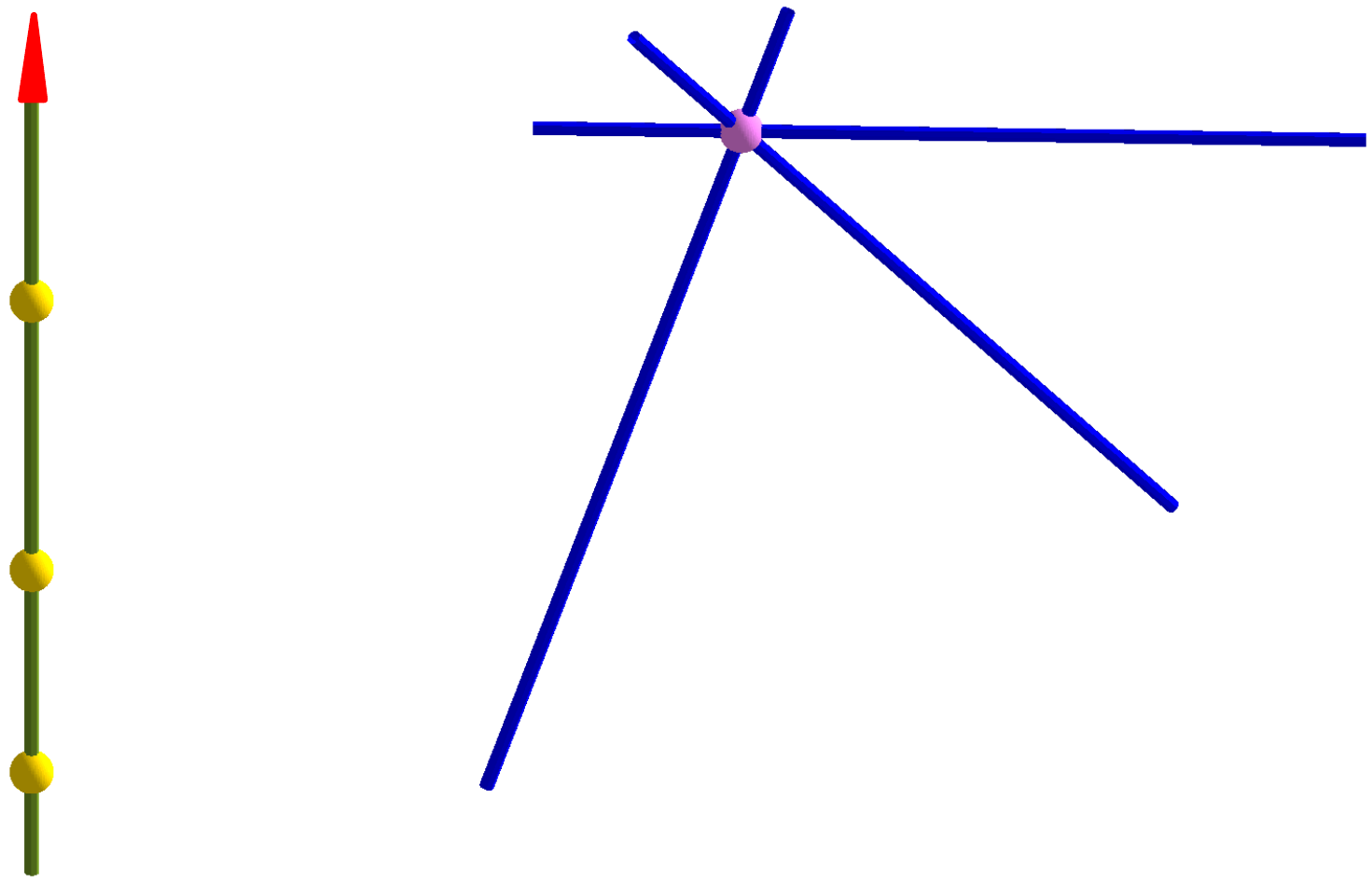}
\begin{small}
\put(5,5){$\go P_3$}
\put(5,19){$\go P_2$}
\put(5,41){$\go P_1$}
\put(39,8.5){$\go g_3$}
\put(75.5,25.5){$\go g_2$}
\put(4,31){$\go p$}
\put(4.5,58){$\go U$}
\put(58.5,56.5){$\go V=\go U\sigma$}
\put(93.5,48){$\go g_1$}
\end{small} 
  \end{overpic}
}  
\end{center} 
\caption{Sketch of the types of non-planar pentapods of $\mathcal{P}$, which do not fulfill the affine relation: (a) Type 1. (b) Type 2. (c) Type 3. (d) Type 4. 
Pentapods fulfilling the affine relation belong to Type 5.
}
  \label{fig25}
\end{figure}
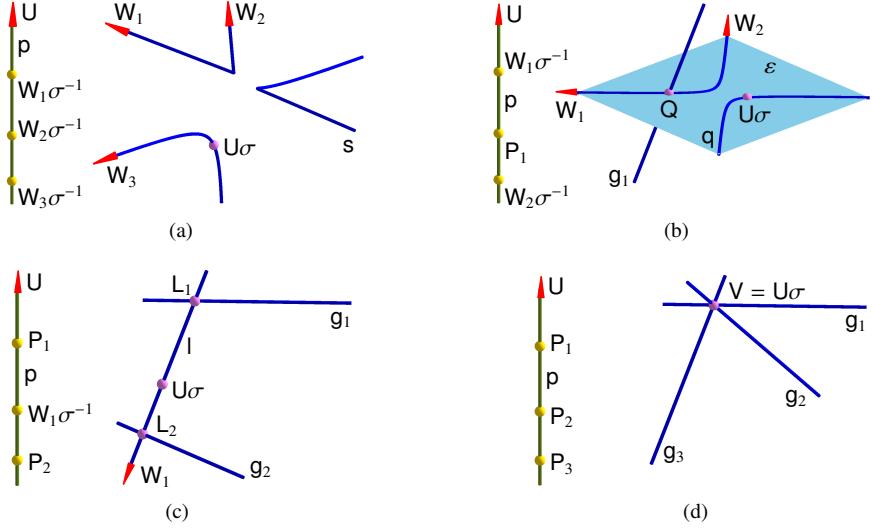

According to Borras and Thomas \cite{borras3} we can distinguish different types  
with respect to the number of roots of $d_0=0$, for which the system Eq.\ (\ref{centraleq_nonplanar}) is consistent. 
This yields the following classification:

\begin{thm}
A non-planar pentapod of $\mathcal{P}$, which does not fulfill the affine relation (AR), belongs to one of the following four types: The 
cubic of Corollary \ref{csc}: 
\begin{enumerate}[{Type} 1]
\item
is irreducible: There is a bijection $\sigma$ between $\go p$ and this space curve $\go s$ (see Fig.\ \ref{fig25}a).
\item
splits up into an irreducible conic $\go q$, located in the finite plane $\varepsilon$ and a finite line $\go g_1\notin \varepsilon$, 
which intersects $\go q$ in the point $\go Q$: 
There is a bijection $\sigma$ between $\go p\setminus\left\{\go P_1\right\}$ and  $\go q\setminus\left\{\go Q\right\}$. 
Moreover the finite point $\go P_1$ is mapped to $\go g_1$ (see Fig.\ \ref{fig25}b). 
\item
splits up into the finite lines $\go l$ and the finite skew lines $\go g_1,\go g_2$, which intersects $\go l$ in the point $\go L_1$ and $\go L_2$, respectively: 
There is a bijection $\sigma$ between $\go p\setminus\left\{\go P_1,\go P_2\right\}$ and  $\go l\setminus\left\{\go L_1,\go L_2\right\}$.
Moreover the finite point $\go P_i$ is mapped to $\go g_i$ for $i=1,2$ (see Fig.\ \ref{fig25}c).
\item
splits up into the finite lines $\go g_1,\go g_2,\go g_3$, which are not coplanar but intersect each other in the finite point $\go V$: 
All points of $\go p\setminus\left\{\go P_1,\go P_2,\go P_3\right\}$ are mapped to $\go V$. 
Moreover the finite point $\go P_i$ is mapped to  $\go g_i$ for $i=1,2,3$ (see Fig.\ \ref{fig25}d). 
\end{enumerate}
\end{thm}

Note that in Type $i$ we have  $4-i$ points $\go W_1\sigma^{-1},\ldots , \go W_{4-i}\sigma^{-1}$  on $\go p$ (counted with algebraic multiplicity) 
for $i=1,\ldots, 4$, which are mapped by $\sigma$ to ideal points $\go W_1,\ldots , \go W_{4-i}$ of the base. 
These points of $\go p$ have the special property that their trajectory is in a plane orthogonal to the respective ideal point. 
Therefore each of these point pairs determines a so-called Darboux condition (cf.\ \cite[Item 6 of Chapter II] {bricard} and \cite[Section 4.1]{nawratil_duporcq}). 
Any other finite point of $\go p$ determines a so-called sphere condition; i.e.\ it is located  on a sphere centered in the corresponding finite 
base anchor point. 

The points $\go P_i$ have the special property of possessing circular trajectories, i.e.\ their path is planar and spherical at the same time. 

Note that the ideal point $\go U$ of $\go p$ is in all four cases mapped by $\sigma$ onto a finite point $\go U\sigma$. 
Therefore this point pair determines a so-called Mannheim condition, which is the inverse of the 
Darboux condition; i.e.\ a plane of the moving system orthogonal to $\go p$ slides through a finite point of the base.

\begin{rmk}
The above given correspondence between points on $\go p$ and points on the base can also be seen as the
correspondence of point paths of $\go p$ and the centers of their osculating spheres. It is an old result of 
Sch\"onflies \cite{schoenflies} that this correspondence is cubic. \hfill $\diamond$ 
\end{rmk}

\subsection{$D_{567}= 0$}\label{ref:AR}

\begin{thm}
For a non-planar pentapod of $\mathcal{P}$, which fulfills the affine relation (AR),  
the ideal point $\go U$ of $\go p$ is related to an ideal element of the base 
within the correspondence implied by singular-invariant leg-replacements. 
Pentapods with this property belong to Type 5.
\end{thm}

\noindent
{\sc Proof:} 
Due to Lemma \ref{noch} not all three determinants $D_{156}$, $D_{157}$, $D_{167}$ can be equal to zero.  
Therefore we can assume e.g.\ that $D_{167}\neq 0$\footnote{For $D_{156}\neq 0$ or $D_{157}\neq 0$ the argumentation can be done 
analogously.} holds. Then similar considerations to those given in 
\cite{borras3} yield the following correspondence: 
\begin{equation}\label{centraleq_nonplanar_spezi}
\begin{pmatrix}
D_{267} & -D_{367} &D_{467} \\
D_{126}-aD_{156} & -D_{136} & D_{146}+aD_{167} \\
D_{127}-aD_{157} & aD_{167}-D_{137} & D_{147}
\end{pmatrix}
\begin{pmatrix}
A \\ B \\ C
\end{pmatrix}
=
a
\begin{pmatrix}
D_{167} \\ 0 \\0
\end{pmatrix}.
\end{equation}
Introducing homogeneous coordinates to the first of these three equations show that the ideal point $\go U$ of $\go p$ is mapped 
to an ideal element (ideal point, ideal line or the complete ideal plane) of the base. \hfill $\BewEnde$ 

\begin{lem}\label{prop5}
If a pentapod of Type 5 has a self-motion, then the ideal element has to be a point $\go W$. 
Moreover the self-motion is a Sch\"onflies motion with axis direction $\go W$. 
\end{lem}

\noindent
{\sc Proof:}
We prove by contradiction that  $\go U$ cannot be mapped to more than one ideal point $\go W$ of the base: 
Assume that $\go U$ can be connected with two distinct ideal points $\go W_1$ and $\go W_2$ of the base. 
Then these special two "legs" correspond to two angle conditions (cf.\ \cite[Item 7 of Chapter II]{bricard} and 
\cite[Section 4.1]{nawratil_duporcq}); i.e.\ the angle enclosed by $\go U$ and $\go W_i$ ($i=1,2$) has to be constant during the self-motion. 
As this already fixes the orientation of $\go p$, the pentapod can only have a translational self-motion, which implies 
the coplanarity of the base (cf.\ Remark \ref{rmk:bond}); a contradiction. 

As the angle enclosed by $\go U$ and $\go W$ has to be constant, the self-motion of $\go p$ can only 
be a Sch\"onflies motion with axis direction $\go W$. \hfill $\BewEnde$ \\

\noindent
In the following we analyze the Types 1-5 separately with respect to the existence of self-motions.  
Type 4 can be discussed similar to the planar case, as singular-invariant leg-replacements can be used to get 
the same pentapod illustrated in Fig.\ \ref{fig1}a with the sole difference that the three lines through the finite point $\go V$ are not coplanar. 
Under consideration of \cite[Theorem 6]{nawratil_bond} an analogous argumentation as in item 1 of the proof of Theorem \ref{thm:planar} 
can be done for Type 4, which shows the following result: 

\begin{thm}\label{thm:type4}
A pentapod of Type 4 cannot possess a self-motion.
\end{thm}

For the discussion of the remaining four types we apply the theory of bonds for pentapods with linear platform, 
which is the content of the next section.


\section{Bond theory for pentapods with linear platform}\label{sec:dkp}

Readers, who are not familiar with the following terms of algebraic geometry (ideal, variety, Gr\"obner base, radical), 
are recommended to look up these basics in the articles \cite{hustyetal} and \cite{hustyschrocker}, before studying this section. 

It was shown by Husty \cite{husty} that the sphere condition equals a homogeneous quadratic equations in the Study parameters 
$(e_0:e_1:e_2:e_3:f_0:f_1:f_2:f_3)$. For our choice of the moving frame $\Sigma$ the sphere condition $\Lambda_i$ simplifies to: 
\begin{equation}\label{eq:lambda}
\begin{split}
\Lambda_i:\quad 
&(a_i^2+A_i^2+B_i^2+C_i^2-R_i^2)(e_0^2+e_1^2+e_2^2+e_3^2) 
-2a_iA_i(e_0^2+e_1^2-e_2^2-e_3^2)
-4a_iB_i(e_0e_3+e_1e_2)\\
&+4a_iC_i(e_0e_2-e_1e_3)
-4a_i(e_0f_1-e_1f_0-e_2f_3+e_3f_2)
+4A_i(e_0f_1-e_1f_0+e_2f_3-e_3f_2)\\
&+4B_i(e_0f_2-e_1f_3-e_2f_0+e_3f_1)
+4C_i(e_0f_3+e_1f_2-e_2f_1-e_3f_0)
+4(f_0^2+f_1^2+f_2^2+f_3^2) =0,
\end{split}
\end{equation}
where $R_i$ denotes the radius of the sphere centered in $\go M_i$ on which $\go m_i$ is located.

Now, all real points of the Study parameter space $P^7$ (7-dimensional projective space), 
which are located on the so-called Study quadric $\Psi:\,\sum_{i=0}^3e_if_i=0$, 
correspond to an Euclidean displacement, with exception of the 3-dimensional subspace $e_0=e_1=e_2=e_3=0$, 
as its points cannot fulfill the condition $e_0^2+e_1^2+e_2^2+e_3^2\neq 0$. 
The translation vector $\Vkt t:=(t_1,t_2,t_3)^T$ and the rotation matrix $\Vkt R$ of the corresponding 
Euclidean displacement $\Vkt m_i\mapsto\Vkt R\Vkt m_i + \Vkt t$ are given by:  
\begin{equation*}
t_1=-2(e_0f_1-e_1f_0+e_2f_3-e_3f_2), \quad
t_2=-2(e_0f_2-e_2f_0+e_3f_1-e_1f_3), \quad
t_3=-2(e_0f_3-e_3f_0+e_1f_2-e_2f_1),
\end{equation*}  
and
\begin{equation}\label{mat:R}
\Vkt R = \begin{pmatrix} 
e_0^2+e_1^2-e_2^2-e_3^2 & 2(e_1e_2-e_0e_3) & 2(e_1e_3+e_0e_2)  \\
2(e_1e_2+e_0e_3) & e_0^2-e_1^2+e_2^2-e_3^2 & 2(e_2e_3-e_0e_1)  \\
2(e_1e_3-e_0e_2) & 2(e_2e_3+e_0e_1) & e_0^2-e_1^2-e_2^2+e_3^2
\end{pmatrix},  
\end{equation}  
if the normalizing condition $e_0^2+e_1^2+e_2^2+e_3^2=1$ is fulfilled. 

Now the solution for the direct kinematics of a pentapod with linear platform can be written as the
algebraic variety of the ideal spanned by $\Psi,\Lambda_1,\ldots ,\Lambda_5,e_0^2+e_1^2+e_2^2+e_3^2=1$. 
In the general case the dimension of this variety equals 1, as we also obtain the 1-dimensional rotation about the line $\go p$. 
In the following we present another kinematic mapping, where we get rid of this redundancy.

\subsection{Kinematic mapping for pentapods with linear platform}\label{kinmap}

With respect to the following nine homogeneous motion parameters $(n_0:x_0:\ldots :x_3:y_0:\ldots :y_3)$ with 
$n_0:=f_0^2+f_1^2+f_2^2+f_3^2$ and
\begin{align*}
x_0&:=2(e_0^2+e_1^2+e_2^2+e_3^2), &\quad y_0&:=4(-e_0f_1+e_1f_0+e_2f_3-e_3f_2), \\
x_1&:=2(-e_0^2-e_1^2+e_2^2+e_3^2), &\quad y_1&:=4(e_0f_1-e_1f_0+e_2f_3-e_3f_2), \\
x_2&:=-4(e_0e_3+e_1e_2), &\quad y_2&:=4(e_0f_2-e_1f_3-e_2f_0+e_3f_1), \\
x_3&:=4(e_0e_2-e_1e_3) &\quad y_3&:=4(e_0f_3+e_1f_2-e_2f_1-e_3f_0),
\end{align*}
the sphere condition of Eq.\ (\ref{eq:lambda}) is linear; i.e.\ 
\begin{equation}\label{sphere_compact}
\Lambda_i:\phm\frac{1}{2}(a_i^2+A_i^2+B_i^2+C_i^2-R_i^2)x_0
+a_iA_ix_1+a_iB_ix_2+a_iC_ix_3+a_iy_0+A_iy_1+B_iy_2+C_iy_3+4n_0=0.
\end{equation}
For $x_0=1$ the Euclidean displacement of $\Vkt m_i$ is given by:
\begin{equation}\label{eq:motion}
\begin{pmatrix}
a_i \\ 0 \\0 
\end{pmatrix} 
\mapsto 
\begin{pmatrix}
-a_ix_1-y_1 \\ -a_ix_2-y_2 \\-a_ix_3-y_3 
\end{pmatrix}.  
\end{equation} 

Note that $(n_0:x_0:\ldots :x_3:y_0:\ldots :y_3)$ can be interpreted as a point in the 8-dimensional projective space $P^8$. 
In order to determine the image of the kinematic mapping, we compute a Gr\"obner bases of the ideal generated by
$n_0-(f_0^2+f_1^2+f_2^2+f_3^2),\dots,y_3-4(e_0f_3+e_1f_2-e_2f_1-e_3f_0)$ and $e_0f_0+e_1f_1+e_2f_2+e_3f_3$ eliminating
the Study parameters $e_0,\dots,f_3$. It contains three quadric elements in the remaining variables, namely: 
\begin{equation}\label{phis}
\Phi_1:\phm x_1^2+x_2^2+x_3^2-x_0^2=0,\qquad
\Phi_2:\phm y_1^2+y_2^2+y_3^2-8x_0n_0=0,\qquad
\Phi_3:\phm x_1y_1+x_2y_2+x_3y_3-x_0y_0=0.
\end{equation} 
Because the degree of the image variety (which is computed from the Gr\"obner bases) is equal to 8, the three quadrics generate
the ideal of the image variety $\mathcal{I}$. The image itself consists of all real points in the zero set with $x_0\ne 0$.
Note that $\mathcal{I}$ is of dimension 5 instead of 6, as we removed the rotations around the line $\go p$, 
which do not change the spherical condition.

\begin{definition}\label{confset}
The intersection of $\mathcal{I}$ with the five hyperplanes of Eq.\ (\ref{sphere_compact}) is the complex configuration set $\mathcal{C}$ of the pentapod. 
Its real points are called real configurations. 
\end{definition}

\begin{lem}\label{los8}
A generic pentapod with linear platform has eight solutions for  the direct kinematics over $\CC$.
\end{lem}

\noindent
{\sc Proof:}
For the solution of the direct kinematics we have to intersect the image variety $\mathcal{I}$ with the five hyperplanes of $P^8$ given 
by Eq.\ (\ref{sphere_compact}). In the generic case the five hyperplanes have a linear 3-space $L$ in common, 
which intersects $\mathcal{I}$ in a finite number of points (= complex configuration set $\mathcal{C}$) 
whose cardinality equals the degree of $\mathcal{I}$. \hfill $\BewEnde$

\begin{rmk}
Lemma \ref{los8} fits with the results obtained in \cite{zhang}, where the number of 
eight solutions for  the direct kinematics problem over $\CC$ was given for planar pentapods with linear platform. 
Due to Lemma \ref{los8} this number also holds for the non-planar case.
\hfill $\diamond$
\end{rmk}

\subsection{Bonds}\label{sec:bond}

Let us assume that the pentapod under consideration has a 1-dimensional self-motion. In this case the complex 
configuration set $\mathcal{C}$ of Definition \ref{confset} is a curve. The points on this configuration curve, which do not correspond 
to Euclidean displacements as they fulfill $x_0=0$, are the so-called bonds. 
This gives a rough idea of bonds, which is developed in detail within this section.

The intersection of the 5-fold $\mathcal{I}$ with the hyperplane $x_0=0$ yields the so-called boundary of $\mathcal{I}$.
A Gr\"obner bases computation of the ideal generated by the generators of the ideal of $\mathcal{I}$ and by $x_0$
reveals 
that some perfect squares, for instance $(x_2y_3-x_3y_2)^2$, are contained in this ideal.
Therefore the radical contains the elements $\Gamma_4,\Gamma_5,\Gamma_6$ below. Its zero set 
is a 4-fold of degree 4, hence a variety of minimal degree (degree = codimension + 1). It is given by 
the following set of equations:
\begin{align}
\Gamma_1&:\phm x_1^2+x_2^2+x_3^2=0, &\quad 
\Gamma_2&:\phm y_1^2+y_2^2+y_3^2=0, &\quad 
\Gamma_3&:\phm x_1y_1+x_2y_2+x_3y_3=0, \\
\Gamma_4&:\phm x_1y_2-x_2y_1=0, &\quad 
\Gamma_5&:\phm x_1y_3-x_3y_1=0, &\quad 
\Gamma_6&:\phm x_2y_3-x_3y_2=0, 
\end{align}
and therefore it is independent of $y_0$ and $n_0$. 

\begin{definition}
The intersection of $\mathcal{C}$ with the boundary of $\mathcal{I}$ is the set $\mathcal{B}$ of bonds. 
\end{definition}

Due to $x_0=0$ the set $\mathcal{B}$ of bonds is independent of the leg lengths $R_i$ (cf.\ Eq.\ (\ref{sphere_compact})), 
and therefore $\mathcal{B}$  only depends on the geometry of the pentapod with linear platform. 

\begin{rmk}
Bonds of pentapods where already introduced in \cite{nawratil_bond} and \cite{gns2}. 
In \cite{nawratil_bond} they were defined with respect to the Study parameters and in 
\cite{gns2} with respect to a special compactification of SE(3), which can be seen as a generalization of the 
method presented above. But both approaches are not suited for the study of 
pentapods with linear platform, due to the rotational redundancy about $\go p$. \hfill $\diamond$
\end{rmk} 

\noindent
In the following we state two necessary conditions for the existence of self-motions in terms of bonds: 

Assume that a pentapod with linear platform has a $1$-dimensional configuration set $\mathcal{C}$; i.e.\ there exists a configuration curve $\go c$ on 
the 5-fold $\mathcal{I}$. Therefore the corresponding bond set $\mathcal{B}$ contains at least one bond $\beta$ (up to conjugation). 
Therefore the existence of a bond is the first necessary condition.

As we have mobility 1, the pentapod with linear platform fulfills the necessary condition of being infinitesimal movable in 
each pose of the $1$-dimensional motion of $\go p$. This is equivalent with the existence of a 
$1$-dimensional tangent space in each point of the configuration curve $\go c$.  
As $\beta\in \go c$ holds, this implies a second necessary condition for mobility 1.

\begin{rmk}\label{max}
Therefore the 3-space $L$ has to intersect the 5-fold $\mathcal{I}$ in the bond $\beta$ at least of multiplicity 2. 
As this also holds for the conjugate of $\beta$, only one pair of conjugate complex bonds can exist (due to Bezout's theorem). 
Moreover if $\beta$ (and its conjugate) is a singular point of $\mathcal{I}$ then the second necessary condition is 
trivially fulfilled. \hfill $\diamond$
\end{rmk}


\section{Non-planar pentapods of $\mathcal{P}$ with self-motions}\label{sec:contemp_duporcq}

The following determination of all non-planar pentapods of $\mathcal{P}$ with self-motions is based on the 
two necessary conditions given in Section \ref{sec:bond}. 
For an overview of the workflow/results structured by types please see Table \ref{table1}.

\begin{table}[top]
\begin{center}
\begin{tabular}{ | c ||  c | c |c |}
\hline 
Type  & Necessary Condition 1 & Necessary Condition 2 & Sufficiency \& Leg Parameters \\ \hline \hline 
1  & Theorem \ref{lem1:type1} & Theorem \ref{lem2:type1} & Theorem \ref{lem3:type1}  \\ \hline 
2  & Theorem \ref{lem1:type2} & Theorem \ref{lem2:type2} & Theorem \ref{lem3:type2} \\ \hline 
3  & Theorem \ref{thm:type3} &  &  \\ \hline 
5  & Theorem \ref{lem1:type5} & Theorem \ref{lem2:type5} & Theorem \ref{lem3:type5} \\ \hline 
\end{tabular}
\caption{Overview of the Theorems, where the first and second necessary condition for mobility
arising from bond theory are applied to the remaining types (for Type 4 see Theorem \ref{thm:type4}) of non-planar pentapods of $\mathcal{P}$. 
Moreover the theorems are listed, where the sufficiency of the obtained necessary conditions is proven. 
Within these theorems also the leg parameters for a self-motion are given. 
 }\label{table1}
\end{center}
\end{table}

\subsection{First necessary condition}

In this section we only apply  the first necessary condition, namely the existence of a bond. 

\begin{thm}\label{lem1:type5}
The base anchor points of possible leg-replacements of a pentapod of Type 5 with self-motions
have to be located on an irreducible cubic ellipse $\go s^*$ on a cylinder of revolution. 
\end{thm}

\noindent
{\sc Proof:}
Given is a pentapod of Type 5. W.l.o.g.\ we can assume that $\go M_1,\ldots ,\go M_4$ span a tetrahedron. 
Due to the properties of Type 5 (cf.\ Lemma \ref{prop5}) we can replace the sphere condition implied by the fifth leg by  
an angle condition; i.e.\ the angle  $\phi$ enclosed by the ideal point $\go m_5$ of $\go p$ and the ideal point $\go M_5$ of the base is 
constant.

Now we can choose the fixed frame $\Sigma_0$ that $\go M_1$ is the origin, $\go M_5$ is the ideal point of the $x$-axis, and 
$\go M_2$ is located in the $xy$-plane. The moving frame $\Sigma$ is chosen in a way that $\go m_1$ is its origin. 
With respect to these coordinate systems the angle condition reads as follows: 
\begin{equation}
\sphericalangle_5:\phm x_1+wx_0=0
\end{equation}
where $w$ denotes $\arccos{(\phi)}$. 
The equations of $\Lambda_i$ for $i=1,2,3,4$ are given in Eq.\ (\ref{sphere_compact}) under consideration of 
$A_1=B_1=C_1=C_2=a_1=0$. 

We set $x_0=0$ and start the computation of the bonds: 
As $\go M_1,\ldots ,\go M_4$ are non-planar ($\Leftrightarrow$ $K\neq 0$ with 
\begin{equation}
K=A_2(B_3C_4-B_4C_3)+B_2(A_4C_3-A_3C_4)
\end{equation}
holds) we can solve 
$\Lambda_1,\ldots ,\Lambda_4,\sphericalangle_5$ for $n_0,x_1,y_1,y_2,y_3$ w.l.o.g.. 
As we get $x_1=0$, the condition $\Gamma_1$ implies $x_2=\pm ix_3$. W.l.o.g.\ we 
only discuss the upper branch as the lower one can be done analogously. 

Now the numerator of $\Gamma_3$ factors into $x_3F_3[26]$ and the numerator of $\Gamma_4$ splits up 
into $x_3F_4[10]$. Therefore we have to distinguish two cases:
\begin{enumerate}[1.]
\item
$x_3=0$:
Now the numerator of $\Gamma_2$ factors into $y_0^2F_2[41]$, where $F_2$ is quadratic 
with respect to $a_4$. The discriminant of $F_2$ with respect to $a_4$ equals:
\begin{equation}
-K^2\left[
(a_2A_3-a_3A_2)^2+(a_2B_3-a_3B_2)^2+a_2^2C_3^2
\right].
\end{equation}
This expression can never be greater than zero. It is equal to zero if the following three conditions are fulfilled:
\begin{equation}
a_2A_3-a_3A_2=0,\qquad
a_2B_3-a_3B_2=0, \qquad
a_2C_3=0.
\end{equation}
As $a_2=a_3=0$ contradicts assumption (i) we have to discuss the following cases:
	\begin{enumerate}[(a)]
	\item
	$a_2=0$: This implies $A_2=B_2=0$ and therefore the first and second leg coincide; a contradiction.
	\item
	$a_3=0$, $a_2\neq 0$: We get $A_3=B_3=C_3=0$ and therefore the first and third leg coincide; a contradiction.
	\item
	$a_2a_3\neq 0$: We get $C_3=0$, $A_3=a_3A_2/a_2$ and $B_3=a_3B_2/a_2$ but this implies $K=0$; a contradiction. 
	\end{enumerate}
As a consequence the case $x_3=0$ does not imply any solution to our problem.
\item
$x_3\neq 0$: In this case $F_3[26]=0$ and $F_4[10]=0$ have to hold. Computing the resultant of these two expressions with 
respect to $y_0$ yields $-x_3K(G_r+iG_c)$ with
\begin{equation}\label{GrGc}
\begin{split}
G_r&=\phm \left[a_4B_3(a_2-a_3)+a_3B_4(a_4-a_2)\right]B_2+a_2(a_3-a_4)(B_3B_4-C_3C_4), \\
G_c&=-\left[a_4C_3(a_2-a_3)+a_3C_4(a_4-a_2)\right]B_2-a_2(a_3-a_4)(B_3C_4+B_4C_3).
\end{split}
\end{equation}
These two expressions do not depend on the $A_i$ coordinates. We denote the orthogonal projection 
of the base anchor point $\go M_i$ onto the $yz$-plane of the fixed frame $\Sigma_0$ by $\go M_i^{\prime}$ for $i=1,\ldots, 4$. 
In the following we want to show that $\go M_1^{\prime},\ldots,\go M_4^{\prime}$ has to be pairwise distinct as well as 
$\go m_1,\ldots ,\go m_4$. 
The proof is done by contradiction, where the following cases have to be discussed: 
	\begin{enumerate}[(a)]
	\item
	Two platform anchor points coincide: W.l.o.g.\ we can set $\go m_1=\go m_2$; i.e.\ $a_2=0$. Then $G_r$ and $G_c$ simplify to
	\begin{equation}
	G_r=-a_3a_4B_2(B_3-B_4), \qquad G_c=a_3a_4B_2(C_3-C_4). 
	\end{equation}
	As $a_3a_4=0$ contradicts assumption (i), and $\go M_3^{\prime}=\go M_4^{\prime}$ ($\Rightarrow$ $\go M_3$, $\go M_4$, $\go M_5$ collinear) 
	implies a contradiction to assumption (ii) we remain with the case $\go M_1^{\prime}=\go M_2^{\prime}$:
	
		In this case $\go m_1=\go m_2$ is located on a circle in a plane orthogonal to the axis of the Sch\"onflies motion. 
		As the point $\go m_1=\go m_2$ (and therefore the complete line $\go p$) cannot be translated in direction of $\go M_5$, the problem 
		reduces to a planar one (projection to the $yz$-plane of $\Sigma_0$). 
		As a circular translation of the resulting planar manipulator is not possible (as otherwise the base has to be planar; cf.\ Remark \ref{rmk:bond}), 
		anchor points have to coincide (cf.\ item 2 in the proof of Theorem \ref{thm:planar}). 
		\begin{enumerate}[i.]
		\item
		If further base anchor points coincide beside $\go M_1^{\prime}=\go M_2^{\prime}$ we get again a contradiction to the assumption $\go M_1,\ldots ,\go M_4$ are non-planar. 
		\item
		Further platform anchor points (beside $\go m_1^{\prime}=\go m_2^{\prime}$) can only coincide without contradicting assumption (i) or (ii)
		if $\go p$ is parallel to the axis of the Sch\"onflies motion. 
		In this case the platform of the planar manipulator collapse into a point and we only get the 
		trivial rotation about the line $\go p$ as uncontrollable motion while $\go p$ itself remains fix.
		\end{enumerate}
	\item
	No platform anchor points coincide and two projected base anchor points coincide: W.l.o.g.\ we can assume that 
	$\go M_1^{\prime}=\go M_2^{\prime}$ holds; i.e.\ $B_2=0$. Then $G_r$ and $G_c$ simplify to
	\begin{equation}
	G_r=a_2(a_3-a_4)(B_3B_4-C_3C_4), \qquad G_c=-a_2(a_3-a_4)(B_3C_4+B_4C_3), 
	\end{equation}
	which cannot vanish without contradiction. 
\end{enumerate}
As a consequence $\go M_1^{\prime},\ldots,\go M_4^{\prime}$ have to be pairwise distinct, as well as 
$\go m_1,\ldots ,\go m_4$. Therefore there exists a uniquely defined  M\"obius transformation $\tau$ with $\go m_i\mapsto \go M_i^{\prime}$ for $i=1,2,3$ 
(cf.\ Section \ref{geom:basic}); i.e.\ $B_i+iC_i=\tau(a_i)$ holds with $\tau$ of Eq.\ (\ref{tau}) and   
\begin{equation}\label{moeb}
z_1= \frac{a_2-a_3}{B_2-B_3-iC_3},\quad 
z_2=0, \quad
z_3=\frac{a_2B_3-B_2a_3+ia_2C_3}{(B_2-B_3-iC_3)(B_3+iC_3)B_2}, \quad 
z_4=\frac{a_2a_3B_2}{B_3+iC_3}.
\end{equation}
Now it can easily be verified that the conditions $G_r=0$ and $G_c=0$ of Eq.\ (\ref{GrGc}) imply that $\tau$ also 
maps $\go m_4$ onto $M_4^{\prime}$. 
As $\go M_1,\ldots ,\go M_4$ span a tetrahedron, $\go M_1^{\prime},\ldots,\go M_4^{\prime}$ has to be located on a circle.
Therefore the base anchor points of possible leg-replacements have to belong to the cylinder of revolution $\Theta$ 
through $\go M_1,\ldots,\go M_4$ with generators in direction of $\go M_5$. 
\end{enumerate}

\noindent
In the following we study the possible leg-replacements for this case in more detail:  
By homogenizing the matrix of Eq.\ (\ref{matrix_cubic}) it is not difficult to see that the corresponding matrix reads as follows: 
\begin{equation}
\begin{pmatrix}
a_2 & A_2 & B_2 & 0   & a_2A_2 & a_2B_2 & 0 \\
a_3 & A_3 & B_3 & C_3 & a_3A_3 & a_3B_3 & a_3C_3 \\
a_4 &	A_4 & B_4 & C_4 & a_4A_4 & a_4B_4 & a_4C_4 \\
0   &  0  & 0   & 0   & 1      & 0      & 0 
\end{pmatrix}.
\end{equation}
As $\go M_1,\ldots ,\go M_4$ are not coplanar, we have $D_{167}\neq 0$. 
But due to $D_{156}= 0$ and $D_{157}= 0$ Eq.\ (\ref{centraleq_nonplanar_spezi}) simplifies to
\begin{equation}
\begin{pmatrix}
D_{267} & -D_{367} &D_{467} \\
D_{126} & -D_{136} & D_{146}+aD_{167} \\
D_{127} & aD_{167}-D_{137} & D_{147}
\end{pmatrix}
\begin{pmatrix}
A \\ B \\ C
\end{pmatrix}
=
a
\begin{pmatrix}
D_{167} \\ 0 \\0
\end{pmatrix}.
\end{equation}
Solving this system of linear equations yields a solution of the form given in Eq.\ (\ref{cramer}), 
but now  $d_0(a)$ is a quadratic expression in $a$ and $d_1(a)$ a cubic one. 
Therefore the base anchor points belong to a cubic curve $\go s^*$, which has to be located on $\Theta$. 
 
Moreover $\go s^*$ cannot split up into three generators as $\go M_1^{\prime},\ldots,\go M_4^{\prime}$ are pairwise distinct. 
The cubic $\go s^*$ can also not split up into a conic $\go q$ and a generator $\go g_1$ for the following reason: 
This case equals Type 2 where $\go P_1$ is not a finite point but the ideal point $\go U$ of $\go p$. 
Therefore $\go M_i$ has to be located on $\go q\setminus\left\{\go Q\right\}$, as otherwise 
$\go m_i$ equals $\go U$, which does not yield a sphere condition. As this has to hold for $i=1,\ldots ,4$ 
we get a contradiction to the non-planarity assumption. 

Therefore the cubic curve has to be an irreducible cubic curve $\go s^*$ on a cylinder of revolution. 
From the following theorem of projective geometry it is clear that the ideal plane cannot be an osculating plane of $\go s^*$  
or be tangent to it: 
{\it The osculating plane in a point $\go X$ of a cubic equals the tangent plane to the cone of chords with respect $\go X$ along the 
tangent of $\go X$ (which is a generator of the cone of chords).} 
Therefore the remaining two intersection points of $\go s^*$ with the plane at infinity are conjugate complex. 
\hfill $\BewEnde$


\begin{thm}\label{lem1:type1}
The irreducible cubic $\go s$ of a pentapod of Type 1 with self-motions has to be a cubic ellipse located on a cylinder of revolution. 
\end{thm}

\noindent
{\sc Proof:} 
$\go s$ has at least one real intersection point with the ideal plane, which is denoted by $\go M_4$. 
As the ideal point $\go m_5$ of $\go p$ is mapped to a finite point $\go M_5$ of the base, 
the point $\go M_4\sigma^{-1}$ has to be a finite point of $\go p$, which is denoted by $\go m_4$. 
Therefore this point pair $(\go M_4,\go m_4)$ determines a Darboux condition $\Omega_4$ and the 
point pair  $(\go M_5,\go m_5)$ a Mannheim condition $\Pi_5$. 
The remaining three point pairs  $(\go M_i,\go m_i)$ imply sphere conditions $\Lambda_i$ for $i=1,2,3$. 

Moreover we choose the fixed frame $\Sigma_0$ that 
$\go M_1$ is the origin, $\go M_4$ the ideal point of the $x$-axis and $\go M_2$ is located 
in the $xy$-plane. Moreover we can define the moving frame $\Sigma$ in a way that 
$\go m_1$ is its origin. 
With respect to these coordinate systems our conditions can be written as: 
\begin{equation}
\Omega_4:\phm p_4x_0+a_4x_1+y_1=0, \qquad
\Pi_5:\phm p_5x_0+A_5x_1+B_5x_2+C_5x_3+y_0=0,
\end{equation}
where $(p_4,0,0)^T$ are the coordinates of the intersection point of the Darboux plane and the $x$-axis of $\Sigma_0$, and 
$(p_5,0,0)^T$ are the coordinates of the intersection point of the Mannheim plane and the $x$-axis of $\Sigma$. 
The equations of $\Lambda_i$ for $i=1,2,3$ are given in Eq.\ (\ref{sphere_compact}) under consideration of 
$A_1=B_1=C_1=C_2=a_1=0$. 

We set $x_0=0$ and start the computation of the bonds: 
Due to the properties of Type 1 no four base points can be coplanar ($\Rightarrow$ $B_2C_3C_5\neq 0$) 
and no two platform anchor points can coincide. Under these assumptions we can solve the equations $\Lambda_1,\Lambda_2,\Lambda_3,\Omega_4,\Pi_5$ for $y_0,y_1,y_2,y_3,n_0$. 
Now the numerator of $\Gamma_4$ factors into $x_1F$ with 
\begin{equation}\label{ex:F}
F:=(a_2A_2-a_4A_2-a_2A_5)x_1+(a_2B_2-a_4B_2-a_2B_5)x_2-a_2C_5x_3.
\end{equation}
Therefore we have to distinguish two cases:
\begin{enumerate}[1.]
\item
$x_1\neq 0$: 
In this case $F=0$ has to hold, which can be solve w.l.o.g. for $x_3$. 
Then the numerator of $\Gamma_6$ factors into $x_2G$ with
\begin{equation}
G:=(A_2x_1+B_2x_2)(a_2-a_4)(a_3C_3-a_3C_5-a_4C_3)+
(A_3x_1+B_3x_2)(a_3-a_4)a_2C_5-
(A_5x_1+B_5x_2)(a_3-a_4)a_2C_3. 
\end{equation}  
We distinguish two cases: 
	\begin{enumerate}[(a)]
	\item
	$x_2\neq 0$: In this case $G=0$ has to hold. We define  $H:=A_2(a_2-a_4)(a_3C_3-a_3C_5-a_4C_3)+A_3(a_3-a_4)a_2C_5-A_5(a_3-a_4)a_2C_3$ and 
	discuss the following two cases: 
		\begin{enumerate}[i.]
		\item
		$H\neq 0$: Under this assumption we 
		can solve $G=0$ for $x_1$. 
		Then the remaining equations only imply one condition which is quadratic with respect to $B_3$. The discriminant of this 
		condition with respect to $B_3$ equals:
		\begin{equation}
		-H^2\left[(a_2A_2-a_4A_2-a_2A_5)^2+(a_2B_2-a_4B_2-a_2B_5)^2+a_2^2C_5^2\right].
		\end{equation}
		Therefore $B_3$ cannot be real; a contradiction.
		\item
		$H=0$: We can solve $H=0$ for $A_3$ w.l.o.g.. Then we can solve the $G=0$ for $B_3$ w.l.o.g., which already yields the 
		contradiction, as now the points $\go M_1,\go M_2,\go M_3 ,\go M_5$ are coplanar.
		\end{enumerate}
	\item
	$x_2=0$: Now the numerator of $\Gamma_1$ factors into 
	\begin{equation}
	x_1^2\left[A_2^2(a_2-a_4)^2-2a_2A_2A_5(a_2-a_4)+(A_5^2+C_5^2)a_2^2\right].  
	\end{equation}
	The discriminant with respect to $A_2$ equals $-C_5^2$ and therefore we get a contradiction. 
	\end{enumerate}
\item
$x_1=0$: From $\Gamma_1$ we get $x_2=\pm x_3i$. In the following we only discuss the case $x_2= x_3i$, as the 
other one can be done analogously. Now the numerator of $\Gamma_6$ factors into $x_3^2(G_r+iG_c)$ with
\begin{equation}
G_r:=(B_2B_3-B_3B_5+C_3C_5)a_2-B_2(B_3-B_5)a_3, \quad
G_c:=B_2(C_3-C_5)a_3-(B_2C_3-B_3C_5-B_5C_3)a_2.
\end{equation}
This are the corresponding expressions to Eq.\ (\ref{GrGc}).
Therefore the two point sets $\go M_1^{\prime},\go M_2^{\prime},\go M_3^{\prime},\go M_5^{\prime}$ and 
$\go m_1,\go m_2,\go m_3,\go m_5$ are again M\"obius equivalent.  
As a consequence $\go s$ has to be located on the cylinder of revolution $\Theta$ 
through $\go M_1^{\prime},\go M_2^{\prime},\go M_3^{\prime},\go M_5^{\prime}$ with generators in direction of $\go M_4$. 
By using again the theorem of projective geometry given at the end of the last proof we are done. 
\hfill $\BewEnde$
\end{enumerate}


\begin{thm}\label{lem1:type2}
The conic $\go q$ of a pentapod of Type 2 with self-motions has to be located 
on a cylinder of revolution, where one generator is the line $\go g_1$.  
\end{thm}  

\noindent
{\sc Proof:}
The proof can be done analogously to the one of Theorem \ref{lem1:type1}. In order to streamline the presentation it is given in 
Appendix A. It should only be noted that $\go q$ can be an ellipse or a circle, respectively, as $\go q$ is the 
planar section of a cylinder of revolution. \hfill $\BewEnde$


\begin{thm}\label{thm:type3}
A pentapod of Type 3 cannot possess a self-motion.
\end{thm}

\noindent
{\sc Proof:}
In order to improve the readability of the paper the proof of the non-existence of 
pentapods of Type 3 with self-motions is given in Appendix B. \hfill $\BewEnde$


\subsection{Second necessary condition}

Due to the obtained results only pentapods of Type 1,2,5 remain as candidates for self-motions. 
In this section we check them with respect to the second necessary condition.

\begin{thm}\label{lem2:type1}
The irreducible cubic $\go s$ of a pentapod of Type 1 with self-motions has to be a straight cubic circle. 
\end{thm}

\noindent
{\sc Proof:}
Due to Theorem \ref{lem1:type1} the cubic $\go s$ has three pairwise distinct points at infinity, which are denoted by 
$\go M_2, \go M_3, \go M_4$. Note that $\go M_4$ is real and that $\go M_2, \go M_3$ are conjugate complex; i.e.\ $\overline{\go M_2}=\go M_3$. 
The corresponding platform anchor points are denoted by $\go m_2, \go m_3, \go m_4$ where $\overline{\go m_2}=\go m_3$ holds. 
Therefore we get three Darboux conditions $\Omega_i$ implied by the point pairs $(\go M_i,\go m_i)$ for $i=2,3,4$. 
The ideal point of the line $\go p$ is again denoted by $\go m_5$ and its corresponding base anchor point with $\go M_5$.
Therefore this point pair implies one Mannheim condition $\Pi_5$. 
The pentapod is completed by a sphere condition $\Lambda_1$ determined by the two finite points $\go M_1$ and $\go m_1$.

The fixed frame $\Sigma_0$ is chosen that  
$\go M_1$ is the origin and that $\go M_2$ and $\go M_3$ are located in the $xy$-plane in 
direction $(1,B_2,0)$ and $(1,\overline{B_2},0)$, respectively. 
As $\go M_2, \go M_3, \go M_4$ cannot be collinear, $\go M_4$ is the ideal point in direction of $(A_4,B_4,1)$. 
Moreover we locate the origin of the moving frame $\Sigma$ in $\go m_1$. 
With respect to these coordinate systems $\Sigma$ and $\Sigma_0$ our conditions can be written as: 
\begin{equation}
\begin{split}
\Omega_j&:\phm p_jx_0+a_jx_1+a_j\overline{B_j}x_2+y_1+ \overline{B_j}y_2=0, \\
\Omega_4&:\phm p_4x_0+a_4A_4x_1+a_4B_4x_2+a_4x_3+A_4y_1+B_4y_2+y_3=0, \\
\Pi_5&:\phm p_5x_0+A_5x_1+B_5x_2+C_5x_3+y_0=0,
\end{split}
\end{equation}
where $(p_j,0,0)^T$ for $j=2,3$ are the coordinates of the intersection point of the Darboux plane and the $x$-axis of $\Sigma_0$,
$(0,0,p_4)^T$ are the coordinates of the intersection point of the Darboux plane and the $z$-axis of $\Sigma_0$, and 
$(p_5,0,0)^T$ are the coordinates of the intersection point of the Mannheim plane and the $x$-axis of $\Sigma$. 
The equation of $\Lambda_1$ is given in Eq.\ (\ref{sphere_compact}) under consideration of 
$A_1=B_1=C_1=a_1=0$. 

With respect to the chosen frames $\Sigma$ and $\Sigma_0$ the first necessary condition is fulfilled if 
\begin{equation}\label{ex:Br+Bc}
B_r = \frac{A_4B_4}{A_4^2+1}, \quad
B_c=\pm \frac{\sqrt{A_4^2+B_4^2+1}}{A_4^2+1}, 
\end{equation}
holds with $B_2=B_r+iB_c$ and $B_c\neq 0$. Then the bond $\beta$ reads as follows: 
\begin{equation}
n_0=0,\quad
x_0=0,\quad 
x_1 = -\overline{B_2}, \quad
x_2=1, \quad
x_3=A_4\overline{B_2}-B_4, 
\end{equation}
\begin{equation}
y_0= \overline{B_2}(A_5-A_4C_5)-B_5+B_4C_5, \quad
y_1=\overline{B_2}\overline{a_2},\quad
y_2=-\overline{a_2}, \quad
y_3=-\overline{a_2}(A_4\overline{B_2}-B_4).
\end{equation}

Now we apply the second necessary condition; i.e.\ 
the eight tangent-hyperplanes to $\Phi_1,\Phi_2,\Phi_3,\Lambda_1,\Omega_2,\Omega_3,\Omega_4,\Pi_5$ in the bond $\beta$ have to have a line in common. 
Therefore we compute the gradients of these eight hypersurfaces with respect to the unknown $n_0,x_0,\ldots ,x_3,y_0,\ldots,y_3$ in the bond $\beta$. 
The resulting $8\times 9$ matrix $\Vkt J$ has rank 8 ($\Rightarrow$ $\beta$ is a regular point of the 5-fold $\mathcal{I}$). 
For the necessary condition $rk(\Vkt J)<8$ the determinants of all  $8\times 8$ submatrices of $\Vkt J$ have to vanish. 
The numerator of the determinant of the $8\times 8$ submatrix of $\Vkt J$, obtained by  
removing the column steaming from the partial derivative with respect to $y_2$, factors into 
$(ia_r-a_c-ia_4)(ia_r+a_c)(A_4^2+B_4^2+1)L_1L_2$ with 
\begin{equation}\label{ls}
L_1=A_4\sqrt{A_4^2+B_4^2+1}\mp iB_4,\qquad
L_2=B_5+B_5A_4^2-B_4C_5-A_4A_5B_4 \pm i(A_5-A_4C_5)\sqrt{A_4^2+B_4^2+1},
\end{equation}
and $a_2=a_r+ia_c$, where $a_c\neq 0$ holds.
It can easily be seen that $L_2$ can only vanish if $\go M_1,\go M_4,\go M_5$ are collinear, which yields a contradiction. 
Therefore $L_1=0$ has to hold, which can only be the case for $A_4=B_4=0$ ($\Rightarrow$ $rk(\Vkt J)=7$). 
This implies $B_r=0$ and $B_c=\pm i$, which shows that $\go s$ is a straight cubic circle. \hfill $\BewEnde$


\begin{thm}\label{lem2:type2}
The conic $\go q$ of a pentapod of Type 2 with self-motions has to be 
a circle and the line $\go g_1$ is orthogonal to its carrier plane (= degenerated case of a straight cubic circle). 
\end{thm}

\noindent
{\sc Proof:}
The proof can exactly be done as for Theorem \ref{lem2:type1} under consideration of $a_4=0$, $C_5=0$ 
and that no two platform anchor points can coincide beside $\go m_1$ and $\go m_4$. 

It should only be noted that in this case $L_2$ vanishes for $\go M_1=\go M_5$, which yields a contradiction to the properties 
of Type 2.\footnote{In this case the cubic of base points splits up into three lines; one real and two conjugate complex ones, which intersect each other in 
$\go M_1=\go M_5$ (complex version of Type 4).}
Therefore we remain again with the solution $A_4=B_4=0$, which implies $B_r=0$ and $B_c=\pm i$. 
This proves the theorem. \hfill $\BewEnde$


\begin{thm}\label{lem2:type5}
The cubic $\go s^*$ of a pentapod of Type 5 with self-motions has to be an irreducible straight cubic circle. 
\end{thm}

\noindent
{\sc Proof:}
The proof of this theorem can be done in a similar fashion as those of Theorems \ref{lem2:type1} and \ref{lem2:type2}. 
In order to streamline the presentation the proof of Theorem \ref{lem2:type5} is given in Appendix C. \hfill $\BewEnde$\\

\noindent
Due to the Theorems \ref{thm:type4},\ref{thm:type3},\ref{lem2:type1},\ref{lem2:type2},\ref{lem2:type5} 
the condition of Duprocq \cite{duporcq}, that the centers of the 
spheres have to be located on a straight cubic circle, is valid for non-planar pentapods of $\mathcal{P}$. 
Note that Duporcq also mentioned explicitly the special cases of Type 2 and Type 5 beside the general case of Type 1.
Therefore it remains to show if this so-called Duporcq condition is already sufficient for the existence of a self-motion. 
This is done in the next section.


\subsection{Sufficiency of the Duporcq condition}

The sufficiency is proven separately for the Types 1,2,5. Moreover in each of the 
three proofs also the leg parameters for a self-motion are given.

\begin{thm}\label{lem3:type5}
A pentapod of Type 5 fulfilling the Duporcq condition has a  1-parametric set of self-motions (over $\CC$). 
With respect to the coordinatization used in the proof of Theorem \ref{lem2:type5} (under consideration of $A_4=B_4=0$ and $B_2=i$) 
the leg parameters are given by: 
\begin{equation}
w=\frac{C_5}{a_5},\quad
p_2=-\frac{(a_3-a_5)(A_5-iB_5)}{a_5},\quad
p_3=-\frac{(a_2-a_5)(A_5+iB_5)}{a_5},
\end{equation}
and the following condition remains in $R_1$ and $R_5$:
\begin{equation}\label{rem:eq}
(a_5^2+B_5^2+C_5^2)(a_2+a_3-a_5)+(R_1^2-R_5^2-a_2a_5-a_3a_5+a_5^2)a_5=0. 
\end{equation}
\end{thm}

\noindent
{\sc Proof:}
We use the same coordinatization as given in the first two paragraphs of the proof of Theorem \ref{lem2:type5} under consideration of $A_4=B_4=0$ and 
$B_2=i$. We distinguish two cases:
\begin{enumerate}[1.]
\item
$C_5-a_5w\neq 0$: Under this assumption we can solve $\Lambda_1,\Omega_2,\Omega_3,\sphericalangle_4,\Lambda_5,\Phi_3$ for $x_3,y_0,y_1,y_2,y_3,n_0$. 
Plugging the obtained expressions into $\Phi_i$ yields the equations $\Phi_i^*$ in $x_0,x_1,x_2$ for $i=1,2$. 
Now we compute the resultant $\Xi$ of the numerator of  $\Phi_1^*$ (which is quadratic in $x_0,x_1,x_2$) 
and the numerator of  $\Phi_2^*$ (which is quartic in $x_0,x_1,x_2$) with respect to $x_2$. 
$\Xi$ factors into $x_0^4N[11464]$, where $N$ is quartic in $x_0,x_1$. 
As $x_0\neq 0$ has to hold (cf.\ Section \ref{kinmap}), $N$ has to be fulfilled identically. Therefore we 
denote the coefficient of $x_0^ix_1^j$ of $N$ by $N_{ij}$.
Then $N_{04}$ factors into $16E_1^2E_2^2$ with
\begin{equation}
E_1=(a_3-a_5)(A_5-iB_5)+a_5p_2,\qquad
E_2=(a_2-a_5)(A_5+iB_5)+a_5p_3.
\end{equation}
W.l.o.g.\ we can set $E_1$ equal to zero and solve it for $p_2$. 
Then the numerator of $N_{13}$ factors into $32a_3(a_3-a_5)(A_5-iB_5)(C_5-a_5w)^2E_2^2$.
Therefore $E_2=0$ has to hold, which can be solved for $p_3$ w.l.o.g..
Now the  numerator of $N_{22}$ splits up into
\begin{equation}
64(a_5^2+B_5^2)a_2a_3(a_2-a_5)(a_3-a_5)(C_5-a_5w)^4,
\end{equation}
which cannot vanish without contradiction. 
\item
$w=C_5/a_5$: 
We can solve the equations  $\Lambda_1,\Omega_2,\Omega_3,\sphericalangle_4,\Lambda_5$ for $x_3,y_1,y_2,y_3,n_0$ and plug the 
obtained expressions into $\Phi_i$, which yields quadratic equation $\Phi_i^*$ in $x_0,x_1,x_2,y_0$ for $i=1,2,3$. 
Now then numerator of $\Phi_1^*$ and $\Phi_3^*$ do not depend on $y_0$ in contrast to the numerator of $\Phi_2^*$ (coefficient of $y_0^2$ equals $4a_5^2$). 
Therefore we compute the resultant $\Xi$ of the numerators of  $\Phi_1^*$ 
and the numerator of  $\Phi_3^*$ with respect to $x_2$.
$\Xi$ factors into $x_0^2a_5^2N[208]$, where $N$ is quadratic in $x_0,x_1$. 
We denote the coefficient of $x_0^ix_1^j$ of $N$ again by $N_{ij}$.
In the following we show that $E_1=0$ and $E_2=0$ has to hold:

$N_{02}$ factors into $4a_5^2E_1E_2$, which can be solve w.l.o.g.\ for $p_2$. 
Then $N_{11}$ splits up into 
\begin{equation}
-2a_5E_2\left[
a_3(A_5^2+B_5^2)+(a_2+a_3-a_5)C_5^2+(1-a_2-a_3)a_5^2+(R_1^2-R_5^2)a_5-(A_5-iB_5)a_5p_3
\right].
\end{equation}
Either $E_2=0$ holds and we are done or the last factor vanishes. In the latter case we can compute $p_3$ w.l.o.g.. Then 
$N_{20}$ factors into $(C_5^2-a_5^2)E_2^2$. For $C_5=\pm a_5$ the condition $\Phi_1$ equals $x_1^2+x_2^2=0$, which cannot yield 
a real self-motion. Therefore $E_2=0$ has to hold. 

Summed up we have proven that $E_1=0$ and $E_2=0$ have to hold. These equations can be solved for $p_2,p_3$
w.l.o.g.. Plugging the obtained expressions into $N$ shows that only the condition given in Eq.\ (\ref{rem:eq}) remains. 
This condition can always be solved for $R_5^2$ and the self-motion is again obtained by back-substitution, which finishes the 
proof of the sufficiency.  \hfill $\BewEnde$
\end{enumerate}


\begin{thm}\label{lem3:type1}
A pentapod of Type 1 fulfilling the Duporcq condition has a  1-parametric set of self-motions (over $\CC$). 
With respect to the coordinatization used in the proof of Theorem \ref{lem2:type1} (under consideration of $A_4=B_4=0$ and $B_2=i$) 
the leg parameters are given by: 
\begin{equation}
p_2=-\frac{A_5(a_2a_3-a_4^2)-i(a_2a_3-a_4^2)B_5}{(a_3-a_4)^2},\quad
p_3=-\frac{A_5(a_2a_3-a_4^2)+i(a_2a_3-a_4^2)B_5}{(a_2-a_4)^2},\quad
p_4=\frac{C_5(a_2a_3-a_4^2)}{(a_2-a_4)(a_3-a_4)},
\end{equation}
and the following condition remains in $R_1$ and $p_5$:
\begin{equation}\label{last}
\begin{split}
&(a_2-a_4)^2(a_3-a_4)^2
\left[
2(a_2a_3-a_4^2)p_5-(a_2+a_3-2a_4)R_1^2-(2a_2a_3-a_2a_4-a_3a_4)a_4 
\right] + \\
&(a_2a_3-a_4^2)^2(a_2+a_3-2a_4)(A_5^2+B_5^2+C_5^2)=0.
\end{split}
\end{equation}
\end{thm}


\begin{thm}\label{lem3:type2}
A pentapod of Type 2 fulfilling the Duporcq condition has a  1-parametric set of self-motions (over $\CC$). 
With respect to the coordinatization used in the proof of Theorem \ref{lem3:type1} (under consideration of $a_4=C_5=0$)
the leg parameters are given by: 
\begin{equation}
p_2=-\frac{a_2(A_5-iB_5)}{a_3},\quad
p_3=-\frac{a_3(A_5+iB_5)}{a_2},\quad
p_4=0,
\end{equation}
and the following condition remains in $R_1$ and $p_5$:
\begin{equation}\label{fin:eq}
(A_5^2+B_5^2)(a_2+a_3)+2a_2a_3p_5-R_1^2(a_2+a_3)=0.
\end{equation}
\end{thm}

\noindent
{\sc Proof of Theorem \ref{lem3:type1} and Theorem \ref{lem3:type2}:} 
The proofs of these theorems can be done in a fashion similar to the one of Theorem \ref{lem3:type5}. 
For this reason and in order to streamline the presentation the corresponding proofs are given in Appendix D and Appendix E, respectively. \hfill $\BewEnde$ \\


\noindent
Note that the proof of the sufficiency in the Theorems \ref{lem3:type5}, \ref{lem3:type1} and \ref{lem3:type2} 
was only done over $\CC$; i.e.\ the self-motion has not to be real.\footnote{In contrast the conditions given in Theorem \ref{thm:planar} 
and those for the cases ($\alpha,\beta,\gamma$) of Section \ref{intro} are even sufficient for the existence of real self-motions.}
Arguments of reality were only used to exclude some special cases.

\begin{figure}[top]
\begin{center} 
\subfigure[]{ 
 \begin{overpic}
    [width=74mm]{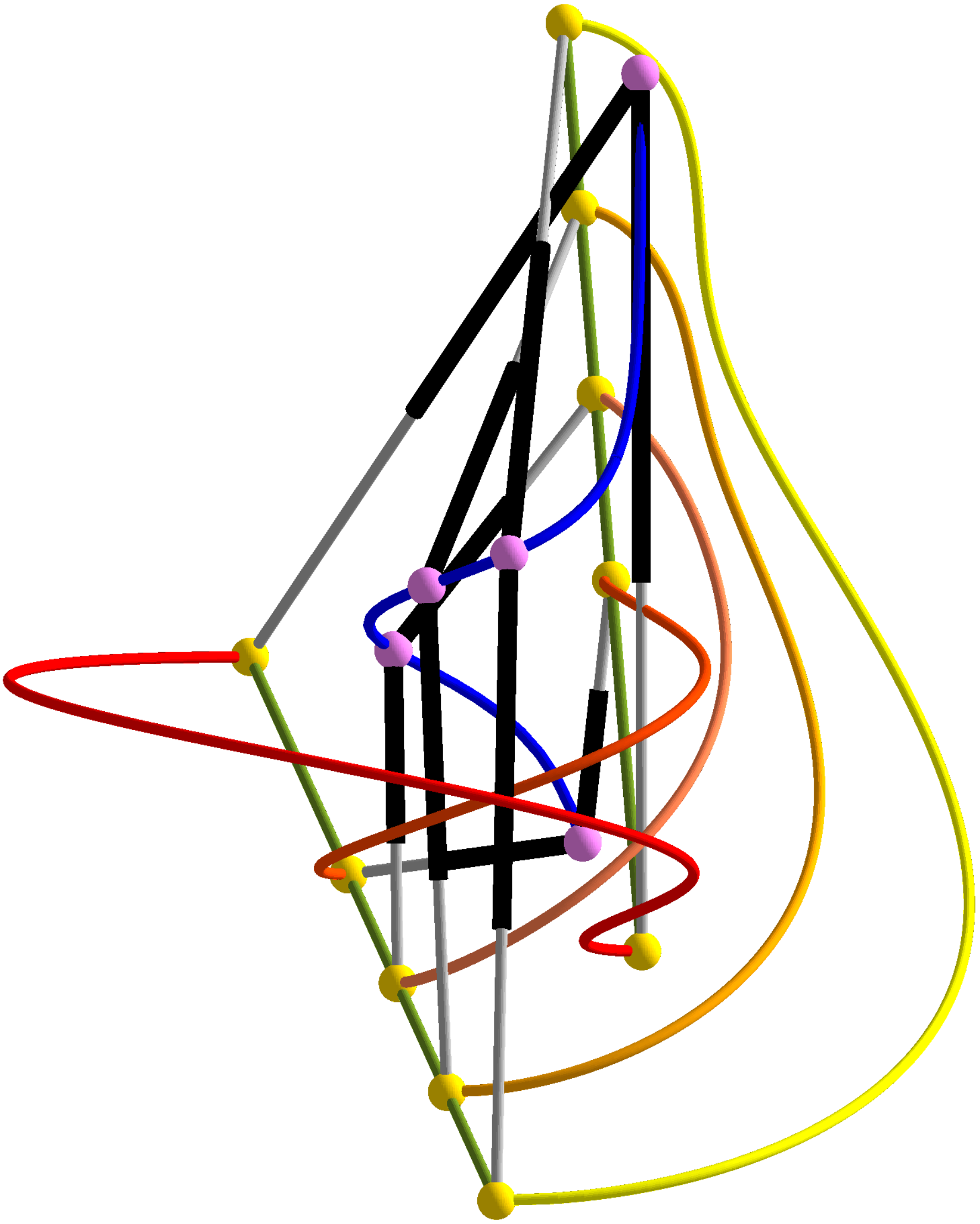}
  \contourlength{0.2mm}
  \begin{small} 
    \put(22,25.5){\contour{white}{$\go m_6^-$}}    
    \put(26,18.5){\contour{white}{$\go m_1^-$}}    
    \put(30,10){\contour{white}{$\go m_8^-$}}    
    \put(33.8,1.5){\contour{white}{$\go m_9^-$}}    
    \put(15,48.5){\contour{white}{$\go m_7^-$}}    
    \put(46.5,18.5){\contour{white}{$\go m_7^+$}}    
    \put(53.5,51){\contour{white}{$\go m_6^+$}}    
    \put(43,67.5){\contour{white}{$\go m_1^+$}}    
    \put(39,83){\contour{white}{$\go m_8^+$}}    
    \put(40,97){\contour{white}{$\go m_9^+$}}
    \put(54,95){\contour{white}{$\go M_7$}}
    \put(43,50.5){\contour{white}{$\go M_9$}}    
    \put(28.8,53){\contour{white}{$\go M_8$}}    
    \put(27,44){\contour{white}{$\go M_1$}}    
    \put(43,27.5){\contour{white}{$\go M_6$}}    
    \put(45,58.5){\contour{white}{$\go s$}}    
	\end{small} 
  \end{overpic}
}
\qquad
\subfigure[]{
\begin{overpic}
    [width=45mm]{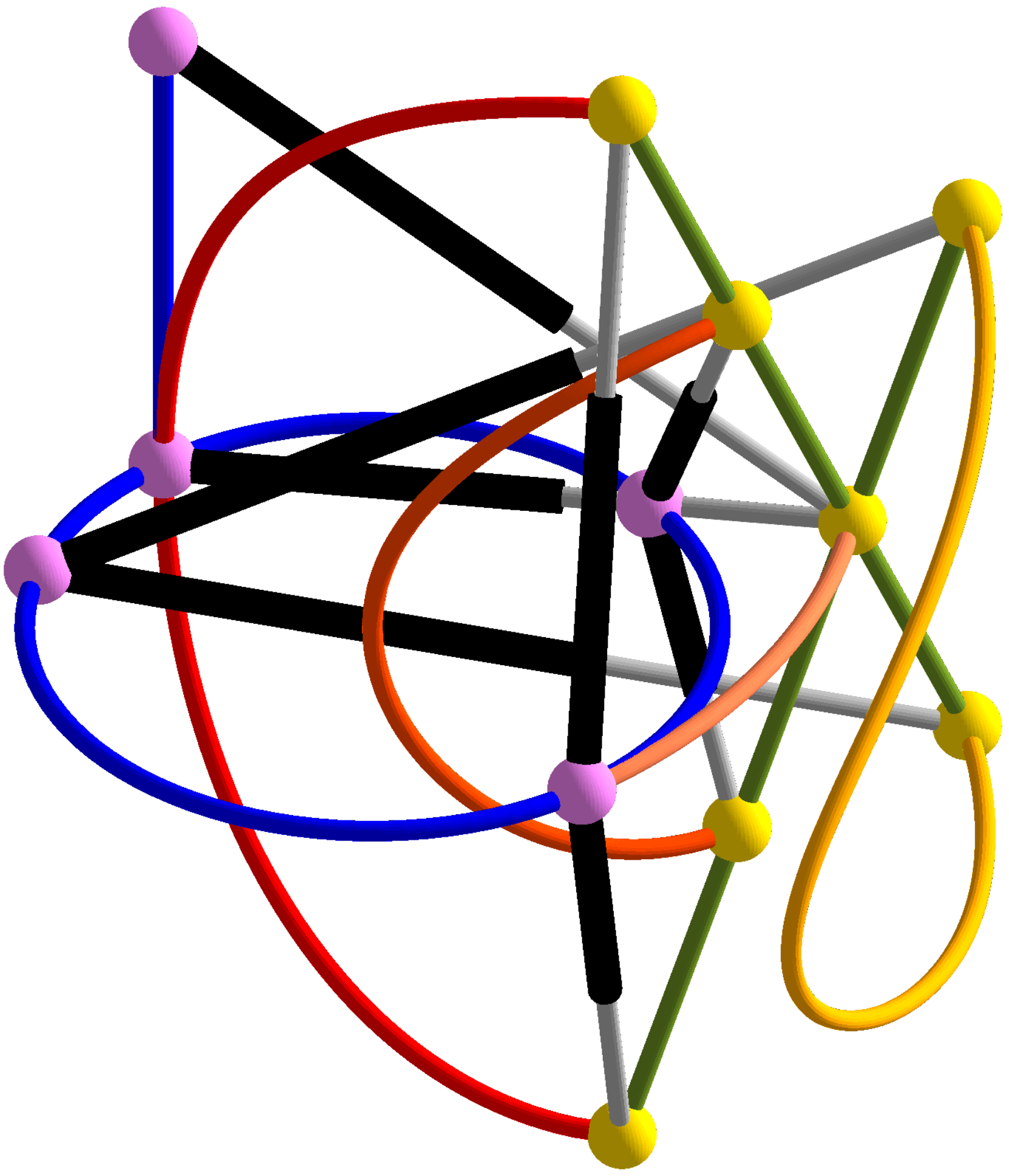}
  \contourlength{0.2mm}
  \begin{small} 
    \put(60.5,79.5){\contour{white}{$\go m_6^-$}}    
    \put(86,35){\contour{white}{$\go m_8^-$}}    
    \put(56.5,92){\contour{white}{$\go m_7^-$}}    
    \put(56.5,2){\contour{white}{$\go m_7^+$}}    
    \put(58.5,22){\contour{white}{$\go m_6^+$}}    
    \put(76,54.5){\contour{white}{$\go m_1^-=\go m_1^+$}}    
    \put(86,81){\contour{white}{$\go m_8^+$}}    
    \put(40,35){\contour{white}{$\go M_7$}}    
    \put(4,43.5){\contour{white}{$\go M_8$}}    
    \put(4.5,62.5){\contour{white}{$\go M_1$}}    
    \put(4.5,92.5){\contour{white}{$\go M_9$}}    
    \put(47,50.5){\contour{white}{$\go M_6$}}    
    \put(3,32){\contour{white}{$\go q$}} 
	\end{small} 
  \end{overpic}
} 
\hfill
\end{center} 
\caption{Trajectories of the upper branch of the self-motion, where  the starting pose at time $t^-$ and 
the end pose at time $t^+$ are illustrated: (a) Example \ref{ex:1} and (b) Example \ref{ex:2}. 
Animations of these two self-motions can be downloaded from the homepage of the first author ({\tt http://www.geometrie.tuwien.ac.at/nawratil}).
}
  \label{fig2}
\end{figure}
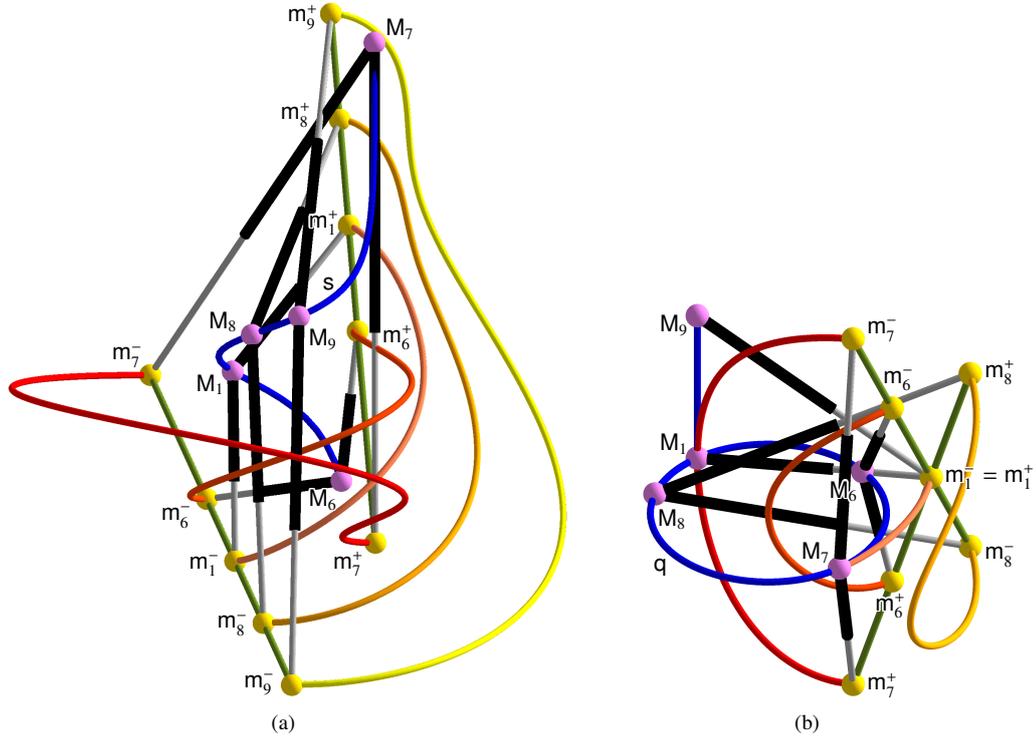

This can best be seen for the self-motions obtained for Type 5, as they belong to the class of 
Borel-Bricard motions. These are the only non-trivial motions where all points of the moving space 
have spherical trajectories (cf.\ \cite[Chapter VI]{bricard}; see also \cite{herve}). 
Note that this special case was also discussed in detail by Krames \cite[Section 5]{krames}.  
In this case $\Phi_1^*$ equals $x_0^2(C_5^2-a_5^2)+a_5^2(x_1^2+x_2^2)=0$. This already shows the following result: 

\begin{cor}\label{real5}
The 1-parametric set of self-motions given in Theorem \ref{lem3:type5} is real 
if  $|C_5|<|a_5|$ holds; complex otherwise. 
\end{cor}
Until now we are not able to give a corresponding easy characterization for 
designs of Type 1 and 2 with real/complex self-motions. 
But the following two examples prove that real self-motions exist:


\begin{ex}\label{ex:1} 
This example of a pentapod of Type 1 with a real self-motion is based on the formulas of $\Lambda_1,\Omega_2,\Omega_3,\Omega_4,\Pi_5$ 
given in the proof of Theorem \ref{lem2:type1}. 
The geometry of the pentapod is determined by: 
\begin{equation}
a_2=B_2=i,\quad a_3=B_3=-i,\quad a_4=2,\quad A_4=B_4=0,\quad A_5=B_5=C_5=1.
\end{equation}
For the leg-parameters:
\begin{equation}
R_1=\sqrt{3},\quad
p_2=-\frac{3}{25}-\frac{21}{25}i, \quad
p_3=-\frac{3}{25}+\frac{21}{25}i, \quad
p_4=-\frac{3}{5},\quad
p_5=\frac{46}{75},
\end{equation} 
which are in accordance with Theorem \ref{lem3:type1}, 
the pentapod has the following self-motion (under consideration of $x_0=1$; cf.\ Eq.\ (\ref{eq:motion})): 
\begin{align}
x_1&=\frac{7}{4}t^2-\frac{7}{5}t-\frac{161}{300}\mp\frac{T}{300}, &\quad
x_2&=\frac{1}{4}t^2-\frac{1}{5}t-\frac{23}{300}\pm\frac{7T}{300}, &\quad
x_3&=t, \\
y_1&=-\frac{1}{4}t^2+\frac{1}{5}t+\frac{59}{300}\mp\frac{7T}{300}, &\quad
y_2&=\frac{7}{4}t^2-\frac{7}{5}t-\frac{413}{300}\mp\frac{T}{300}, &\quad
y_3&=-2t+\frac{3}{5},
\end{align}
with $T=\sqrt{-(75t^2-30t-41)(75t^2-90t+31)}$. Both branches (upper and lower one) are real for $t\in\left[t^-,t^+\right]$ with
\begin{equation}
t^-=\frac{1}{5}-\frac{2}{15}\sqrt{33},\quad
t^+=\frac{1}{5}+\frac{2}{15}\sqrt{33}.
\end{equation}
For this example we also show how to compute the finite base anchor point 
$\go M$ (with coordinates $(A,B,C)^T$), the finite platform anchor point $\go m$ (with coordinates $(a,0,0)^T$) and the leg length $R$ 
of a further leg. Its corresponding sphere condition $\Lambda$ has to be a linear combination of the given equations $\Lambda_1,\Omega_2,\Omega_3,\Omega_4,\Pi_5$; i.e.\ 
\begin{equation}
\mu_1\Lambda_1+\mu_2\Omega_2+\mu_3\Omega_3+\mu_4\Omega_4+\mu_5\Pi_5-\Lambda=0
\end{equation}
for any choice of $n_0,x_0,\ldots ,x_3,y_0,\ldots, y_3$. Therefore their nine coefficients imply nine equations 
in the ten unknowns $\mu_1,\ldots,\mu_5,R,A,B,C,a$. This system has the following solution (in dependence of $a$): 
\begin{equation}
\begin{split}
\mu_1&=1,\quad
\mu_2=A+Bi,\quad
\mu_3=A-Bi,\quad
\mu_4=2C,\quad
\mu_5=2a, \\
R^2&=\frac{
(3a^2-8a+9)(25a^4-64a^3+146a^2-136a+100)
}
{
75(a^2+1)(a-2)^2
},
\end{split}
\end{equation}
with
\begin{equation}
A = \frac{a(a-1)}{a^2+1}, \quad
B = \frac{a(a+1)}{a^2+1},\quad
C = \frac{a}{a-2}. 
\end{equation}
The last equation gives the bijection $\sigma$ between points $\go m$ of $\go p$ and 
points $\go M$ of the irreducible straight cubic circle $\go s$ (cf.\ Eq.\ (\ref{cramer})). 

In Fig.\ \ref{fig2}a the trajectories of the platform anchor points $\go m_1$ ($a=0$),
$\go m_6$ ($a=1$), $\go m_7$ ($a=3$), $\go m_8$ ($a=-1$) and $\go m_9$ ($a=-2$) 
are displayed for the upper branch of the self-motion. \hfill $\diamond$
\end{ex}

\begin{ex}\label{ex:2} 
This example of a pentapod of Type 2 with a real self-motion is based on the formulas
of $\Lambda_1,\Omega_2,\Omega_3,\Omega_4,\Pi_5$ given in the proof of Theorem \ref{lem2:type1} 
under consideration of $a_4=0$ and $C_5=0$.  
The geometry of the pentapod is determined by: 
\begin{equation}
B_2=i,\quad B_3=-i,\quad 
a_2=1+i,\quad a_3=1-i,\quad
A_4=B_4=0,\quad A_5=B_5=1.
\end{equation}
For the leg-parameters:
\begin{equation}
R_1=2,\quad
p_2=-1-i, \quad
p_3=-1+i, \quad
p_4=0,\quad
p_5=1,
\end{equation} 
which are in accordance with Theorem \ref{lem3:type2}, 
the pentapod has the following self-motion (under consideration of $x_0=1$; cf.\ Eq.\ (\ref{eq:motion})): 
\begin{equation}
x_1=-\frac{1}{2}t^2, \quad
x_2=\pm\frac{T}{2}, \quad
x_3=t, \quad
y_1=\frac{1}{2}t^2+1\mp\frac{T}{2}, \quad
y_2=-\frac{1}{2}t^2-1\mp\frac{T}{2}, \quad
y_3=0,
\end{equation}
with $T=\sqrt{-t^4-4t^2+4}$. Both branches (upper and lower one) are real for $t\in\left[t^-,t^+\right]$ with
\begin{equation}
t^-=-\sqrt{2\sqrt{2}-2},\quad
t^+=\sqrt{2\sqrt{2}-2}.
\end{equation}
Analogous considerations as in Example \ref{ex:1} show the following bijection $\sigma$ between points $\go m$ of $\go p\setminus\left\{\go m_1\right\}$ and 
points $\go M$ of the circle $\go q\setminus\left\{\go M_1\right\}$:
\begin{equation}
A = \frac{a(a-2)}{a^2-2a+2},\quad  B = \frac{a^2}{a^2-2a+2}, \quad C=0.
\end{equation}
The point $\go m_1$ is mapped to the line $[\go M_1,\go M_9]$, which equals the $z$-axis of $\Sigma_0$. 

In Fig.\ \ref{fig2}b the trajectories of the platform anchor points $\go m_1=\go m_9$ ($a=0$),
$\go m_6$ ($a=1$), $\go m_7$ ($a=2$) and $\go m_8$ ($a=-1$)  
are displayed for the upper branch of the self-motion.  \hfill $\diamond$
\end{ex}

\begin{rmk} 
Further examples of pentapods of Type 1 and Type 2 are implied by 
a remarkable motion, where all points of a hyperboloid, which carries two reguli of lines, have spherical trajectories. 
This well-studied motion is also known as BBM-II motion in the literature (e.g.\ \cite{hartmann}).
It is known (cf.\ \cite[page 24]{hartmann} and \cite[page 188]{krames2}) that the corresponding sphere centers of lines, belonging to one 
regulus\footnote{The corresponding sphere centers of lines belonging to the other regulus are again 
located on a line (cf.\ \cite[page 24]{hartmann}), which imply  architecturally singular pentapod designs.},
are located on irreducible straight cubic circles, which imply examples of self-motions of Type 1. 
Note that there also exist degenerated cases where the hyperboloid splits up into two orthogonal planes, 
which imply examples of self-motions of Type 2. \hfill $\diamond$
\end{rmk}


\section{Conclusions for practical applications}\label{conclusion}

We introduced a novel kinematic mapping for pentapods with linear platform (cf.\ Section \ref{kinmap}), which 
can be used for an efficient solution of the direct kinematics problem. 
Beside this achievement we listed all pentapods with linear platform which 
\begin{itemize}[$\star$]
\item
are architecturally singular (cf.\ Corollary \ref{cor:arch}),
\item
possess self-motion (over $\CC$) without having an architecture singularity.  
\end{itemize}
The latter are either the designs ($\alpha,\beta,\gamma$) given in Section \ref{intro}, pentapods of 
Types 1,2,5 fulfilling the Duporcq condition (cf.\ Section \ref{sec:contemp_duporcq}) or the manipulator given in Theorem \ref{thm:planar}. 

Clearly, architecturally singular pentapods are not suited for practical application and therefore engineers should be aware of these designs. 
The usage of pentapods with self-motions within the design process is a double-edged sword; on the one side they should be avoided for 
reasons of safety\footnote{A self-motion is dangerous because it is uncontrollable and thus a hazard to man and machine.}
and on the other side they have a simplified direct kinematics for the following reason:

As the bonds are independent of the set of leg lengths (cf.\ Section \ref{sec:bond}), 
they always appear as solution of the direct kinematics problem 	
even though the given set $R_1,\ldots ,R_5$ does not cause a self-motion of the manipulator. 
Recall that a bond of a self-motion corresponds to a complex configuration on the boundary (cf.\ Remark \ref{max}). 
Therefore four solutions ($=$ two conjugate complex bonds of multiplicity 2; cf.\ Remark \ref{max}) of the direct kinematics of a pentapod 
with linear platform possessing a self-motion are always located on the boundary of $\mathcal{I}$. 
This yields the following corollary: 

\begin{cor}
A pentapod with linear platform possessing a self-motion (over $\CC$) can have a maximum of four real configurations 
(instead of generically 8, cf.\ Theorem \ref{los8}), if the given set of leg lengths does not imply a self-motion. 
The direct kinematic problem of these pentapods reduces to the solution of a polynomial of degree 4. 
\end{cor}

As this quartic equation can be solved explicitly, these pentapods seem to be of special interest for practical application.
We demonstrate this result on the basis of the following example:

\begin{ex}\label{ex:3}
Continuation of Example \ref{ex:1}: 
We consider the pentapod with platform anchor points 
$\go m_1,\go m_6,\ldots ,\go m_9$ and base anchor points $\go M_1,\go M_6,\ldots ,\go M_9$ 
of Example \ref{ex:1} (see Fig.\ \ref{fig2}a). But now we want to solve the direct kinematics problem of this 
pentapod of Type 1 for the following given set of leg lengths:
\begin{equation}
R_1=2,\quad R_6=1,\quad R_7=5,\quad R_8=3,\quad R_9=4,
\end{equation}
which does not cause a self-motion.  
We can solve the corresponding system of equations $\Lambda_1,\Lambda_6,\ldots,\Lambda_9$ for 
$n_0,y_0,y_1,y_2$ and $y_3$. Moreover we can set $x_0=1$. Now $\Phi_3$ of Eq.\ (\ref{phis}) is only linear in $x_1$ and $x_2$ and we can solve it 
for $x_1$. Then $\Phi_1$ and $\Phi_2$ are only quadratic in $x_2$, and therefore the resultant of these two expressions with respect to $x_2$ yields:
\begin{equation}
4316636297+69486876480x_3+241133479200x_3^2-291209472000x_3^3+76425120000x_3^4=0,
\end{equation}
which is only of degree 4 in $x_3$.
\hfill $\diamond$
\end{ex}

But it is even possible to use this advantage of self-motions without any risk (cf.\ footnote 9), by 
designing pentapods with linear platform, which only have complex self-motions. 
In the following we list two sets of such designs: 
\begin{itemize}
\item[(I)]
Pentapods of Type 5 with $|C_5|\geq|a_5|$ (cf.\ Corollary \ref{real5}). 
\item[(II)]
We can also solve the planar case (cf.\ Section \ref{p5coplanar}) with the bond based approach used for the 
non-planar one. This study shows, that a planar pentapod of $\mathcal{P}$ (cf.\ Definition \ref{df:P}) 
has a bond if and only if the vertex $\go V$ (cf.\ proof of Theorem \ref{thm:planar}) is an ideal point. 
Moreover the second necessary condition implied by the theory of bonds is only fulfilled if 
the affine relation (AR) holds\footnote{In this case the  bond (and its conjugate) are singular points of  the 5-fold $\mathcal{I}$ (cf.\ Remark \ref{max}).}. 
 
\begin{rmk}\label{quadratically} 
Note that the condition (AR) equals the 
linear constraint given in \cite{btt}, where the observation was reported that these planar pentapods with linear platform only possess 
a maximum of four real solutions of the direct kinematics problem (without giving an explanation for this behavior). 
This problem can even be solved quadratically as the solutions are symmetric with respect to the base plane (cf.\ \cite{btt}). 
Therefore this also holds for the design ($\gamma$) and for the designs ($\alpha,\beta$) under the extra condition of a planar base.  \hfill $\diamond$
\end{rmk}

The constraint (AR) is even sufficient for the existence of a self-motion (over $\CC$). 
Now we design the pentapods in a way that the distance between the parallel lines  
$[\go M_i,\go V]$ and $[\go M_j,\go V]$, which are fibers of the affinity $\kappa$, is 
\begin{enumerate}[$\bullet$]
\item
equal or less than the distance dist$(\go m_i,\go m_j)$ between their images: This yields the pentapods 
characterized in Theorem \ref{thm:planar}, which all have real self-motions (cf.\ footnote 8).
\item
greater than the distance dist$(\go m_i,\go m_j)$ between their images:  
Then the line $\go p$ cannot be oriented that 
dist$(\go M_i^{\prime},\go M_j^{\prime})$=dist$(\go m_i^{\prime},\go m_j^{\prime})$ holds, which already shows  
that this design-set (II) is free of real self-motion. 
\end{enumerate} 
\end{itemize}
Therefore the authors recommend engineers to design pentapods with linear platform 
within the set (I) in the non-planar case and 
within the set (II) in the planar one, respectively. 
For reasons of completeness we also give the following corollary: 

\begin{cor}
The following pentapods with linear platform, which do not possess a self-motion, have a maximum of six real configurations 
(instead of generically 8, cf.\ Theorem \ref{los8}): 
\begin{enumerate}
\item
The irreducible cubic $\go s$ of a pentapod of Type 1 is a cubic ellipse located on a cylinder of revolution.
\item
The conic $\go q$ of a pentapod of Type 2 belongs to a cylinder of revolution, where one generator is the line $\go g_1$.
\item
A pentapod of Type 5 with one of the following two properties: 
	\begin{enumerate}
	\item
	Either the base anchor points of possible leg-replacements are located on an irreducible cubic ellipse $\go s^*$ on a cylinder of revolution.
	\item
	or it has the following design (under consideration of footnote 1):
	$\go m_2=\go m_3$, $\go m_4=\go m_5$ and $[\go M_2,\go M_3]$ is parallel to $[\go M_4,\go M_5]$.
	\end{enumerate}
\item
Planar pentapod of $\mathcal{P}$, where the associated point $\go V$ is an ideal point (cf.\ Fig.\ \ref{fig1}c).  
\end{enumerate}
The direct kinematic problem of the listed pentapods reduces to the solution of a polynomial of degree 6.  
\end{cor}

\noindent
{\sc Proof:} 
Items 1, 2 and 3(a) are a direct consequence of the Theorems \ref{lem1:type1}, \ref{lem1:type2} and \ref{lem1:type5}, respectively.  
Item 4 follows from the above given discussion of design-set (II) and item 3(b) can be seen as its corresponding non-planar case (cf.\ Fig.\ \ref{fig1}c).   
In the latter case the cubic splits up into 3 parallel (but non-planar) lines. This case is hidden in item 2(a) of the proof of Theorem \ref{lem1:type5}. 

Finally, it should be noted that the direct kinematics problem is only cubic for item 4 as the pentapod is planar (cf.\ Remark \ref{quadratically}). 
\hfill $\BewEnde$\\

\noindent
We want to close the paper by referring to \cite{nawratil_new} where the remaining problem of 
characterizing designs of Type 1 and Type 2 without real self-motions is discussed, as they 
imply further "save" designs with a closed form solution.


\section*{Acknowledgments}
The first author's research is funded by the Austrian Science Fund (FWF): P24927-N25 - ``Stewart Gough platforms with self-motions''. 
The second author's research is supported by the Austrian Science Fund (FWF): 
W1214-N15/DK9 and P26607 - ``Algebraic Methods in Kinematics: Motion Factorisation and Bond Theory''.


\newpage

\section*{Appendix A: Proof of Theorem \ref{lem1:type2}}

We can start with the same set of equations as in the proof of Theorem \ref{lem1:type1}, but we have the extra conditions 
$\go m_1=\go m_4$ and that $\go M_1,\go M_2,\go M_3,\go M_5$ are located in a plane, which does not contain $\go M_4$.  
Due to the properties of Type 2 not all base points are coplanar ($\Rightarrow$ $B_2C_3C_5\neq 0$) 
and no two platform anchor points can coincide beside $\go m_1$ and $\go m_4$ ($\Rightarrow$  $a_1=a_4=0$). 
Therefore we can express the coplanarity of $\go M_1,\go M_2,\go M_3,\go M_5$ as 
\begin{equation}
A_3 = \frac{A_2B_3C_5-A_2B_5C_3+B_2C_3A_5}{B_2C_5}.
\end{equation}

We set $x_0=0$ and start the computation of the bonds: 
We can solve the equations $\Lambda_1,\Lambda_2,\Lambda_3,\Omega_4,\Pi_5$ for $y_0,y_1,y_2,y_3,n_0$ w.l.o.g..
Now the numerator of $\Gamma_4$ factors into $x_1F$ with $F$ of Eq.\ (\ref{ex:F}). 
Therefore we have to distinguish two cases:
\begin{enumerate}[1.]
\item
$x_1\neq 0$: 
In this case $F=0$ has to hold, which can be solve w.l.o.g. for $x_3$. 
Then the numerator of $\Gamma_6$ factors into: 
\begin{equation}
x_2a_3(x_2B_2+x_1A_2)(B_3C_5-B_5C_3+B_2C_3-B_2C_5).
\end{equation}  
The last factor cannot vanish, as otherwise the base anchor points $\go M_2,\go M_3,\go M_5$ are collinear, a contradiction. 
Therefore we remain with two cases: 
	\begin{enumerate}[a.]
	\item
	$x_2 = -x_1A_2/B_2$. Now the numerator of $\Gamma_1$ factors into: 
	\begin{equation}
	x_1^2\left[A_2^2(B_5^2+C_5^2)-2A_2A_5B_2B_5+B_2^2(A_5^2+C_5^2)\right]. 
	\end{equation}
	The discriminant with respect to $A_2$ equals $-C_5^2(A_5^2+B_5^2+C_5^2)$ and therefore we get a contradiction. 
	\item
	$x_2=0$: In this case the numerator of $\Gamma_1$ factors into $x_1^2[C_5^2+(A_2-A_5)^2]$
	which can also not vanish without contradiction. 
	\end{enumerate}
\item
$x_1=0$: This case is exactly the same as the one discussed in item 2 of the proof of Theorem \ref{lem1:type1}, 
which already yields the result. \hfill $\BewEnde$
\end{enumerate}


\section*{Appendix B: Proof of Theorem \ref{thm:type3}}
 
We can make leg-replacements such that 
$\go m_1=\go m_2=\go P_1$, $\go m_3=\go m_4=\go P_2$, $\go M_1,\go M_2\in\go L_1$ and 
$\go M_3,\go M_4\in\go L_2$. Moreover we can choose $\go M_2$ and $\go M_3$ as ideal points of 
$\go L_1$ and $\go L_2$, respectively. Therefore the point pairs $(\go M_i,\go m_i)$ determine 
sphere conditions $\Lambda_i$ for $i=1,4$ and Darboux condition $\Omega_i$ for $i=2,3$. 
Moreover we can assume that $\go M_1$ is located in the Darboux plane of $(\go M_2,\go m_2)$
and that $\go M_4$ is located in the Darboux plane of $(\go M_3,\go m_3)$.
Finally we can assume that $\go m_5$ is the ideal point of the line $\go p$. Therefore the 
point pairs $(\go M_5,\go m_5)$ determines a Mannheim condition $\Pi_5$. 

W.l.o.g.\ we choose the fixed frame $\Sigma_0$ that $\go M_1$ equals its origin and that $\go M_2$ and $\go M_3$ are located in the 
$xy$-plane symmetric with respect to the $x$-axis. Therefore the directions of $\go M_2$ and $\go M_3$ are given by 
$(1,B_2,0)$ and $(1,-B_2,0)$, respectively. 
Moreover we can define the moving frame $\Sigma$ in a way that $\go m_1$ is its origin. 
With respect to these coordinate systems our conditions can be written as: 
\begin{equation}
\begin{split}
\Omega_2&:\phm y_1+B_2y_2=0, \\
\Omega_3&:\phm (A_4-B_2B_4)x_0+a_3x_1-a_3B_2x_2+y_1-B_2y_2=0,\\
\Pi_5&:\phm p_5x_0+A_5x_1+B_5x_2+C_5x_3+y_0=0,
\end{split}
\end{equation}
where $(p_5,0,0)^T$ are the coordinates of the intersection point of the Mannheim plane and the $x$-axis of $\Sigma$. 
The equations of $\Lambda_i$ for $i=1,4$ are given in Eq.\ (\ref{sphere_compact}) under consideration of 
$A_1=B_1=C_1=a_1=0$ and $a_4=a_3$.

Now we set $x_0=0$ and prove that no bonds can exist. W.l.o.g.\ we can solve $\Lambda_1,\Omega_2,\Omega_3,\Pi_5$ for $n_0,y_0,y_1,y_2$. 
Then the numerator of $\Gamma_4$ can only vanish in the following two cases: 
\begin{enumerate}[1.]
\item
$x_1=B_2x_2$: The numerator of $\Gamma_2$ implies $y_3=0$. Now the numerator of $\Lambda_4$ equals:
\begin{equation}
a_4\left[
x_2(B_2A_4-B_2A_5+B_4-B_5)+(C_4-C_5)x_3
\right].
\end{equation}
We distinguish two cases: 
	\begin{enumerate}[(a)]
	\item
	$C_4\neq C_5$: Under this assumption we can solve the last factor for $x_3$. Then $x_2^2$ factors out from the numerator of $\Gamma_1$ and 
	we remain with only one condition, which is quadratic with respect to $A_5$. The corresponding discriminant 
	equals $-(C_4-C_5)^2(B_2^2+1)$
	and therefore no solution exists. 
	\item
	$C_4= C_5$: In this case $\Lambda_4$ can only vanish for $x_2=0$ as $B_2A_4-B_2A_5+B_4-B_5=0$ implies the collinearity of $\go M_3,\go M_4,\go M_5$, a contradiction. 
	Then $\Gamma_1$ cannot vanish without contradiction. 
	\end{enumerate}
\item
$x_1=-B_2x_2$: Now the numerator of $\Gamma_6$ equals $-x_2(x_3a_4+y_3)$. Therefore we have to distinguish two cases:
	\begin{enumerate}[(a)]
	\item
	$x_2=0$: Now $\Gamma_1$ and $\Gamma_2$ imply $x_3=y_3=0$, a contradiction.
	\item
	$y_3=-x_3a_4$: Then $\Lambda_4$ factors into 
	\begin{equation}
	a_4\left[x_2(B_2A_5-B_5)-C_5x_3\right].
	\end{equation}
	We distinguish two cases:
		\begin{enumerate}[i.]
		\item
		$C_5\neq 0$: Under this assumption we can solve the last factor for $x_3$. Then $x_2^2$ factors out from the numerator of $\Gamma_1$ and 
		we remain with only one condition, which is quadratic with respect to $A_5$. The corresponding discriminant equals $-C_5^2(B_2^2+1)$
		and therefore no solution exists. 
		\item
		$C_5=0$: In this case $\Lambda_4$ can only vanish for $x_2=0$ as $B_2A_5-B_5=0$ implies the collinearity of $\go M_1,\go M_2,\go M_5$, a contradiction. 
		Then $\Gamma_1$ cannot vanish without contradiction. \hfill $\BewEnde$
		\end{enumerate}
	\end{enumerate}
\end{enumerate}


\section*{Appendix C: Proof of Theorem \ref{lem2:type5}}

Due to Theorem \ref{lem1:type5} the irreducible cubic $\go s^*$ 
has three pairwise distinct points at infinity, which are denoted by 
$\go M_2, \go M_3,\go M_4$. Note that $\go M_4$ is real and that $\go M_2, \go M_3$ are conjugate complex; i.e.\ $\overline{\go M_2}=\go M_3$. 
The corresponding platform anchor points are denoted by $\go m_2, \go m_3, \go m_4$ where $\overline{\go m_2}=\go m_3$ holds. 
Therefore we get two Darboux conditions $\Omega_i$  implied by the point pairs $(\go M_i,\go m_i)$ for $i=2,3$. 
Moreover the point pair $(\go M_4,\go m_4)$ implies the angle condition $\sphericalangle_4$. 
The pentapod is completed by two sphere conditions $\Lambda_j$, which are determined by the two finite points $\go M_j$ and $\go m_j$ for $j=1,5$.

We choose the fixed frame $\Sigma_0$  that 
$\go M_1$ is the origin and that $\go M_2$ and $\go M_3$ are located in the $xy$-plane in 
direction $(1,B_2,0)$ and $(1,\overline{B_2},0)$, respectively. 
As $\go M_2, \go M_3, \go M_4$ cannot be collinear, $\go M_4$ is the ideal point in direction of $(A_4,B_4,1)$. 
Moreover we can define the moving frame $\Sigma$ in a way that $\go m_1$ is its origin. 
With respect to these coordinate systems our conditions can be written as: 
\begin{equation}
\Omega_i:\phm p_ix_0+a_ix_1+a_i\overline{B_i}x_2+y_1+ \overline{B_i}y_2=0, \qquad
\sphericalangle_4:\phm wx_0+A_4x_1+B_4x_2+x_3=0 
\end{equation}
where $(p_i,0,0)^T$ for $i=2,3$ are the coordinates of the intersection point of the Darboux plane and the $x$-axis of $\Sigma_0$
and $w$ denotes $\arccos{(\phi)}$. 
The equations of $\Lambda_j$ is given in Eq.\ (\ref{sphere_compact}) under consideration of 
$A_1=B_1=C_1=a_1=0$ for $j=1,5$.

In the following we substitute $B_2=B_r+iB_c$ and $a_2=a_r+ia_c$ with $B_r,B_c,a_r,a_c\in\RR$ and $B_ca_c\neq 0$. 
Then we set $x_0=0$ and start the computation of bonds: We can solve the equations $\Lambda_1,\Omega_2,\Omega_3,\sphericalangle_4,\Lambda_5$ 
for $x_3,y_1,y_2,y_3,n_0$ w.l.o.g..
Now the numerator of $\Gamma_4$ can only vanish without contradiction for $x_1 = (-B_r\pm iB_c)x_2$. We only discuss the upper sign as the 
conjugate solution can be done analogously. Then the numerator of $\Gamma_6$ can be solved for $y_0$ w.l.o.g.. 
From the numerator of $\Gamma_1$ we can factor out $x_2^2$ and we remain with only one condition.
From its imaginary part and real part we can compute $B_r$ and $B_c$ given in Eq.\ (\ref{ex:Br+Bc}).

Then we compute again the $8\times 9$ matrix $\Vkt J$ with respect to the obtained bond. 
For the necessary condition $rk(\Vkt J)<8$ the determinants of all  $8\times 8$ submatrices of $\Vkt J$ have to vanish. 
The numerator of the determinant of the $8\times 8$ submatrix of $\Vkt J$,  
obtained by removing the column steaming from the partial derivative with respect to $x_2$, factors into 
$(ia_r+a_c-ia_5)(ia_r+a_c)(A_4^2+B_4^2+1)L_1L_2$ with $L_1$ and $L_2$ of Eq.\ (\ref{ls}).  
Analogous arguments as in the proof of Theorem \ref{lem2:type1} shows that $\go s^*$ has to be a 
straight cubic circle. \hfill $\BewEnde$


\section*{Appendix D: Proof of Theorem \ref{lem3:type1}}

We use the same coordinatization as given in the first three paragraphs of the proof of Theorem \ref{lem2:type1} under consideration of $A_4=B_4=0$ and 
$B_2=i$. 

W.l.o.g.\ we can solve the equations $\Lambda_1,\Omega_2,\Omega_3,\Omega_4,\Pi_5$ for $y_0,y_1,y_2,y_3,n_0$.
Plugging the obtained expressions into $\Phi_i$ yields quadratic equation $\Phi_i^*$ in $x_0,\ldots,x_3$ for $i=1,2,3$. 
Therefore $\Phi_i^*$ corresponds with a quadric in the homogeneous 3-space (spanned by $x_0,\ldots,x_3$ ). 
In the following we want to determine the parameters $R_1,p_2,p_3,p_4,p_5$ in a way that these three quadrics have a curve in common. 
This can be done as follows: 

We compute the resultant  $\Xi_k$ of $\Phi_i^*$ and $\Phi_j^*$ with respect to $x_1$ for pairwise distinct $i,j,k\in\left\{1,2,3\right\}$. 
$\Xi_1$, $\Xi_2$ and $\Xi_3$ are homogeneous quartic expressions in $x_0,x_2,x_3$, but they are only quadratic with respect to $x_2$. 
Therefore we eliminate this unknown by computing the resultant $\Upsilon_k$ of $\Xi_i$ and $\Xi_j$ for pairwise distinct $i,j,k\in\left\{1,2,3\right\}$. 
Then the greatest common divisor $N$ of $\Upsilon_1$, $\Upsilon_2$ and $\Upsilon_3$ has to vanish. It turns out that $N$ has 411 terms and that 
it is homogeneous of degree 4 in $x_0,x_3$. For a self-motion of the line $\go p$ the expression $N$ has to be fulfilled independently of $x_0,x_3$.
Therefore we denote the coefficient of $x_0^ix_3^j$ of $N$ by $N_{ij}$.

In the following we show that the three conditions $E_3=0$, $E_4=0$, $E_5=0$ have to be fulfilled with: 
\begin{equation}
\begin{split}
E_3&=A_5(a_2a_3-a_4^2)+p_2(a_3-a_4)^2-i(a_2a_3-a_4^2)B_5, \\
E_4&=A_5(a_2a_3-a_4^2)+p_3(a_2-a_4)^2+i(a_2a_3-a_4^2)B_5, \\
E_5&=C_5(a_2a_3-a_4^2)-p_4(a_2-a_4)(a_3-a_4).
\end{split}
\end{equation}
This can be seen as follows: 
$N_{04}=0$ splits up into $E_3E_4$. 
W.l.o.g.\ we can set $E_3$ equal to zero and solve it for $p_2$. 
Then the numerator of $N_{13}$ factors into $-2a_3(A_5-iB_5)E_4E_5$.
Therefore we have to distinguish two cases:
\begin{enumerate}
\item
$E_4=0$: W.l.o.g.\ we can solve this equation for $p_3$. Then the numerator of $N_{22}$ splits up into 
$4a_2a_3(A_5^2+B_5^2)E_5^2$. Therefore $E_5=0$ has to hold and we are done. 
\item
$E_5=0$: W.l.o.g.\ we can solve this equation for $p_4$. Then the numerator of $N_{22}$ and $N_{31}$ splits up into: 
\begin{equation}
a_3(A_5-iB_5)E_4F_{22}[83],\qquad
a_3a_4C_5(A_5-iB_5)E_4F_{31}[73].
\end{equation}
Either $E_4=0$ holds and we are done or $F_{22}=0$ and $F_{31}=0$ have to hold. In the latter case we compute 
$F_{22}+F_{31}$, which factors into $a_3(a_2-a_4)^2(A_5-iB_5)E_4$. Therefore again $E_4=0$ has to be fulfilled. 
\end{enumerate} 
Summed up we have proven that $E_3=0$, $E_4=0$, $E_5=0$ have to hold. These equations can be solved for $p_2,p_3,p_4$
w.l.o.g.. Plugging the obtained expressions into $N$ shows that only the condition 
given in Eq.\ (\ref{last}) remains. We distinguish two cases:
\begin{itemize}[$\bullet$]
\item
$a_2a_3-a_4^2\neq 0$: In this case Eq.\ (\ref{last}) can always be solved for $p_5$. 
Then the self-motion can be computed by back-substitution; i.e.\ we compute the common factor of $\Xi_1,\Xi_2,\Xi_3$ 
and solve it for $x_2$ (which only appears quadratic) and finally the common factor of $\Phi_1^*$, $\Phi_2^*$ and $\Phi_3^*$ (which is linear in $x_1$)
gives the self-motion. 
\item
For the special case $a_2a_3-a_4^2=0$ we get $R_1^2=a_4^2$ from Eq.\ (\ref{last}). 
Moreover if we set $a_2=a_r+ia_c$ with $a_r,a_c\in\RR$ and 
$a_c\neq 0$ the condition $a_2a_3-a_4^2=0$ is equivalent with $a_r^2+a_c^2-a_4^2=0$, which can be solved for $a_4$ w.l.o.g.. 
Now it can easily be seen that $\Phi_1^*(a_r^2+a_c^2)=\Phi_2^*$ holds. Therefore we only remain with the following condition beside $\Phi_1^*$ (which equals $\Phi_1$ 
given in Eq.\ (\ref{phis})):  
\begin{equation}
\Phi_3^*:\phm x_0^2p_5-(x_1^2+x_2^2)a_r-x_3^2\sqrt{a_r^2+a_c^2}+x_0(x_1A_5+x_2B_5+x_3C_5)=0.
\end{equation}
This finishes the proof of the sufficiency. \hfill $\BewEnde$
\end{itemize}


\section*{Appendix E: Proof of Theorem \ref{lem3:type2}}

We start with the same set of equations as in the proof of 
Theorem \ref{lem3:type1} under consideration of $a_4=0$, $C_5=0$ 
and that no two platform anchor points can coincide beside $\go m_1$ and $\go m_4$. 

Moreover we can compute analogously the corresponding expression $N$, which has in this case only 204 terms. 
We denote the coefficient of $x_0^ix_3^j$ of $N$ again by $N_{ij}$.
In the following we show that the three conditions $E_6=0$, $E_7=0$ and $p_4=0$ have to be fulfilled with: 
\begin{equation}
E_6=a_2(A_5-iB_5)+a_3p_2, \qquad
E_7=a_3(A_5+iB_5)+a_2p_3.
\end{equation}
This can be seen as follows:  
$N_{04}=0$ splits up into $a_2a_3E_6E_7$. 
W.l.o.g.\ we can set $E_6$ equal to zero and solve it for $p_2$. 
Then the numerator of $N_{13}$ factors into $2a_2^2(A_5-iB_5)p_4E_7$.
Therefore we have to distinguish two cases:
\begin{enumerate}
\item
$E_7=0$: W.l.o.g.\ we can solve this equation for $p_3$. Then the numerator of $N_{22}$ splits up into 
$4a_2a_3(A_5^2+B_5^2)p_4^2$. Therefore $p_4=0$ has to hold and we are done. 
\item
$p_4=0$: Now the numerator of $N_{22}$ splits up into: 
\begin{equation}
a_2(A_5-iB_5)E_7\left[a_2p_3(A_5-iB_5)+R_1^2(a_2+a_3)-a_2(A_5^2+B_5^2+2a_3p_5)\right].
\end{equation}
If $E_7=0$ holds we are done. Therefore we express $p_3$ from the last factor, which can be done w.l.o.g..
Then $N_{31}$ is fulfilled identically and we only remain with $N_{40}$. Its numerator factors into: 
\begin{equation}
a_2R_1^2\left[
2a_2a_3p_5+(a_2+a_3)(A_5^2+B_5^2-R_1^2)
\right].
\end{equation}
Now we have to distinguish two cases:
	\begin{enumerate}
	\item
	We can solve the last factor for $p_5$ w.l.o.g.. Then it can easily be checked that $E_7$ is fulfilled and we are done. 
	\item
	$R_1=0$: This condition implies a spherical self-motion of the line $\go p$ with center $\go M_1=\go m_1$. 
	Now a one-parametric set of legs with base anchor points on the circle $\go q$ (the corresponding platform anchor points on $\go p$ 
	are given by $\sigma^{-1}$) can be attached without restricting the spherical self-motion. It can easily be seen that 
	any two legs of this set already fix the pose of $\go p$; hence no real self-motion of $\go p$ exists. 
	\end{enumerate}
\end{enumerate}
Summed up we have proven that $E_6=0$, $E_7=0$ and $p_4=0$ have to hold. $E_6=0$, $E_7=0$ can be solved for $p_2,p_3$
w.l.o.g.. Plugging the obtained expressions into $N$ shows that only the condition 
given in Eq.\ (\ref{fin:eq}) remains.  
This condition can always be solved for $p_5$ and the self-motion is again obtained by back-substitution, which finishes the 
proof of the sufficiency.  \hfill $\BewEnde$
\end{document}